%% file: proxy_paper_camera_ready.tex
\documentclass{article}

\usepackage{microtype}
\usepackage{graphicx}
\usepackage{subcaption}
\usepackage{booktabs} %

\usepackage{hyperref}
\usepackage{xurl}

\usepackage[accepted]{icml2026}

\usepackage{amsmath}
\usepackage{amssymb}
\usepackage{mathtools}
\usepackage{amsthm}
\usepackage{xspace}
\usepackage{multirow}
\usepackage{pgfplots}
\usepackage{rotating}
\usepackage{etoc}
\usepackage{pifont}

\pgfplotsset{compat=1.18}
\usepackage{enumitem}
\usepgfplotslibrary{groupplots}
\usepackage{tikz}
\pgfplotsset{
    /pgf/number format/use comma=false,
}

\usepackage{listings}
\usepackage{xcolor}
\usepackage{siunitx}
\usepackage[capitalize,noabbrev]{cleveref}
\crefname{section}{§}{§§}
\Crefname{section}{§}{§§}
\crefformat{section}{#2§#1#3}
\crefmultiformat{section}{\S\S#2#1#3}{ and~#2#1#3}{, #2#1#3}{, and~#2#1#3}
\creflabelformat{equation}{#2\textup{#1}#3}
\makeatletter
\AddToHook{cmd/appendix/before}{%
    \crefalias{section}{appendix}%
    \crefalias{subsection}{appendix}
}
\makeatother

\newcommand{\cmark}{\ding{51}}%
\newcommand{\xmark}{\ding{55}}%

\newcommand{\tok}[1]{\mathtt{\langle #1\rangle}}
\newcommand{\bortok}{\tok{raw}}
\newcommand{\eortok}{\tok{/raw}}
\newcommand{\boctok}{\tok{comp}}
\newcommand{\eoctok}{\tok{/comp}}

\theoremstyle{plain}

\theoremstyle{definition}

\theoremstyle{remark}

\usepackage[textsize=tiny]{todonotes}

\input{math_commands.tex}

\icmltitlerunning{Proxy Compression for Language Modeling}

\begin{document}

\twocolumn[
  \icmltitle{Proxy Compression for Language Modeling}

  \icmlsetsymbol{equal}{*}

  \begin{icmlauthorlist}
    \icmlauthor{Lin Zheng}{equal,hku}
    \icmlauthor{Xinyu Li}{equal,hku}
    \icmlauthor{Qian Liu}{comp}
    \icmlauthor{Xiachong Feng}{hku}
    \icmlauthor{Lingpeng Kong}{hku}
  \end{icmlauthorlist}

  \icmlaffiliation{hku}{University of Hong Kong}
  \icmlaffiliation{comp}{TikTok}

  \icmlcorrespondingauthor{Lin Zheng}{lzheng2@cs.hku.hk}

  \icmlkeywords{language modeling, byte-level modeling, tokenization, compression, efficient training}

  \vskip 0.3in
]

\printAffiliationsAndNotice{\icmlEqualContribution}

\begin{abstract}
Modern language models are trained almost exclusively on token sequences produced by a fixed \emph{tokenizer}, an external lossless compressor often over UTF‑8 byte sequences, thereby coupling the model to that compressor.
This work introduces proxy compression, an alternative training scheme that preserves the efficiency benefits of compressed inputs while providing an end-to-end, raw-byte interface at inference time.
During training, a single language model is jointly trained on raw byte sequences and compressed views generated by external compressors; through the process, the model learns to internally align compressed sequences and raw bytes. This alignment enables strong transfer between the two formats, even when training predominantly on compressed inputs that are discarded at inference.
Extensive experiments on code language modeling demonstrate that proxy compression substantially improves training efficiency and significantly outperforms pure byte-level baselines given fixed compute budgets.
As model scale increases, these gains become more pronounced, and proxy-trained models eventually match or surpass tokenizer approaches, all while operating solely on raw bytes and retaining the inherent robustness of byte-level modeling.
Our code is available at \url{https://github.com/LZhengisme/proxy-compression}.
\end{abstract}

\section{Introduction}
\label{introduction}

\begin{figure*}[tb]
\centering
{\includegraphics[width=0.96\textwidth]{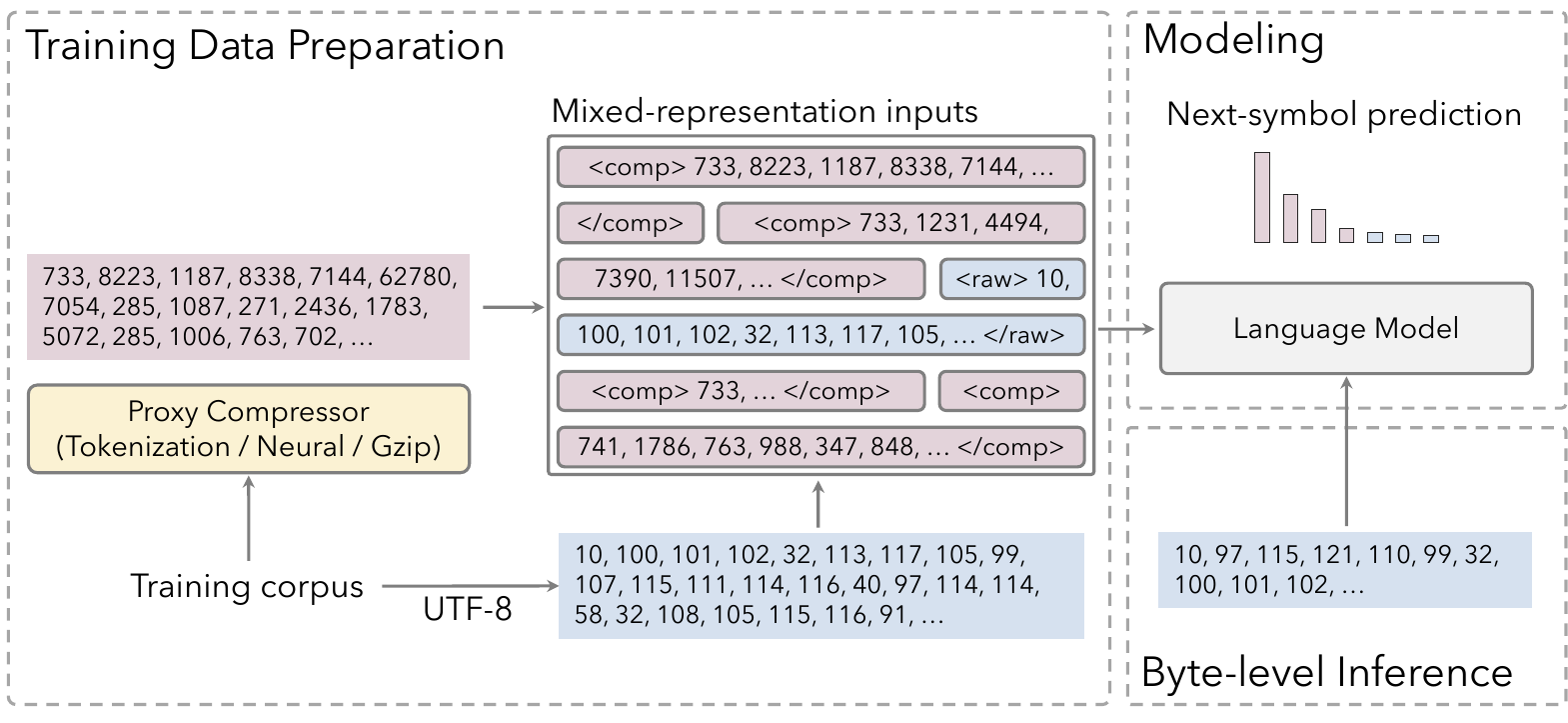}}
\caption{Overview of proxy compression for language modeling. During training, we prepare mixed-representation inputs by combining compressed sequences with raw UTF-8 bytes, which are packed together to train a single language model with next-symbol prediction over both representations. Different representations are associated with special sentinels, such as $\boctok$, $\eoctok$ for compressed data and $\bortok$ and $\eortok$ for raw data. At inference time, the proxy compressor is discarded entirely, and the model operates solely on raw bytes. By training primarily on compressed data (e.g., 90\% of training data in this work), this approach captures the training efficiency benefits of compressed data without hard-wiring the compressor into the model's interface.}
\label{fig:overview}
\end{figure*}

Modern language models are almost always trained on compressed views of data rather than on the raw format \citep{bengio2003neural}. In practice, the compressor, most commonly a tokenizer \citep{sennrich2016bpe,kudo2018sentencepiece} or arithmetic coding with another language model \citep{lester2024training}, maps the raw input into a shorter sequence of discrete tokens that the model actually processes. This compression is computationally essential, as it reduces the sequence length and thus enables efficient training.

However, this design couples the entire modeling stack to a fixed external compressor. Every input and output must pass through it, and at the level of raw data,\footnote{We use \emph{raw} to refer to the UTF-8 byte encoding of text, contrasting it with compressed sequences often derived from that byte stream, such as tokenized text.} the model is no longer strictly end-to-end, as it learns only to manipulate the compressor's outputs. In the case of hand-crafted tokenizers, this coupling introduces well-documented artifacts \citep{phan2024understanding}, including prompt boundary problems \citep{microsoft2023guidance,lundberg2023tokenhealing,athiwaratkun2024token,hayase2025sampling}, retokenization drift \citep{vllm2025retokenizationdrift}, under-trained or glitch tokens \citep{rumbelow2023solidgoldmagikarp,land2024fishing,wang2024tokenizationmatters,yang2024rethinking,yang2024problematictokens}, data mixture leakage \citep{hayase2024datamixture}, biases against low-resource languages \citep{petrov2023languageunfairness,ahia2023languagescost,limisiewicz2024myte}, and increased sensitivity to adversarial inputs \citep{sun2020advbert,pagnoni2024blt}.

\begin{figure}[tb]
\centering
{\includegraphics[width=0.97\columnwidth]{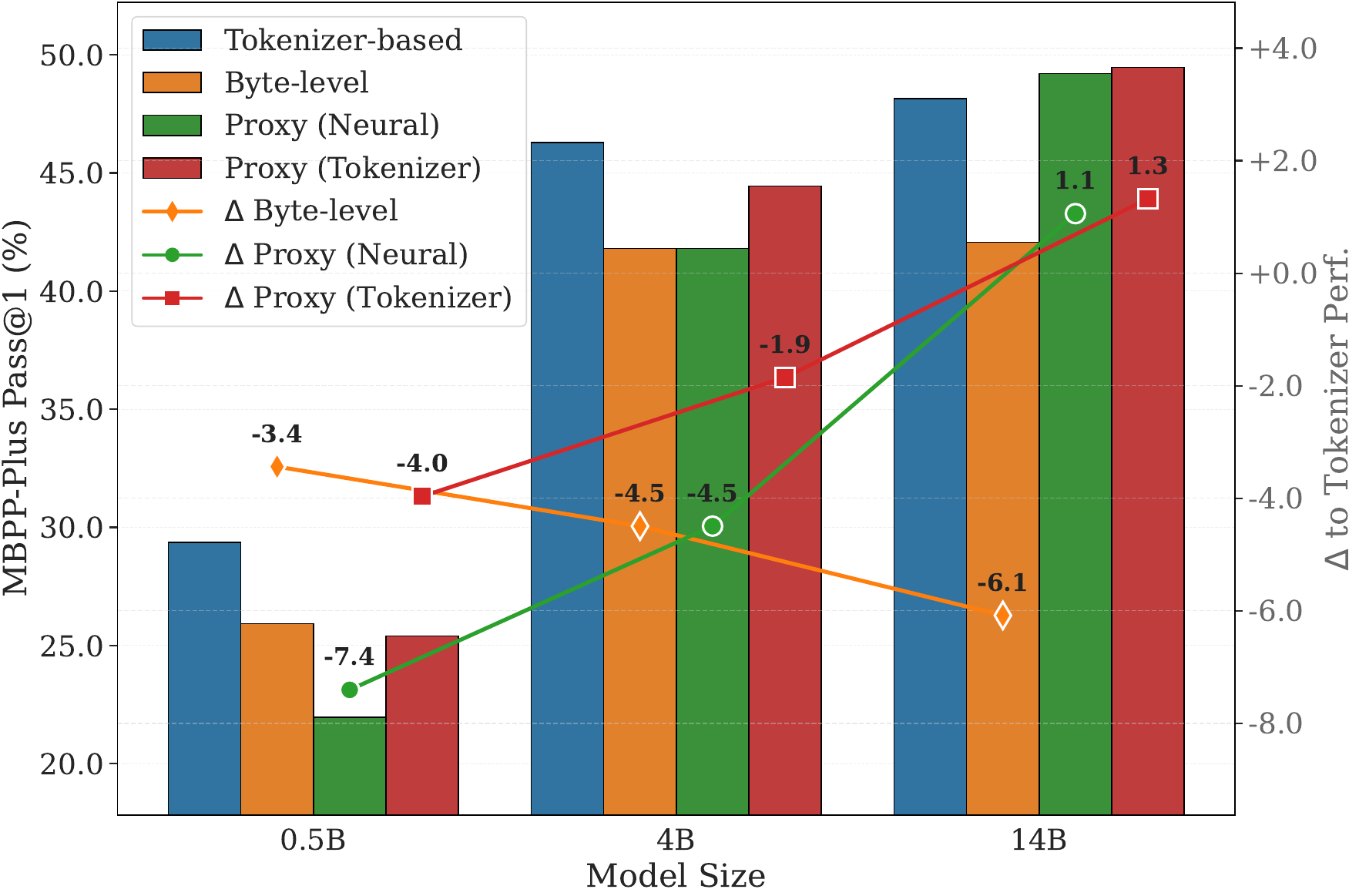}}
\caption{Model performance (Pass@1) on MBPP-Plus across model scales. Bars show absolute performance (left axis); lines show the performance gap ($\Delta$) relative to the tokenizer baseline (right axis). While byte-level models exhibit a persistent or widening gap, proxy-based models progressively close the gap as the model scale increases.}
\label{fig:perf-increase-with-scale}
\end{figure}

In this work, we ask whether it is possible to keep the efficiency benefits of training on compressed data without hard-wiring the compressor into the model's interface. We address this with \emph{proxy compression} (\cref{fig:overview}): treating external compressors as a \emph{training-time proxy} rather than a permanent part of the pipeline. Given a training corpus, we apply an external compressor to produce compressed views of each sequence and mix these compressed sequences with raw UTF-8 counterparts during training. A single language model is trained to perform next-symbol prediction over both representations jointly. At inference time, we can discard the compressor entirely and run the model on raw bytes alone; the compressor has served only to provide shorter proxy sequences during training for efficiency. Our central finding enabling proxy compression is strong \emph{cross-representation transfer}: although the majority of training data is specified in compressed form for efficiency (e.g., 90\%), the model performs surprisingly well on raw-byte inference. Moreover, transfer strength of proxy compression grows with model scale: small models demonstrate weak or negative transfer, while larger models with proxy compression can match or surpass ordinary language models with hard-wired compressors (\cref{fig:perf-increase-with-scale}).

Conceptually, proxy compression makes the learning objective more demanding in a useful way. The model must predict in two different code spaces and implicitly align them, learning an internal mapping between proxy codes and bytes. This encourages the model to treat compressed sequences as informative hints rather than a complete substitute for the raw byte stream. From a coding perspective, a language model defines a compressor via its predictive distribution \citep{deletang2024lmiscompression}; proxy compression can thus be viewed as partially delegating compression to an external proxy while training the model to remain an effective compressor over raw bytes. This navigates a fundamental trade-off: training on purely compressed data is efficient but constrains the model to patterns the compressor exposes, while training with pure raw bytes captures all structure but at high computational cost. Proxy compression gains efficiency from compressed inputs while staying grounded in the underlying byte-level distribution.

We instantiate proxy compression with several families of compressors (\cref{method}): tokenizer-based compression (\cref{method:token-proxies}), which uses conventional tokenizers as proxy compressors; neural proxy compressors (\cref{method:neural-proxies}), which combine a separately trained byte-level model with arithmetic coding \citep{lester2024training}; and generic, structure-agnostic compression via gzip (\cref{method:gzip-proxies}). Our experiments on code language modeling (\cref{experiments}) demonstrate that proxy compression yields strong transfer from compressed training to raw-byte inference. On downstream benchmarks (\cref{experiments:main-transfer}), models that see only about 10\% of their training samples in raw UTF-8 nonetheless outperform pure raw-byte baselines under a fixed compute budget, as they can consume substantially more compressed data while benefiting from strong transfer; at scale, they match or surpass strong tokenizer-based baselines. In-context transfer probes (\cref{experiments:in-context-transfer}) show that proxy-trained models can near-perfectly recover raw inputs from their compressed forms in context. We then investigate what makes a good proxy compressor in \cref{experiments:proxy-compression-comp}, where gzip proxies fail to transfer, possibly due to their unstable outputs, whereas both tokenizer-based and neural proxies support strong transfer. Robustness evaluations (\cref{experiments:robustness}) show that proxy-trained models retain most of the inherent robustness of raw byte-level modeling.

In summary, this work makes the following contributions:
\begin{itemize}[leftmargin=*,itemsep=2pt,topsep=2pt]
\item We introduce \emph{proxy compression}, a mixed-representation training scheme that enables efficient training over compressed data while keeping a simple, raw-byte inference interface, without modifying model architectures.
\item We demonstrate \emph{strong cross-representation transfer} with proxy compression: models trained predominantly on proxy-compressed inputs substantially outperform byte-level baselines and rival tokenizer-based approaches at scale, with the gap narrowing as the model scale increases.
\item We systematically study several proxy compressors, including generic gzip, tokenizer-based compression, and arithmetic-coded neural proxies, showing that more structured compressors (tokenizer-based and neural) are highly effective proxies for training language models, while gzip-based compression fails to transfer effectively.
\end{itemize}

\section{Proxy Compression}
\label{method}
We introduce \emph{proxy compression}, a training scheme that retains the efficiency of compressed inputs while exposing an end-to-end byte-level interface at inference. \cref{method:overview} describes the core framework, \cref{method:token-proxies,method:neural-proxies,method:gzip-proxies} present concrete instantiations of the proxy compressor, and \cref{method:impl_details} provides implementation details.

\subsection{Overview}
\label{method:overview}

Let $x_{\text{raw}}$ denote a raw input sample in UTF-8 bytes, and let $x_{\text{comp}} \coloneqq f(x_{\text{raw}})$ be a representation compressed by a proxy compressor $f$. We assume $x_{\text{comp}}$ is a sequence of discrete symbols from a finite vocabulary, so that both formats can be handled seamlessly by a standard language model.

\paragraph{Mixed-Representation Training.} Our training pipeline operates at the sample level. For each input $x_{\text{raw}}$, we draw a Bernoulli variable such that with probability $r$, the sample is presented in compressed form $x_{\text{comp}}$, otherwise it remains as $x_{\text{raw}}$. The resulting sequences are packed into fixed-length contexts, which may contain both formats within the same context, and a single standard autoregressive model is trained with the usual next-symbol prediction objective on both compressed and raw sequences. Crucially, this scheme requires \emph{no architectural changes}; all modifications are confined to the data preprocessing pipeline.

\paragraph{Byte-level Inference.}
At inference time, the model operates \emph{exclusively} on raw UTF-8 bytes; the proxy compressor is used only during training and can be discarded afterwards. This decoupling leverages compression for training efficiency while retaining a universal byte-level interface.

\paragraph{Format Sentinel Tokens.} To make different representations explicitly distinguishable, we wrap each sequence with special sentinel tokens that indicate its format. Raw and compressed sequences are encoded as $\big[\bortok \circ x_{\text{raw}} \circ \eortok \big]$ and $\big[\boctok \circ x_{\text{comp}} \circ \eoctok \big]$, respectively, where $\circ$ denotes concatenation. These markers allow the model to condition its predictions on the representation type.\footnote{Sentinel tokens also improve performance at smaller model scales with diminishing gains at larger scales; see \cref{app:additional-exp-results:sentinel-ablation}.}

\paragraph{In-context Translation Pairing.} The sampling scheme above treats raw and compressed samples independently. To further encourage cross-representation alignment, we introduce \emph{in-context translation pairing}, which optionally presents both views of the same sample concatenated within a single context (the ordering between $x_{\text{raw}}$ and $x_{\text{comp}}$ is randomized with equal probability),
\begin{align*}
    \big[\bortok \circ x_{\text{raw}} \circ \eortok \circ \boctok \circ x_{\text{comp}} \circ \eoctok \big].
\end{align*}
These paired sequences encourage the model to predict one representation conditioned on the other, effectively learning an in-context ``translation'' between formats. By default, pairing is enabled only during an initial warm-up phase (\cref{method:impl_details}); further analysis is provided in \cref{app:additional-exp-results:pairing-strategy}.

Note that mixed-representation training usually yields a lower effective compression rate than the proxy compressor $f$ alone; see \cref{app:compression-ratio-derivation} for more detailed discussion. We next describe concrete instantiations of $f$ that produce discrete symbol sequences amenable to offline preprocessing.

\subsection{Tokenizer-based Compressors}
\label{method:token-proxies}
A natural instantiation of the proxy compressor $f$ is standard tokenization. Given a trained tokenizer \citep{sennrich2016bpe,kudo2018sentencepiece,liu2025superbpe}, it segments $x_{\text{raw}}$ into token indices from a fixed-size vocabulary to obtain $x_{\text{comp}}$. This satisfies our design criteria: the output is a discrete sequence, and tokenization can be performed entirely offline. Conceptually, the use of tokenization for proxy compression differs from conventional tokenizer-based language model training. Whereas a standard tokenizer-based model operates exclusively in token space and almost never sees raw bytes, our approach instead treats tokens as a \emph{training-time proxy} for efficiency while still retaining a raw-byte interface at inference. We also explored alternative token encodings (e.g., mapping token IDs to fixed-length byte sequences); these variants did not improve over direct token-index representations and are discussed in \cref{app:impl-details:tokenizer}.

\subsection{Neural Compressors}
\label{method:neural-proxies}

Beyond tokenization, we study neural proxy compressors based on arithmetic coding~\citep{lester2024training}. We train a small byte-level language model (${\sim}40$M parameters) on raw UTF-8 data to obtain conditional probability estimates for each byte position of the input $x_{\text{raw}}$. We then apply arithmetic coding with equal-information windows \citep{lester2024training} to $x_{\text{raw}}$ using these probabilities to produce a compressed bitstream, which is then packed into discrete symbols to form $x_{\text{comp}}$.

A naive implementation of this pipeline is prohibitively slow due to the sequential nature of arithmetic coding. To enable neural compressors at scale, we introduce an \emph{entropy-based segmentation} strategy: we use the compressor model to estimate per-byte entropy, identify segment boundaries at high-entropy positions, and compress segments independently in parallel. This is essential for making neural proxy compression practical on large-scale corpora, not only improving compression throughput but also yielding better empirical performance; full details are provided in \cref{app:impl-details:neural}.

\paragraph{Neurally Compressed Representations are Fuzzy.}
Neural compressors behave quite differently from tokenizer-based compressors. Although each raw input $x_{\text{raw}}$ maps deterministically to a unique compressed sequence $x_{\text{comp}}$, the reverse mapping is generally \emph{not} unique: given only $x_{\text{comp}}$ and no record of the compressing model probabilities used during encoding, there might be multiple distinct raw byte sequences corresponding to the same compressed stream. We refer to the compressed inputs as ``fuzzy'' in the sense that neural compression induces a many-to-one mapping from raw bytes to compressed symbols.

Importantly, this ambiguity is highly structured rather than arbitrary. As we analyze in \cref{experiments:proxy-compression-comp}, collisions typically group together raw byte sequences that differ only in superficial, low-entropy details (such as whitespace, newlines, and indentation), while preserving their semantic content. Unlike tokenization, we cannot losslessly decode $x_{\text{comp}}$ back to $x_{\text{raw}}$. However, for proxy compression this controlled ambiguity turns out to be beneficial: it abstracts away formatting noise and implicitly shares representations across semantically equivalent patterns.

\subsection{Gzip Compressors}
\label{method:gzip-proxies}
We also instantiate $f$ using general-purpose compressors. We apply gzip (via Python's standard library with \texttt{gzip.compress(...,mtime=0)} to remove variation due to headers and trailers) directly to $x_{\text{raw}}$ and treat the output byte stream as $x_{\text{comp}}$. On our corpus, gzip achieves a compression rate of roughly $2.5\times$ over $x_{\text{raw}}$. Unlike tokenizer-based and neural proxies, gzip produces streams whose local patterns depend strongly on document-specific compression state; we later show that this results in substantially less stable representations (\cref{experiments:proxy-compression-comp}).

\subsection{Implementation Details}
\label{method:impl_details}

\paragraph{Mixing Training Schedule.}
We set the mixing rate $r = 0.9$, so that 90\% of training samples are presented in compressed form and only 10\% as raw bytes. In this case, tokenizer-based and neural proxy compressors achieve average compression rates of roughly $2.9\times$ and $2.6\times$, respectively. During an initial warm-up phase (first 10k steps), we enable in-context translation pairing (\cref{method:overview}) and linearly increase $r$ from 0.4 to 0.9; after warm-up, pairing is disabled and $r$ remains fixed at 0.9. We also explored starting from different rates such as $r=0.1$ and observed no significant difference in final performance. All proxy-trained models are evaluated exclusively on raw byte inputs at inference time unless otherwise noted (\cref{experiments:analyses}).

\paragraph{Vocabulary.}
Since input sequences now consist of both compressed and raw sequences, we use a shared vocabulary to accommodate both raw bytes and compressed symbols, partitioned as follows: the first $64$ indices in the vocabulary are reserved for special sentinel tokens, followed by $256$ entries for raw UTF-8 byte values, with remaining indices assigned to compressed symbols. For tokenizer-based proxy compressors, we use the OpenCoder tokenizer vocabulary \citep{huang2024opencoder} with $96{,}640$ symbols. For neural compression, we pack every 16 bits into a compressed symbol (a $65{,}536$-way alphabet). The gzip-based proxy compressor works at the byte level over 256 compressed symbols.

\section{Experiments}
\label{experiments}

In this section, we empirically evaluate proxy compressed inputs. We first describe the experimental setup in \cref{experiments:setup}, study transfer from proxy compressors at scale in \cref{experiments:main-transfer,experiments:in-context-transfer}, compare different proxy compressors (\cref{experiments:proxy-compression-comp}), and finally evaluate the robustness of our approach in \cref{experiments:robustness}.

\subsection{Setup}
\label{experiments:setup}

We center our empirical study on code for compute-matched comparisons across input representations; complementary natural-language experiments at 1.5B parameters (\cref{app:additional-exp-results:natural-language}) verify that the transfer mechanism is not domain-specific. All training experiments use the RefineCode corpus~\citep{huang2024opencoder}, where we primarily use its Python subset, totaling roughly 270~GB of Python source code, and its full GitHub split (used in \cref{experiments:main-transfer}) with approximately 3.3~TB of code across multiple programming languages. We train a family of language models following the EvaByte architecture \citep{zheng2025evabyte} at 0.5B, 1.5B, 4B, 7B, and 14B parameters for all input representations. We choose EvaByte because it provides both strong modeling quality and efficient byte-level decoding: our architecture ablation shows that EvaByte achieves lower validation BPB while matching OpenCoder \citep{huang2024opencoder}, a Llama-based transformer architecture \citep{dubey2024llama3}, on downstream code generation (\cref{app:fig:ablation:model_arch}), and its efficient attention and multi-byte prediction mitigate the longer sequences induced by byte-level inference. Training runs for 50K steps with a fixed batch size of 2M sequence symbols (e.g., UTF-8 bytes, tokens, etc.); this yields a comparable training FLOPs budget across representations. As a result, models operating on different representations can consume different amounts of raw data per step. We compare against two baselines: (i) a \emph{byte-level} model trained entirely on raw UTF-8 bytes, and (ii) a \emph{tokenizer-based} model trained on BPE tokens using the OpenCoder tokenizer~\citep{huang2024opencoder}. For evaluation, we focus on code generation tasks including HumanEval \citep{chen2021humaneval}, MBPP \citep{austin2021mbpp}, and their EvalPlus variants \citep{liu2023evalplus}. Full experimental details are provided in \cref{app:additional-details}.

\subsection{Main Results: Scalable Downstream Transfer}
\label{experiments:main-transfer}
\begin{table}[t]
    \centering
    \caption{Downstream pass@1 performance on HumanEval-Plus and MBPP-Plus across different model sizes and input representations. CR denotes compression rate (avg. bytes per symbol). All models are trained with a fixed budget of 100B symbols (tokens or bytes); larger models (7B, 14B) may be undertrained relative to compute-optimal scaling~\citep{hoffmann2022chinchilla}.}
    \label{tab:downstream-transfer}
    \resizebox{\columnwidth}{!}{
    \begin{tabular}{l l c c c c c c}
        \toprule
        \multirow{2}{*}{Task} & \multirow{2}{*}{Model} & \multirow{2}{*}{CR} & \multicolumn{5}{c}{Model Size} \\
        & & & 0.5B & 1.5B & 4B & 7B & 14B \\
        \midrule
        \multirow{4}{*}{HumanEval-Plus}
            & Tokenizer-based & 3.7 & \textbf{17.7} & 18.3 & \textbf{28.0} & \textbf{28.7} & 29.3 \\
            & Byte-level & 1.0 & 15.9 & 18.3 & 22.0 & 23.8 & 24.4 \\
            & Proxy (Neural) & 2.6 & 13.4 & 18.3 & 22.6 & 26.8 & 29.9 \\
            & Proxy (Tokenizer) & 2.9 & 12.2 & \textbf{20.7} & 24.4 & 26.2 & \textbf{30.5} \\
        \midrule
        \multirow{4}{*}{MBPP-Plus}
            & Tokenizer-based & 3.7 & \textbf{29.4} & \textbf{41.0} & \textbf{46.3} & 45.2 & 48.1 \\
            & Byte-level & 1.0 & 25.9 & 33.6 & 41.8 & 41.3 & 42.1 \\
            & Proxy (Neural) & 2.6 & 22.0 & 29.6 & 41.8 & 41.8 & 49.2 \\
            & Proxy (Tokenizer) & 2.9 & 25.4 & 38.4 & 44.4 & \textbf{45.5} & \textbf{49.5} \\
        \bottomrule
    \end{tabular}}
\end{table}

In this section, we address the central question: \emph{can we train primarily on proxy-compressed inputs yet deploy the model purely on raw bytes, without ever exposing compressed formats at inference time?} Concretely, we evaluate whether mixed-representation training with proxy compressors yields effective transfer to raw byte streams on downstream code generation tasks.

\paragraph{Transfer Scales with Model Size.}
As shown in \cref{tab:downstream-transfer}, proxy-compressed inputs yield strong and scalable transfer on downstream tasks. Despite observing raw bytes on only $10\%$ of training samples, proxy-trained byte models closely track or outperform the fully byte-level baseline at model sizes above 1.5B parameters. This indicates that compressed views effectively compensate for the significantly reduced raw-byte exposure by consuming more data and that cross-representation transfer is strong enough to recover most of the performance of fully byte-level training. Moreover, the advantage of proxy training becomes more pronounced with scale. At larger model sizes, proxy-trained models not only significantly outperform byte-level baselines, but also start to achieve or surpass the performance of tokenizer-based models. This can be attributed to the greater capacity of larger models to encode the alignment between compressed and raw views in their weights. Both neural and tokenizer-based proxies perform competitively, with neural compression achieving slightly lower compression (2.6$\times$) than the tokenizer-based compressor (2.9$\times$), yet still matching its performance at scale.

\paragraph{Data versus Compute Efficiency.}
Byte-level and tokenizer-based models exhibit a well-known trade-off~\citep{xue2022byt5,zheng2025evabyte}: tokenizer-based models are more compute-efficient (better performance under a fixed FLOPs budget, as they consume more data per unit of compute), while byte models are more data-efficient (better performance under the same amount of training data, as they allocate more optimization steps and thus more compute per training example).\footnote{This trade-off does not always hold when varying the scale and training horizon; please see \cref{app:additional-exp-results:data-compute-efficiency} for further analysis.}

\begin{figure}[tb]
\centering
\begin{subfigure}[b]{0.49\columnwidth} 
  \centering
  {\includegraphics[width=\textwidth]{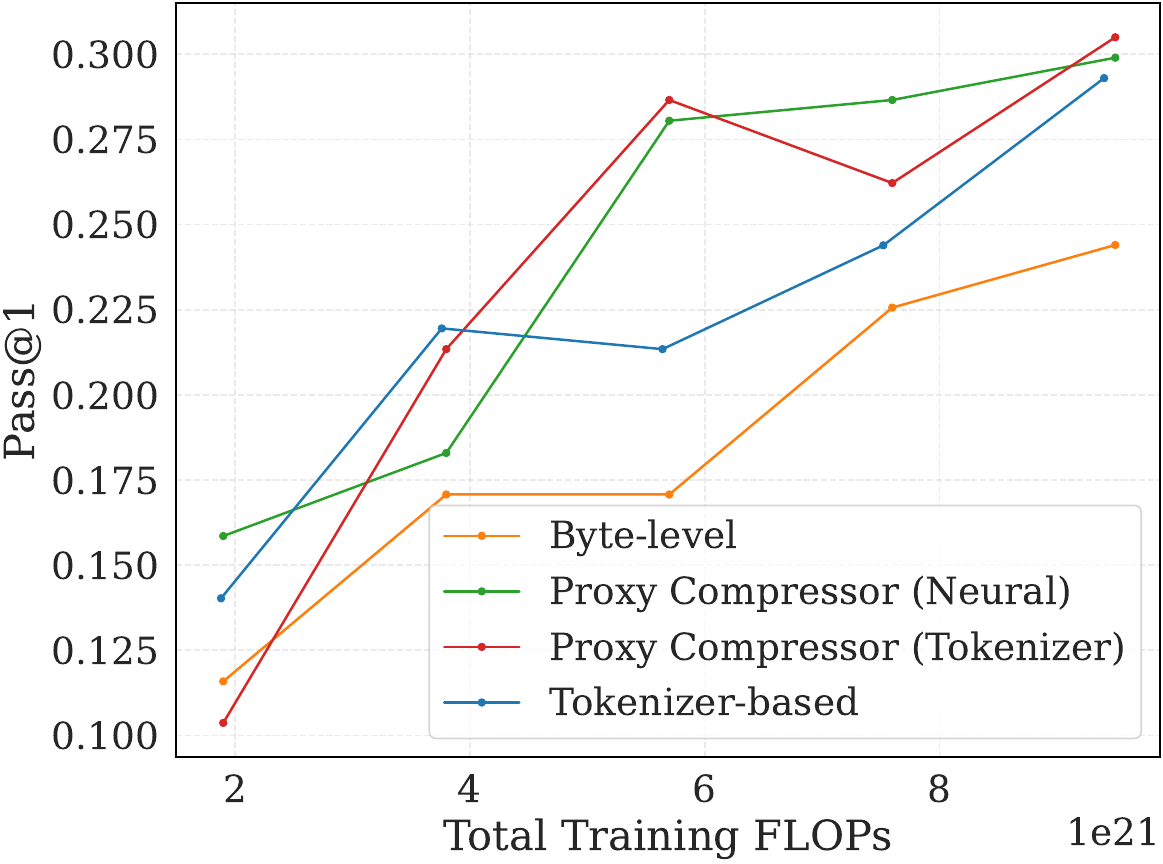}}
\end{subfigure}\hfill
\begin{subfigure}[b]{0.49\columnwidth}
  \centering
  {\includegraphics[width=\textwidth]{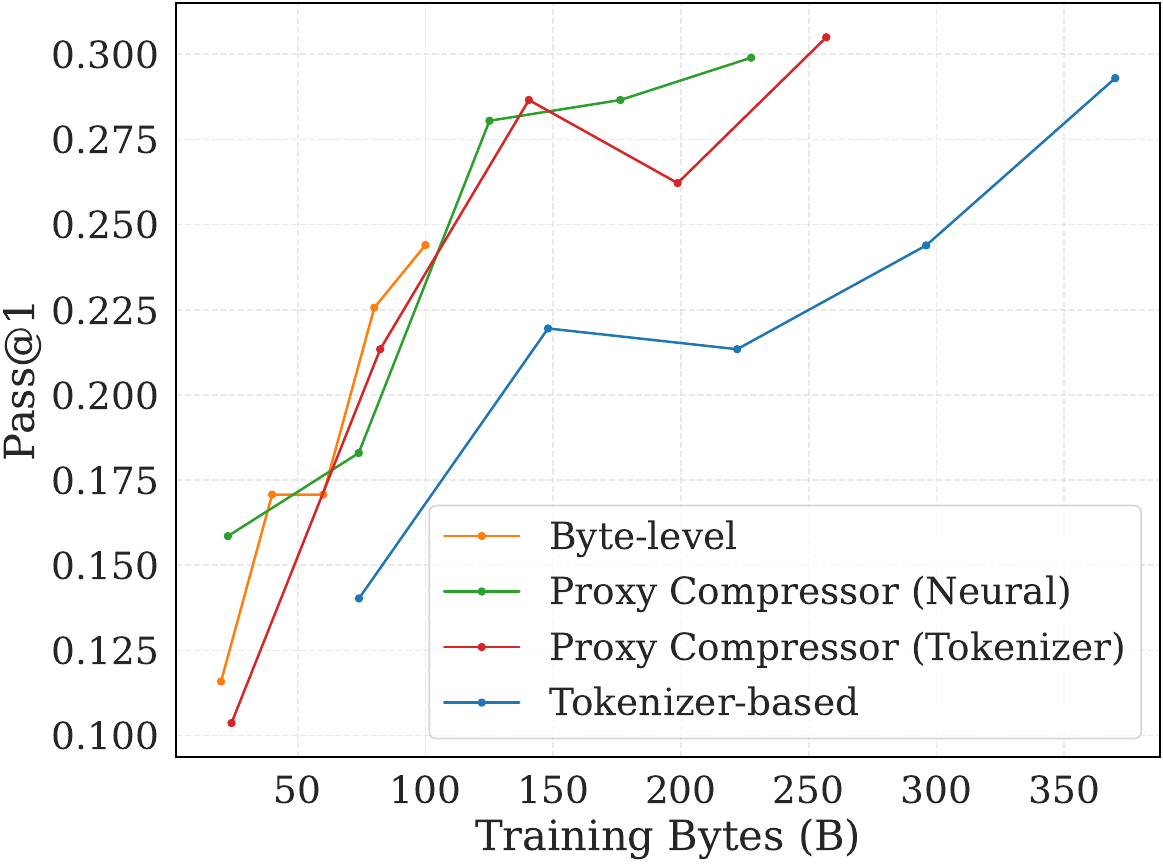}}
\end{subfigure}
\caption{Pass@1 performance on HumanEval-Plus for 14B models under different input representations, compared as a function of training FLOPs (left) and amount of training data (right).}
\label{fig:14B-humaneval_pass1-vs-flops-and-data}
\end{figure}

Under proxy compression, this trade-off shifts systematically with scale. At 0.5B-7B (\cref{fig:0B5-humaneval_pass1-vs-flops-and-data,fig:1B5-humaneval_pass1-vs-flops-and-data,fig:4B-humaneval_pass1-vs-flops-and-data,fig:7B-humaneval_pass1-vs-flops-and-data}), byte-level training remains more data-efficient, while proxy models generally lie between the two baselines and progressively narrow the gap. At 14B (\cref{fig:14B-humaneval_pass1-vs-flops-and-data}), proxy compression is able to capture the best of both regimes, performing comparably to tokenizer baselines under matched FLOPs and retaining byte-level data efficiency while substantially outperforming tokenizer baselines under matched data.

\paragraph{Longer Training Horizons.}
To test whether these trends hold at larger scale, we extend training to 320B symbols (tokens or bytes) on the full RefineCode GitHub split, following the OpenCoder setup~\citep{huang2024opencoder}. \cref{tab:downstream-transfer-longer-training} confirms consistent patterns: at 1.5B parameters, proxy models outperform byte baselines but still trail tokenizer baselines; at 7B, they close the gap and often match or exceed tokenizer-based models. This shows that proxy compression remains effective under longer training horizons.

Overall, these results show that proxy-compressed training enables efficient learning from compressed data, while achieving \emph{scalable downstream transfer} to raw-byte inference, with benefits that amplify at scale.

\begin{table}[t]
    \centering
    \caption{Downstream pass@1 on HumanEval-Plus and MBPP-Plus after training 320B sequence symbols on the full RefineCode GitHub data.}
    \label{tab:downstream-transfer-longer-training}
    \resizebox{0.95\columnwidth}{!}{
    \begin{tabular}{c l c c}
        \toprule
        \# Parameters & Model & HumanEval-Plus & MBPP-Plus \\
        \midrule
        \multirow{4}{*}{1.5B}
            & Tokenizer-based                        & \textbf{17.1} & \textbf{28.0} \\
            & Byte-level                             & 9.1  & 23.3 \\
            & Proxy (Neural)                         & 14.0 & 24.1 \\
            & Proxy (Tokenizer)                      & 12.8 & 25.1 \\
        \midrule
        \multirow{4}{*}{7B}
            & Tokenizer-based                         & 21.3  & \textbf{36.0} \\
            & Byte-level                              & 14.6  & 32.5 \\
            & Proxy (Neural)                          & 21.3  & \textbf{36.0} \\
            & Proxy (Tokenizer)                       & \textbf{22.0}  & 34.9 \\
        \bottomrule
    \end{tabular}}
\end{table}

\subsection{In-context Cross-Representation Transfer}
\label{experiments:in-context-transfer}

The results in \cref{experiments:main-transfer} demonstrate \emph{in-weight transfer}: knowledge learned from compressed inputs transfers to raw-byte inference through the model parameters.
Here, we probe a more explicit form of transfer called \emph{in-context translation}, where both compressed and raw views of the \emph{same} input appear in a single context, and the model must translate between representations on the fly. For each problem on HumanEval-Plus~\citep{chen2021humaneval,liu2023evalplus}, we take the raw prompt $\text{p}_{\text{raw}}$ and oracle solution $\text{s}_{\text{raw}}$, compress them into $\text{p}_{\text{comp}}$ and $\text{s}_{\text{comp}}$, and construct a mixed-representation prompt
\begin{align*}
    \big[ \boctok \circ \text{p}_{\text{comp}} \circ \text{s}_{\text{comp}} \circ \eoctok \circ \bortok \circ \text{p}_{\text{raw}} \big].
\end{align*}
The model must decode the corresponding raw solution $\text{s}_{\text{raw}}$ in raw bytes. We report \emph{oracle-translation pass@1}: whether the model recovers the correct solution given its compressed form in context. To study how in-context transfer depends on explicit pairing as described in \cref{method:overview}, we compare three training schedules (all with mixing ratio $r{=}0.9$): \emph{No pairs} (independent sampling without any paired data), \emph{Warmup-only} (enabling translation pairs for first 10k training steps only), and \emph{Always-on} (pairs throughout training). More details are in \cref{app:additional-exp-results:in-context-transfer}.

\cref{tab:in-context-transfer} reports both ordinary (i.e., no-oracle) and oracle-translation pass@1. Mixed-representation training induces substantial in-context transfer even without explicit pairs: under \emph{No pairs}, oracle-translation pass@1 reaches ${\sim}46\%$ (tokenizer) and ${\sim}33\%$ (neural), well above no-oracle baselines. With \emph{Always-on} pairing, both compressors rapidly achieve near-perfect translation ($>$95\% pass@1), demonstrating that even ambiguous neural compression can be reliably disambiguated in context. Under \emph{Warmup-only}, translation accuracy decays toward the \emph{No pairs} baseline once pairing is removed, yet still outperforming no-oracle baselines, indicating retained transfer capability.

Interestingly, high translation accuracy is not necessary for strong downstream performance (e.g., \cref{experiments:main-transfer}, which uses \emph{Warmup-only} by default). Comparing \emph{Warmup-only} and \emph{Always-on}, we observe a significant drop in oracle-translation accuracy, yet ordinary pass@1 is slightly better for \emph{Warmup-only}. Two confounded factors plausibly contribute: \emph{Always-on} duplicates each sample as a pair and thus sees fewer unique documents under matched FLOPs (\cref{app:compression-ratio-derivation}), and persistent co-presence of both views may encourage reliance on in-context translation, which is absent when only one view is available at inference. The performance of \emph{Warmup-only} is consistent with transfer operating beyond literal sequence-to-sequence translation: the model can still extract and leverage shared structure across representations \emph{in-weight} even without maintaining perfect in-context translation. Extended analyses are provided in \cref{app:additional-exp-results:in-context-transfer}.

\begin{table}[t]
\centering
\caption{Ordinary pass@1 (\%) at 50k training steps and oracle-translation pass@1 (\%) at 10k, 20k, 30k, 40k, and 50k training steps on HumanEval-Plus for different compressor types and pairing schedules (1.5B model; higher is better). We highlight cells with pass rates higher than 90\%.}
\label{tab:in-context-transfer}
\resizebox{0.99\columnwidth}{!}{
\begin{tabular}{l l c ccccc}
\toprule
\multirow{2}{*}{Compressor Type} & \multirow{2}{*}{Model Variant} & \multirow{2}{*}{\begin{tabular}{c}Pass@1\\50k steps\end{tabular}} & \multicolumn{5}{c}{Oracle Translation Pass@1} \\
& & & 10k & 20k & 30k & 40k & 50k \\
\midrule
\multirow{3}{*}{Tokenizer}
& No pairs & 17.0 & 7.30 & 29.3 & 40.2 & 44.5 & 45.7 \\
& Warmup-only & 20.7 & \textbf{90.9} & 31.1 & 42.1 & 53.7 & 45.7 \\
& Always-on & 17.0 & \textbf{93.3} & \textbf{95.7} & \textbf{96.3} & \textbf{96.3} & \textbf{96.3} \\
\midrule
\multirow{3}{*}{Neural}
& No pairs  & 14.6 & 0.6 & 3.0 & 7.9 & 20.7 & 32.9 \\
& Warmup-only & 18.9 & \textbf{90.9} & 39.0 & 14.6 & 22.0 & 38.4 \\
& Always-on & 14.6 & \textbf{94.5} & \textbf{92.1} & \textbf{93.9} & \textbf{95.1} & \textbf{95.1} \\
\bottomrule
\end{tabular}}
\end{table}

\subsection{What Makes a Good Proxy Compressor?}
\label{experiments:proxy-compression-comp}

\begin{figure}[t]
  \centering
  \includegraphics[width=0.98\linewidth]{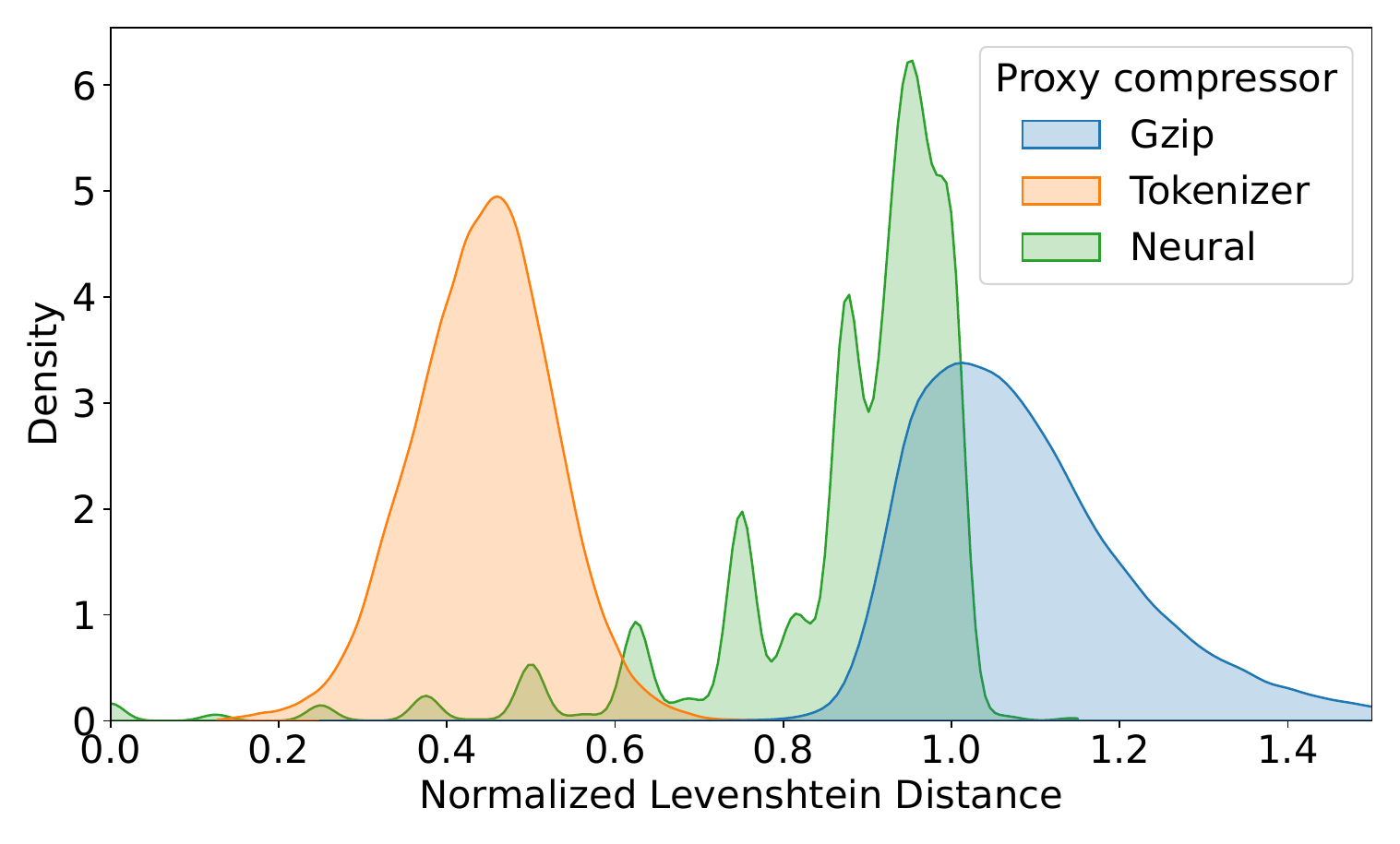}
  \caption{Compressor stability analysis under input perturbation: we apply random 10\% character deletion to 80K samples and measure normalized Levenshtein distance between compressed outputs before and after perturbation.}
  \label{fig:proxy-compression-comp}
\end{figure}

The results above show that tokenizer-based and neural proxies enable strong transfer. We now broaden the comparison to include gzip proxies and ask: what distinguishes effective proxy compressors from ineffective ones? We hypothesize that useful proxy representations should be structurally regular: small changes to the raw input should not cause disproportionate changes to the compressed sequence, and recurring syntactic or semantic patterns should induce recurring compressed patterns. We test this hypothesis by measuring compressor stability: we apply random 10\% character deletion to 80K samples drawn from training data and compute the normalized Levenshtein distance between compressed outputs before and after the perturbation. As shown in \cref{fig:proxy-compression-comp}, tokenization is highly stable, gzip is significantly more unstable (small edits cause large output changes), and neural compression lies in between.

\begin{figure}[tb]
\centering
{\includegraphics[width=0.92\columnwidth]{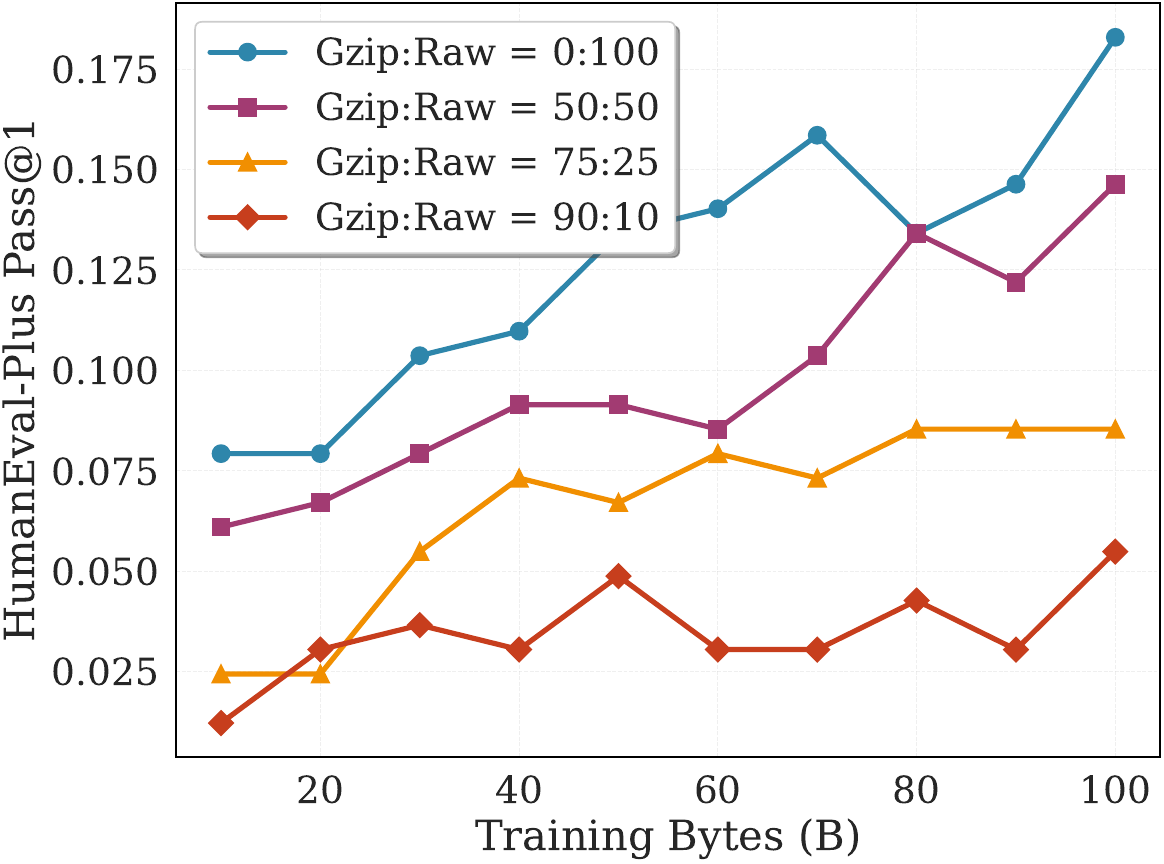}}
\caption{HumanEval-Plus pass@1 of 1.5B gzip-proxy models with different mixing ratios $r$ as a function of training data.}
\label{fig:gzip-performance}
\end{figure}

\paragraph{Gzip Proxies Fail to Transfer.}
\cref{fig:gzip-performance} shows downstream performance for 1.5B models trained with varying gzip-versus-raw ratios. Unlike tokenizer and neural proxies, increasing the proportion of gzip compressed data \emph{degrades} performance: models trained on pure raw bytes (0\% gzip) always outperform gzip-mixed variants. This indicates weak or even negative transfer from gzip-compressed sequences to raw bytes. We attribute this failure to the highly unstable and unstructured outputs from gzip (\cref{fig:proxy-compression-comp}), where small input perturbations cause drastic changes in the compressed sequence, making it difficult for the model to learn consistent patterns. In addition, gzip exploits low-level byte redundancies \emph{per sample} without respecting semantics, producing outputs that resemble noise to a language model. Empirically, we observe that models trained on gzip-compressed data fail to complete partial gzip sequences coherently, suggesting they never learn the underlying compression scheme.

\begin{figure}[t]
\centering
\includegraphics[width=0.92\columnwidth]{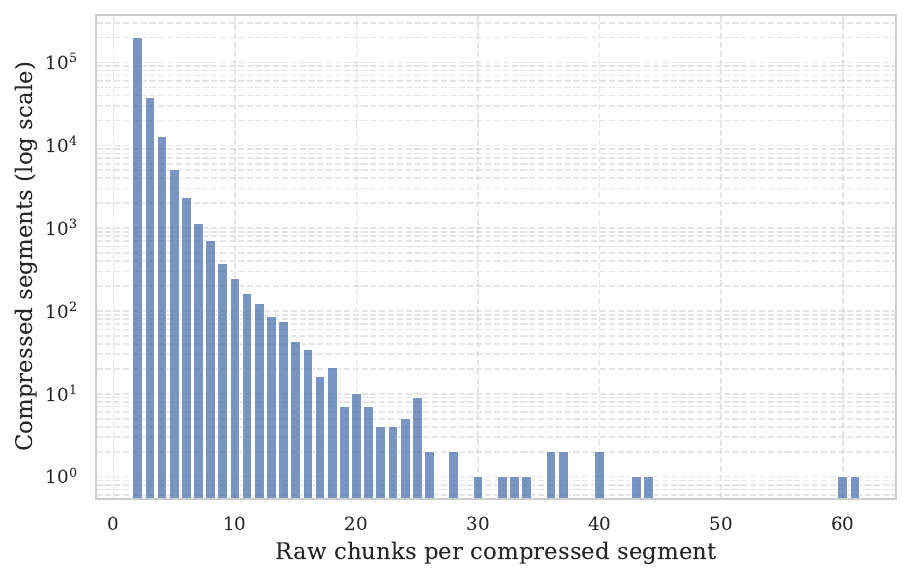}
\caption{Collision statistics for the neural compressor. The $x$-axis is the number of distinct byte chunks that collide on the same compressed segment, and the $y$-axis is the number of such compressed segments in log scale.}
\label{fig:collision-scale}
\end{figure}

\paragraph{Neural Compression: Structured Fuzziness.}
As discussed in \cref{method:neural-proxies}, neural compression is non-invertible: a single compressed sequence may correspond to multiple raw byte chunks (\emph{collisions}). Despite the ambiguity and unstable mapping (\cref{fig:proxy-compression-comp}), neural proxies still achieve strong transfer. We analyze this in \cref{fig:collision-scale}, showing that most compressed segments exhibit collisions. To quantify similarity among colliding chunks, we compute the \emph{longest common prefix (LCP) ratio}: the length of the shared prefix divided by average chunk length. Crucially, these collisions are highly structured: over 90\% of collisions have LCP ratios above 0.8, meaning most colliding chunks are nearly identical except for short suffixes (\cref{app:fig:lcp-ratio}), likely because the suffix entropies are too low for the arithmetic coder to allocate additional bits. This \emph{structured fuzziness} abstracts away formatting noise while preserving semantics, potentially explaining why neural proxies match tokenizer-based proxies despite their ambiguity. See \cref{app:additional-exp-results:neural-proxy-analysis} for a detailed analysis.

\subsection{Robustness Evaluation}
\label{experiments:robustness}
Byte-level models have been shown to be more robust to input perturbations than tokenizer-based models \citep{pagnoni2024blt,hwang2025hnet}. In this section, we investigate whether proxy-trained models inherit this advantage. We evaluate robustness on the HumanEval split of ReCode \citep{wang2023recode}, which applies semantics-preserving perturbations to coding problems in four families: function name rewrites (\emph{Function}), formatting changes (\emph{Format}), syntactic rewrites (\emph{Syntax}), and docstring paraphrases (\emph{Docstrings}). We report standard pass@1 on unperturbed inputs and \emph{Robust Pass@1} (RP), the worst-case pass@1 across $5$ perturbed variants per problem. Full metric definitions and per-family breakdowns are in \cref{app:additional-exp-results:robustness}.

\begin{figure}[tb]
\centering
\begin{subfigure}[b]{0.49\columnwidth} 
  \centering
  {\includegraphics[width=\textwidth]{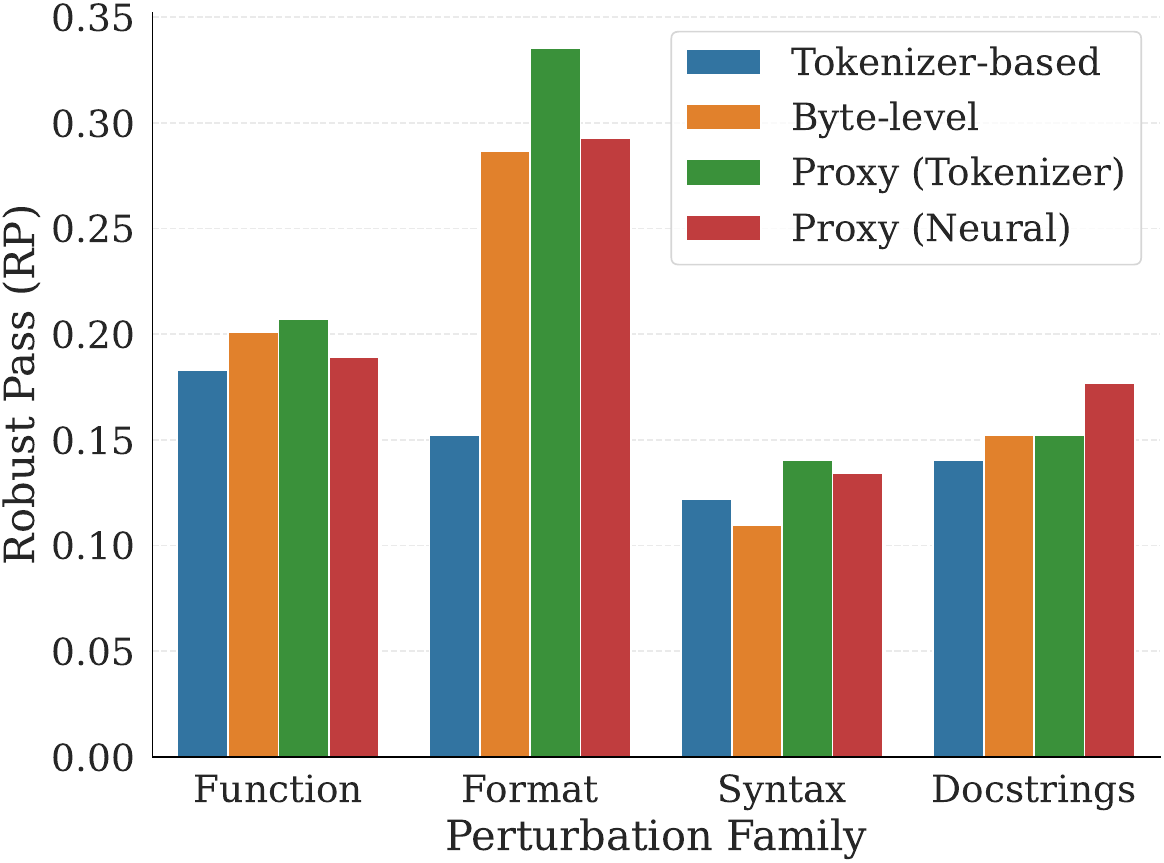}}
\end{subfigure}\hfill
\begin{subfigure}[b]{0.49\columnwidth} 
  \centering
  {\includegraphics[width=\textwidth]{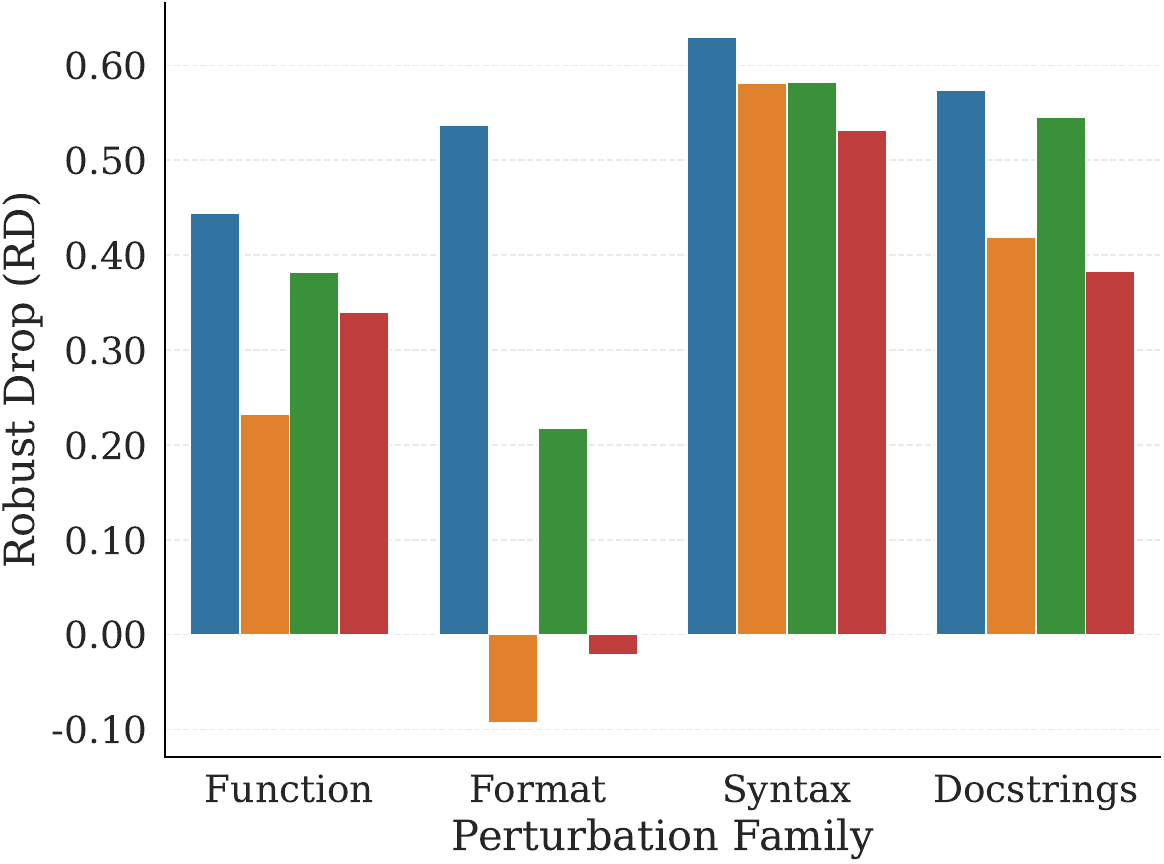}}
\end{subfigure}
\caption{Robust pass (left, higher is better) and robust drop (right, lower is better) performance on ReCode per perturbation family with different models.}
\label{fig:recode-per-family}
\end{figure}

\cref{tab:avg_robust} reports standard pass@1 and macro-averaged robust pass@1 ($\overline{\mathrm{RP}}$). Although the byte-level model achieves lower no-perturbation pass@1, it exhibits much better robustness than tokenizer baselines ($\overline{\mathrm{RP}}$: 18.7 vs.\ 14.9). Proxy-trained models inherit and amplify these robustness advantages: neural proxies attain the highest $\overline{\mathrm{RP}}$, while tokenizer proxies substantially improve robustness over the tokenizer baseline. One plausible explanation is that the fuzzy representations induced by neural compression (\cref{experiments:proxy-compression-comp}) help smooth over superficial noise, encouraging the model to focus on invariant structure during training.
\cref{fig:recode-per-family} reveals that \emph{Format} perturbations drive the largest gap: byte-level and proxy models remain near-nominal under whitespace and indentation changes, whereas the tokenizer baseline degrades sharply. On \emph{Syntax}, all models suffer a large robust drop (RD), though proxy models still improve RP. Overall, despite being trained predominantly on compressed inputs, our proxy-compressed models can mostly retain and in some cases improve upon the robustness of byte-level models; we further verify this with a targeted prompt-boundary study in \cref{app:additional-exp-results:prompt-boundary}.

\begin{table}[t]
\centering
\caption{Robustness evaluation on the HumanEval split of ReCode with 7B models. We report standard pass@1 and robust pass@1 ($\overline{\mathrm{RP}}$), the macro average across perturbation families.}
\label{tab:avg_robust}
\resizebox{0.95\columnwidth}{!}{%
\begin{tabular}{l c c}
\toprule
Model & Pass@1 & Robust Pass@1 ($\overline{\mathrm{RP}}$) $\uparrow$ \\
\midrule
Tokenizer-based        & \textbf{32.9} & 14.9  \\
Byte-level             & 26.2 & 18.7  \\
Proxy (Tokenizer)      & \textbf{32.9} & 19.1  \\
Proxy (Neural)         & 30.5 & \textbf{19.8}  \\
\bottomrule
\end{tabular}
}
\end{table}

\subsection{Analyses}
\label{experiments:analyses}

\begin{figure}[tb]
\centering
\begin{subfigure}[b]{0.49\columnwidth} 
  \centering
  {\includegraphics[width=\textwidth]{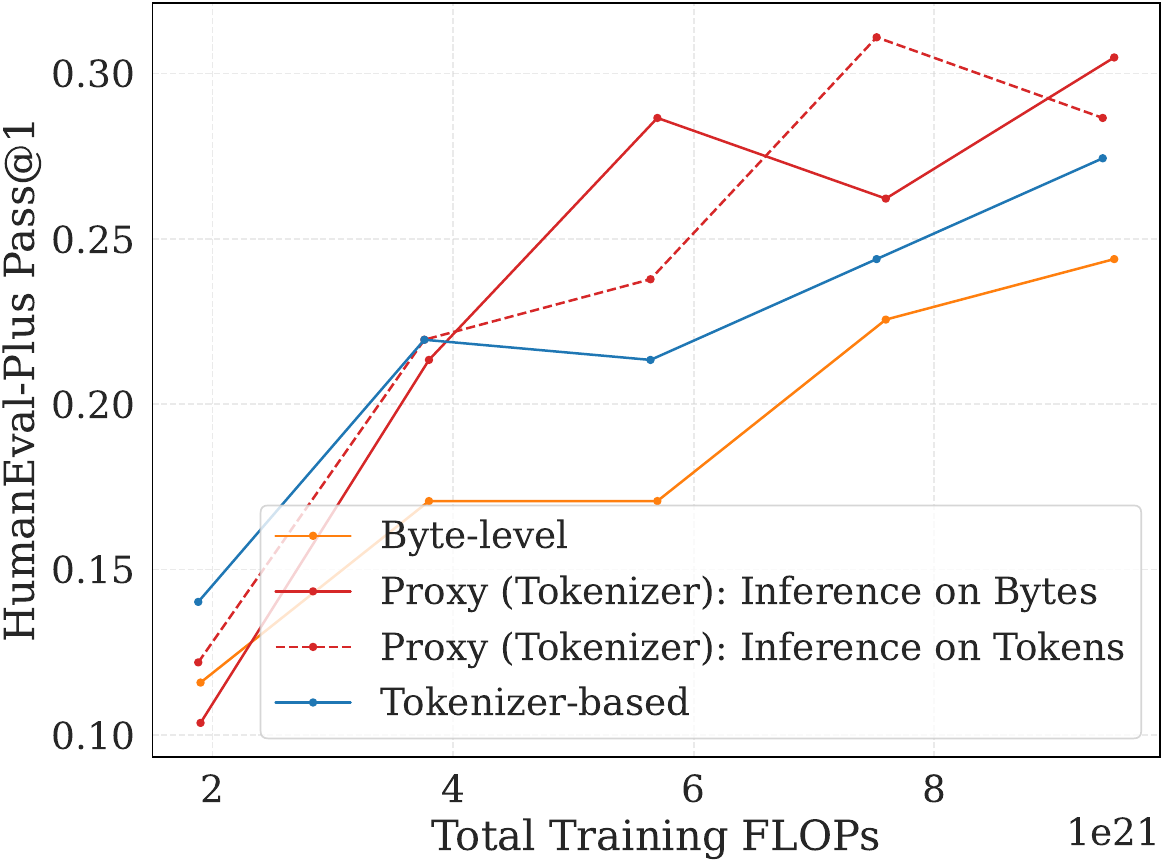}}
\end{subfigure}\hfill
\begin{subfigure}[b]{0.49\columnwidth} 
  \centering
  {\includegraphics[width=\textwidth]{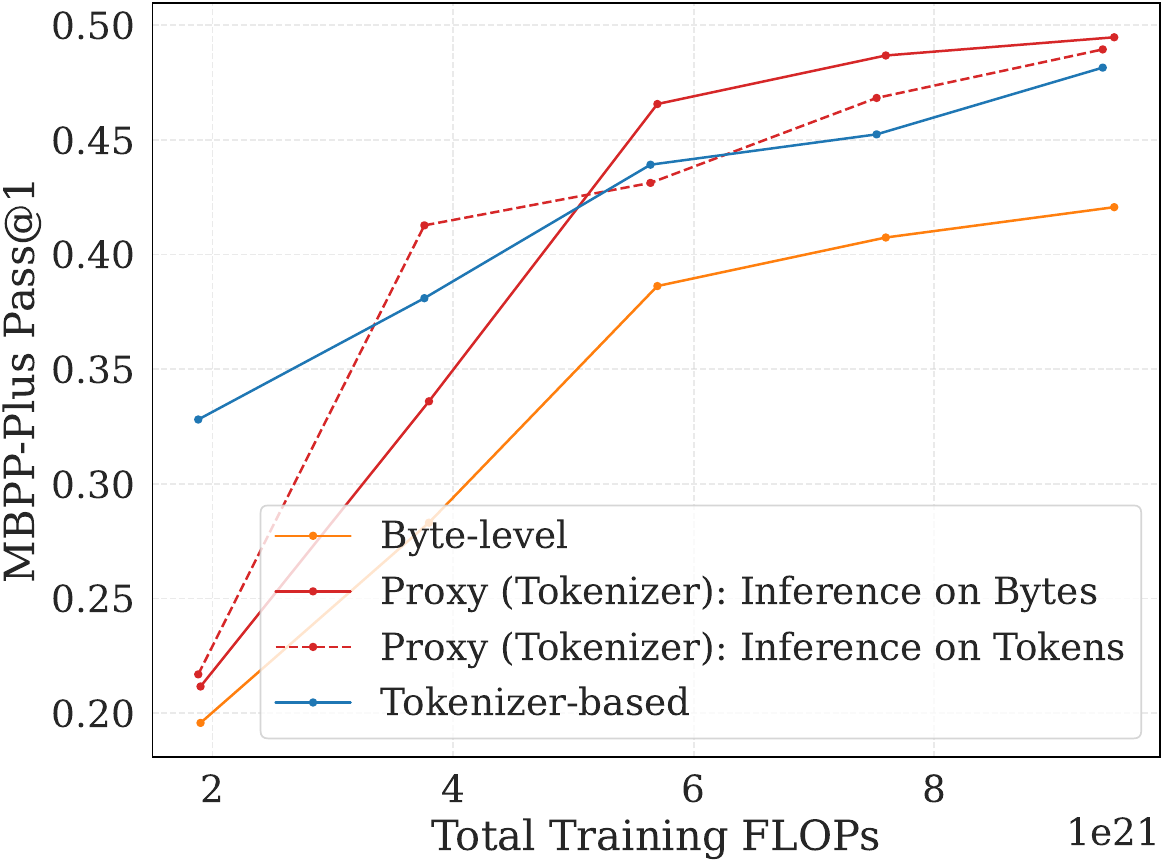}}
\end{subfigure}
\caption{Inference-time interface comparison on HumanEval-Plus (left) and MBPP-Plus (right) for 14B tokenizer-proxy models.}
\label{fig:inference_format}
\end{figure}

\paragraph{Inference-Time Interface.}
For tokenizer-based proxy compression, we can perform inference on either raw bytes or compressed tokens. As shown in \cref{fig:inference_format}, byte-level inference in many cases matches or outperforms token-level inference, though results vary across benchmarks and compute budgets. This is notable because only $10\%$ of training samples are presented in raw-byte form. We attribute this to two factors: (i) cross-representation transfer is sufficiently strong that the model performs well on raw bytes even with limited exposure, and (ii) longer byte sequences afford more test-time compute per problem instance. For neural compression, we report only raw-byte inference: its compressed symbols are training-time proxies and are not directly decodable into text, so they do not define a usable inference interface (\cref{method:neural-proxies}).

\begin{figure}[tb]
\centering
{\includegraphics[width=0.94\columnwidth]{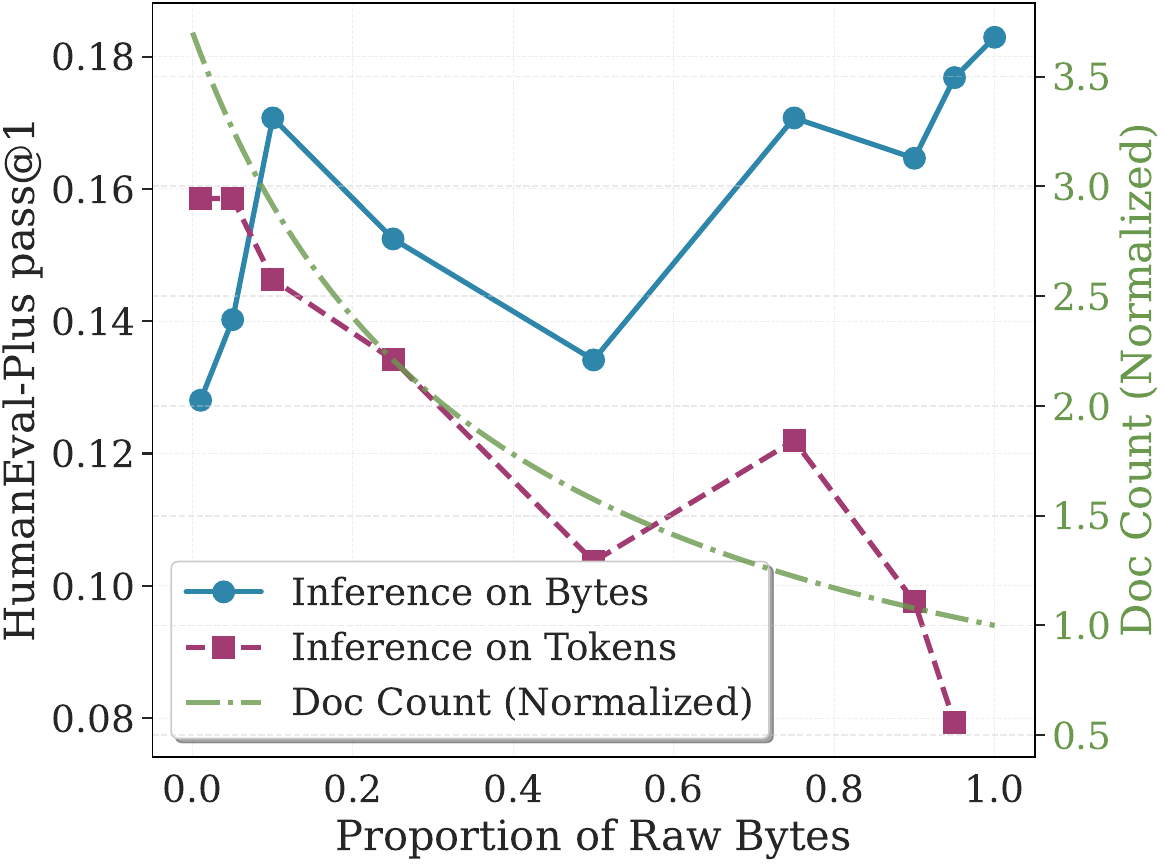}}
\caption{Performance on HumanEval-Plus Pass@1 as a function of raw byte proportion in the training mixture for 1.5B models with tokenizer-based proxy compression. Solid and dashed lines show models performing byte-level and token-level inference, respectively. The dash-dot line (right y-axis) indicates the normalized document count seen during training within a fixed training FLOPs budget.}
\label{fig:ratio-ablation}
\end{figure}

\paragraph{Mixing Ratio Ablation.}
Throughout our experiments, we fix $r=0.9$ as the proportion of compressed samples in the training corpus. To study the effect of raw-compressed data composition in proxy compression training, we train 1.5B parameter models on mixtures containing varying proportions of raw and compressed data, maintaining a constant budget of 50000 training steps. \cref{fig:ratio-ablation} reveals that byte-level inference performance does not degrade monotonically as the proportion of compressed samples increases. This pattern is explained by the document count curve (dash-dot line), which shows that reducing the raw-byte proportion in the mixture increases the number of unique training samples seen within a fixed compute budget. At $r{=}0.9$, models observe approximately $3\times$ more samples than at $r{=}0.0$ (100\% raw bytes), and strong compressed-to-raw transfer allows them to benefit from this additional data. The mixing ratio ablation also reveals \emph{an asymmetry in transfer direction}. For byte-level inference (solid line in \cref{fig:ratio-ablation}), performance remains strong even with only 10\% raw data, indicating robust transfer from compressed to raw representations. In contrast, for token-level inference (dashed line), performance degrades nearly monotonically as the raw-byte proportion increases, suggesting weak transfer from raw back to compressed representations.

\section{Related Work}
\label{related_work}
\paragraph{Input Representations for Language Models.}
Most modern language models rely on external tokenization that segments textual data into sequences of discrete tokens, such as Byte-Pair Encoding \citep{gage1994bpe,sennrich2016bpe}, WordPiece \citep{schuster2012wordpiece,wu2016googlenmt,devlin-etal-2019-bert}, UnigramLM \citep{kudo2018subwordregularization}, and SentencePiece implementations \citep{kudo2018sentencepiece}. Tokenization compresses raw text into shorter token sequences, and recent work has pushed this design along multiple axes, including analyzing the overall tokenization pipeline design \citep{bostrom2020bpe,zouhar2023tokenization,schmidt2024tokenization,dagan2024getting}, studying non-canonical tokenization behavior \citep{cao2021evaluate,geh2024signal,geh2025adversarial,zheng2025brokentokens}, exploring vocabulary scaling \citep{tao2024scaling,yu2025scone,huang2025overtokenized}, and improving compression \citep{fried2023incoder,gee2023mwt,schmidt2025boundless,liu2025superbpe}.

Besides tokenization, recent work also explores distinct approaches to constructing compressed representations, such as gzip-based compressors \citep{jiang2023zipclassification,lester2024training}, arithmetic coding with equal-information windows \citep{lester2024training}, concept-level semantic units \citep{barrault2024lcm,qu2025dlcm}, morphology-driven byte representations \citep{limisiewicz2024myte} that improve efficiency and fairness in multilingual settings, and pixel-rendered text \citep{salesky2021robust,rust2023pixel,salesky2023multilingualpixel,li2023glyphdiffusion,lee23pix2struct,lotz2023textrendering,tai2024pixar,gao2024improving,wei2025deepseekocr,cheng2025glyph}.

\paragraph{Byte-level Models.}
Alternatively, one can directly use byte-level sequences as the input representation \citep{sutskever2011generating,graves2013generating,radford2017learning,chung2017hierarchical,hwang2017character,alrfou2019character,choe2019bridging,xue2022byt5,wang2024mambabyte,neitemeier2025hierarchical,zheng2025evabyte,minixhofer2025bolmo}. To mitigate the computational cost of longer sequences, recent work aims to pool input sequences into shorter representations inside the model architecture, including downsampling the input in a context-independent manner \citep{jaegle2021perceiver,jaegle2021perceiverio,hawthorne2022perceiverar,clark2022canine,nawrot2022hourglass,yu2023megabyte,pagnoni2024blt,slagle2024spacebyte,videau2025aunets}. Recent work also explores adaptive compression with dynamically adjusted granularity based on the input content \citep{tay2022charformer,godey2022manta,nawrot2023dynamicpool,kallini2024mrt5,ahia2024magnet,owodunni2025flexitokens,geng2025zip2zip,hwang2025hnet,minixhofer2025bolmo}. Beyond text-only language modeling, byte-level models are also used for multi-modal learning \citep{egli2025multiscale,zheng2025evabyte}. More generally, byte-level models can be used to process various types of digital file contents \citep{park2023rgbnomore,wu2024bgpt,perez2024jpegfiles,han2024jpeglm,horton2024bytes,jolicoeur2025bytecraft,li2025bytegen,alcazar2025tempest}.

Our work differs from prior approaches by leveraging compressed representations as a proxy during training while still retaining a byte-level interface at inference, without architectural modifications.

\section{Conclusion}
\label{conclusion}
We introduced proxy compression, a mixed-representation training scheme that decouples the efficiency benefits of compressed training from the inference-time interface. By jointly training on raw inputs and compressed views produced by external compressors, models learn to align representations and transfer effectively from compressed inputs to raw-byte inference. Extensive experiments on code language modeling demonstrate that proxy-trained models substantially outperform pure byte-level baselines under fixed compute budgets, and at scale match or surpass tokenizer-based models, all while operating solely on raw bytes. Our systematic study of proxy compressors further reveals that tokenizer-based and neural proxies support strong transfer, whereas generic compression (gzip) fails to provide useful training signal, suggesting that structured, semantically meaningful compression is key to effective proxy training.

\vspace{-2pt}\paragraph{Limitations.}
Our evaluation primarily focuses on code language modeling; our natural-language results provide initial evidence beyond code, but larger-scale natural-language, multilingual, and mixed-domain validation remains future work. Because inference is performed on raw bytes, deployment efficiency depends on the underlying byte-level model architecture; EvaByte \citep{zheng2025evabyte} mitigates this cost in our experiments, but proxy compression itself does not remove the general challenge of efficient byte-level decoding. The trade-offs among compression rate, transfer strength, and compute efficiency are also not yet fully characterized (e.g., more aggressive compression may amplify efficiency gains but could degrade transfer quality). Finally, incorporating proxy compression directly into model architecture design, rather than treating it purely as a data preprocessing step, may unlock further performance or efficiency improvements.

\section*{Impact Statement}
This paper presents work whose goal is to advance the field of Machine Learning. There are many potential societal consequences of our work, none of which we feel must be specifically highlighted here.

\bibliography{proxy_paper}
\bibliographystyle{icml2026}

\newpage
\appendix
\onecolumn
\textbf{\huge Appendix}
\vspace{0.05in}

\section{Token-level Compression Ratios for Proxy Compression}
\label{app:compression-ratio-derivation}
We implement mixed-representation training at the sample level. This indicates that although documents will be compressed with a certain probability $r$, the actual ratio of compressed tokens present in the context will differ from the sample-level mixing rate. In this section, we provide a rough estimate of the \emph{token-level} compression ratio.

For simplicity, we consider a fixed batch size per training step, that is, the total number of symbols is fixed, denoted by $M$. For raw byte models, we assume each batch consists of $N_b$ input documents each with average sample length $L_b$; for a compression-based model, this comprises $N_t$ input samples with average length $L_t$. We also denote the compression rate brought by compression as $C = \frac{L_b}{L_t}$. As a result, we have
\begin{align*}
    M = N_t L_t = N_b L_b = N_b C L_t \rightarrow N_t = C N_b,
\end{align*}
which indicates that compression-based models will approximately consume $C\times$ more samples than raw byte models. Now consider our proxy compression framework, where we additionally maintain a mixing ratio $r$ such that with probability $r$ an input sample will be compressed with a compressor at rate $C$. Denoting the total number of input samples (either compressed or not) per training step as $N'$, we have
\begin{align*}
    M = rN'L_t + (1-r)N'L_b = rN'\frac{L_b}{C} + (1-r)N'L_b = \left[\frac{r}{C} + (1-r)\right]N'L_b,
\end{align*}
and by comparing the above, we obtain
\begin{align*}
    N' = \frac{1}{\frac{r}{C} + (1-r)}N_b,
\end{align*}
which roughly estimates the number of input samples consumed in our compressed-raw mixture. For instance, for tokenizer-based proxies at a rate $r = 0.9$ with the OpenCoder tokenizer \citep{huang2024opencoder}, which delivers a compression rate $C \approx 3.7$, the actual token-level compression rate would be $\frac{1}{\frac{r}{C} + (1-r)} \approx 2.91$.

\paragraph{Mixing with In-context Translation Pairing.}
When we employ translation pairs into training (\cref{method:overview}), the actual compression rate is further decreased, as the same input sample will be duplicated twice. Since for each translation pair, it increments the amount of both compressed and raw data, we calibrate the mixing rate to take this into account. In particular, with the mixing rate $r$, we assume there will be $rN$ samples compressed, while $(1-r)N$ samples including both the raw and compressed versions. So the actual compression and raw samples would be $r+1-r:1-r$. 

If $r > 0.5$, we mix translation pairs with compressed samples; conversely, if $r \leq 0.5$, we mix translation pairs with raw samples, as it would be impossible to attain a compressed-raw fraction larger than $0.5$ if we solely mix translation and raw bytes. Concretely, if we desire a compressed fraction $r$, then the calibrated mixing rate $r'$ of mixing translation pairs with probability $r'$ must satisfy
\begin{align*}
\begin{cases}
    \frac{r'+(1-r')}{r'} = \frac{r}{1-r}, &\text{ if } r > 0.5, \\
    \frac{r'}{r'+(1-r')} = \frac{r}{1-r}, &\text{ if } r \leq 0.5.
\end{cases}
\end{align*}
This gives
\begin{align*}
    r' = \begin{cases}
        \frac{1}{r} - 1, &r>0.5,\\
        \frac{r}{1-r}, &r\leq0.5.
    \end{cases}
\end{align*}
As a result, we can draw training examples with translation pairs appearing with probability $r'$, mixed with compressed (resp. raw) samples, where the effective fraction of compressed versus raw samples matches the intended target $r > 0.5$ (resp. $r \leq 0.5$).

We now investigate the delivered token-level compression rate with translation pairs. Assume $r > 0.5$ which implies $r' = \frac{1}{r} - 1$. Then for each training batch, we have
\begin{align*}
    M = (1-r')N'L_t + r'N'(L_b+L_t) = (1-r')N'\frac{L_b}{C} + r'N'(L_b+\frac{L_b}{C}) = \left[\frac{1}{C} + r'\right]N'L_b.
\end{align*}
This gives
\begin{align*}
    N' = \frac{1}{\frac{1}{C} + \frac{1}{r} - 1}N_b,
\end{align*}
with the configuration above, this yields an actual compression rate of $N'\approx 2.62N_b$, which is a bit slower than that without translation. As a result, we only employ translation pairing as a warmup phase at the beginning of training and disable it later to maximize the amount of unique data per training step.

\paragraph{In-context Translation Pairing Warmup.}
Also notice that when warmup phases are incorporated, we consider annealing $r(i)$ linearly from the start point $a$ and ending rate $b$ with $T$ steps (in this work, $a = 0.4$ and $b = 0.9$ by default).
For $r \leq 0.5$, we can follow similar reasoning and derive
\begin{align*}
    M = (1-r'(i))N_i'L_b + r'(i)N_i'(L_b+L_t) = (1-r'(i))N_i'L_b + r'(i)N_i'(L_b+\frac{L_b}{C}) = \left[1 + \frac{r'(i)}{C}\right]N_i'L_b.
\end{align*}
This gives
\begin{align*}
    N_i' = \frac{1}{1 + \frac{r'(i)}{C}}N_b = \frac{1}{1 + \frac{r(i)}{1-r(i)}\frac{1}{C}}N_b.
\end{align*}
The compression rate for $r > 0.5$ can be calculated similarly. We evaluate the average number of samples processed at each step as $\overline{N'} \coloneqq \frac{1}{T}\sum_{i=1}^T N_i'$. This can be evaluated either by enumerating each term in the summation, or by approximating with the piecewise integral average, which can be evaluated in closed form to give a quick estimate. For instance, in our default schedule, we anneal $r(i) = 0.4 + \frac{0.5i}{10000}$ to increase from $0.4$ to $0.9$ at the first 10000 training steps $i$. This gives the average rate $\overline{N'} \approx 1.38N_b$, in other words, during the warmup phase, we only process on average $1.38\times$ as many samples per step as a pure raw byte model training step.

\section{Implementation Details of Proxy Compressors}
\label{app:impl-details}

\subsection{Tokenizer-based Proxy Compression}
\label{app:impl-details:tokenizer}
This section provides additional implementation details on tokenizer-based proxy compression, including representation formats, encoding strategies, and pre-tokenization schemes. \cref{tab:ablation_token_compressor_design} summarizes ablation results across these design choices.

\paragraph{Representation Formats.}
We consider two formats for representing tokenizer outputs:
\begin{itemize}[leftmargin=*, itemsep=2pt, topsep=2pt]
    \item \textbf{Token indices} (default): Each token ID is mapped to a dedicated entry in the language model vocabulary and treated as a single symbol. This achieves the highest compression rate ($3.7\times$ with the OpenCoder tokenizer) and yields the best downstream performance.
    \item \textbf{Token bytes}: Token IDs are serialized as fixed-length byte sequences. We choose the smallest $B$ such that $256^B \geq V$, where $V$ is the tokenizer vocabulary size. For OpenCoder ($V = 96{,}640$), this gives $B = 3$, so each token becomes 3 bytes. Under this scheme, a length-$L$ token sequence produced by the tokenizer is converted into a sequence of $B L$ symbols. This format keeps both raw and compressed representations in byte space, which we hypothesized might ease cross-representation alignment. However, it reduces the effective compression rate and does not improve transfer (\cref{tab:ablation_token_compressor_design}).
\end{itemize}
A practical advantage of token bytes is flexibility: since the language model vocabulary is decoupled from the tokenizer vocabulary, arbitrarily large tokenizers (including ``superword'' vocabularies that merge across word boundaries \citep{liu2025superbpe}) can be supported without changing the model's embedding table. For our token-byte experiments, we train a BPE tokenizer with a default vocabulary size $V = 65{,}536$ on our pretraining corpus, so that each token ID can be represented as exactly 2 bytes.

We also explore a \emph{double-byte} variant that expands the language model vocabulary from 256 to $65{,}536$. Under this scheme, each token ID maps to a single symbol, rather than being serialized into multiple bytes. As shown in \cref{tab:ablation_token_compressor_design}, double-byte achieves a higher compression rate ($3.5\times$) than simple bytes ($1.7\times$) but still underperforms direct token-index representations.

\paragraph{Byte Encoding Strategies.}
For the token-bytes format, we explored several encoding strategies:
\begin{itemize}[leftmargin=*, itemsep=2pt, topsep=2pt]
    \item \textbf{Default (fixed-length)}: Each token ID is converted to a $B$-byte big-endian representation. All tokens occupy the same number of bytes, providing uniform representation complexity. Note that this implicitly encodes frequency information, since BPE typically assigns lower IDs to more frequent tokens.
    \item \textbf{Huffman coding}: Variable-length byte codes are assigned based on token frequency, with more frequent tokens receiving shorter codes. Codes are byte-aligned and satisfy the Kraft-McMillan inequality for prefix-free decodability. In practice, this yields only marginal compression improvements over fixed-length encoding.
    \item \textbf{Gray coding}: Tokens are first sorted lexicographically by their UTF-8 surface forms and assigned sequential ranks. These ranks are then converted to Gray codes, ensuring that lexicographically similar tokens differ by only one bit in their byte representations. This preserves locality in byte surfaces and slightly improves performance.
\end{itemize}
As shown in \cref{tab:ablation_token_compressor_design}, these encoding strategies yield marginal improvements in downstream performance within the token-bytes format, and all underperform direct token-index representations.

\paragraph{Pretokenization Schemes.}
Standard BPE pipelines first segment raw text into ``words'' via pre-tokenization (typically regex-based), then apply iterative merges to produce subwords. We experimented with several pre-tokenization schemes:
\begin{itemize}[leftmargin=*, itemsep=2pt, topsep=2pt]
    \item \textbf{Default}: Standard regex-based pre-tokenization~\citep{radford2019gpt2,openai2023gpt4,dagan2024getting}.
    \item \textbf{Line-separated}: Pre-tokenization allows tokens to span multiple words within a line~\citep{fried2023incoder}.
    \item \textbf{SuperBPE}: Extended pre-tokenization that permits merges across whitespace boundaries~\citep{liu2025superbpe}.
\end{itemize}
Line-separated and SuperBPE pre-tokenization improve compression rates ($2.3\times$ and $2.9\times$ respectively, compared to $1.7\times$ for default), but do not match the performance of direct token-index representations.

\paragraph{Tokenizer Algorithms.}
We also compared Unigram-based tokenization~\citep{kudo2018subwordregularization} and found no significant difference in transfer performance at matched compression rates, suggesting that the choice of tokenization algorithms is less critical.

\begin{table}[t]
\centering
\caption{Ablation study on design choices of tokenizer proxy compression. All models are 1.5B parameters trained on 100B sequence symbols from the Python subset. Direct token-index representation achieves the best compression-performance trade-off.}
\label{tab:ablation_token_compressor_design}
\begin{tabular}{lccccc}
\toprule
Format & Vocab Size & Pre-tokenization & Encoding & Compression Rate & Pass@1 (\%) \\
\midrule
Byte         & 256   & Default  & Default        & 1.7 & 14.6 \\
Byte         & 256   & Default  & Gray Coding    & 1.7 & 15.2 \\
Byte         & 256   & Default  & Huffman Coding & 1.8 & 18.3 \\
Double-byte  & 65536 & Default  & Default        & 3.5 & 15.2 \\
Byte         & 256   & Line     & Default        & 2.3 & 18.3 \\
Byte         & 256   & SuperBPE & Default        & 2.9 & 16.5 \\
\midrule
Token-index  & 96640 & Default  & ---            & 3.7 & 20.7 \\
\bottomrule
\end{tabular}
\end{table}

\subsection{Neural Proxy Compression}
\label{app:impl-details:neural}

This section provides implementation details for neural proxy compression, including the compressor architecture, entropy-based segmentation algorithm, and parallelization strategies. \cref{tab:ablation_neural_compressor_design} summarizes ablation results across key design choices.

\paragraph{Compressor Model.}
Our neural proxy compressor combines a small byte-level language model with arithmetic coding. We train a $\sim$40M-parameter byte-level Transformer on our pretraining corpus, with the dimension size 512, 12 layers, 8 attention heads. Training uses a learning rate of $1\text{e}\!-\!3$ for 200k steps with batch size $64 \times 2048$ bytes. The model operates over a fixed alphabet of size $256$, and produces conditional distributions $p(b_t \mid b_{<t})$ for every byte in the input $x_{\text{raw}} = (b_1, \dots, b_T)$. These distributions drive the arithmetic coder to output a compressed bitstream.

\paragraph{Entropy-based Segmentation.}
Naively compressing training data is prohibitively slow: (1) the arithmetic coder is inherently sequential, and (2) the equal-information window scheme in \citet{lester2024training} requires frequent termination and context truncation, triggering new forward passes for the language model to obtain new probabilities. To increase parallelism, we pre-segment training samples and compress each segment independently.

We implement an \emph{entropy-based adaptive segmentation} strategy: for each input example in UTF-8 bytes, we run the model forward pass to obtain per-position logits and compute the per-byte cross-entropy $h_t \coloneqq -\log p_t(b_t)$ for $t = 1,\dots,L$. Segment boundaries are constructed where the entropy profile indicates low predictability based on two criteria:
\begin{enumerate}[leftmargin=*, itemsep=2pt, topsep=2pt]
    \item \textbf{Absolute entropy}: positions where $h_t$ exceeds a global threshold (e.g., 95th percentile).
    \item \textbf{Entropy jumps}: positions where the finite difference $\Delta h_t = h_t - h_{t-1}$ exceeds a monotonicity threshold, indicating sudden changes in predictability.
\end{enumerate}
Similar entropy-based criteria also appear in BLTs \citep{pagnoni2024blt} for dynamic byte patchification within the model architecture; here we use them only for segmenting inputs for parallel arithmetic coding. By default, we use the 95th percentile as the threshold for both criteria.

\paragraph{Vectorized Arithmetic Coding.}
We implement a batched arithmetic coder to process multiple segments in parallel with equal-information windows \citep{lester2024training}. Given a predefined bit threshold $\tau$, compression proceeds iteratively for each segment:
\begin{enumerate}[leftmargin=*, itemsep=2pt, topsep=2pt]
    \item Run a forward pass of the compressor model to obtain next-byte distributions (on GPU).
    \item Perform arithmetic coding and count output bits in the resulting compressed bitstream (on CPU). If the bitstream for the current window exceeds $\tau$ bits, emit the first $\tau$ bits, discard the consumed byte context, and return to step 1 with the truncated context.
\end{enumerate}
We design a pipelined implementation to overlap GPU forward passes with CPU encoding across iterations.

\paragraph{Caching.}
We exploit substantial repetition in pre-training corpus with a global segment cache: before compressing a segment, we check whether it exists in the cache. Cache hits will retrieve the pre-computed compressed bitstream, while cache misses will trigger compression as usual. Caching significantly reduces redundant computation during arithmetic coding.

\paragraph{Bitstream Packing.}
After compressing each segment to a bitstream, we pack fixed-length bit chunks into discrete symbols. We define a bitwidth $N$ and pack consecutive $N$-bit chunks from the bitstream into symbols. By default, $N = 16$, which yields a vocabulary size $V = 65{,}536$ for the language model embedding table; we ablate this design choice in \cref{tab:ablation_neural_compressor_design}.

\paragraph{End-to-End Pipeline.}
The compressor runs fully offline in parallel via a multi-process pipeline. Each worker:
\begin{enumerate}[leftmargin=*, itemsep=2pt, topsep=2pt]
    \item Reads a shard of the corpus.
    \item Applies entropy-based segmentation (on GPUs).
    \item Compresses segments with arithmetic coding, equal-info windows \citep{lester2024training}, and cache lookup (GPU/CPU pipelining).
    \item Packs the resulting compressed bitstream into fixed-bit symbols.
    \item Writes segmentation metadata and compressed sequences.
\end{enumerate}
At training time, the proxy compressor simply reads the pre-computed compressed data and presents them to the mixed-representation training pipeline (\cref{method:overview}). Our pipeline design improves efficiency significantly: we process ${\sim}3.3$TB of pretraining data at 0.57 GB/hour per process, compared to 0.005 GB/hour for a naive fully-sequential implementation.

\paragraph{Why Neural Compression Cannot Decode.}
Unlike tokenization, neural compression is not invertible and thus does not support decoding. Several factors contribute to this: (i) probability tables are not stored during compression, (ii) our use of segmentation and equal-information windowing \citep{lester2024training} produces compressed segments of varying byte lengths, and these boundaries are unknown at decode time, and (iii) at inference time, when new compressed symbols are predicted by the language model instead of produced by the compressor, the underlying compressor probabilities that would be needed for decoding are by definition unavailable. These factors cause the same compressed sequence to correspond to multiple plausible raw byte sequences, or the ``fuzziness'' discussed in \cref{method:neural-proxies}.

\paragraph{Design Choice Ablations.}
\cref{tab:ablation_neural_compressor_design} ablates key design choices in our neural compression: segmentation strategy, equal-information bit window size, and symbol packing granularity. For \emph{segmentation}, we observe fixed-length segmentation performs poorly, likely due to abrupt changes with semantic-agnostic boundaries. Line-based segmentation (splitting at newlines) works well for code, improves results, but remains unsatisfactory. Entropy-based segmentation is general and yields the best performance by placing boundaries at natural transition points. When using equal-info windows \citep{lester2024training}, it improves performance but degrades compression rates, as window resets introduce overhead in terms of compression. We find 16-bit windows strike the best balance and increasing to 32 bits improves compression rate but greatly degrades performance. For packing, using 16-bit per symbol substantially outperforms the 8-bit variant. In addition, the choice between 95th and 98th percentile for the threshold in entropy segmentation has minimal impact on downstream performance. Overall, entropy-based segmentation with 16-bit equal-information windows and 16-bit symbol packing yields the best trade-off.

\begin{table}[t]
\centering
\caption{Ablation study on design choices of neural proxy compression with 1.5B models. Compressor CR denotes the compression rate of the proxy compressor in isolation (i.e., without mixing with raw bytes). Pre-segmentation controls how documents are split before arithmetic coding; EqualInfoAC Bits specifies the target information per coding window~\citep{lester2024training}; Packing Bits determines how many bits are grouped into each discrete symbol ($V$ is the resulting vocabulary size). The bottom row shows our final configuration, which achieves an effective compression rate of approximately $2.6$ under the mixed-representation training scheme with $r{=}0.9$ (see~\cref{app:compression-ratio-derivation}).}
\label{tab:ablation_neural_compressor_design}
\begin{tabular}{ccccc}
\toprule
Pre-segmentation & EqualInfoAC Bits & Packing Bits & Compressor CR & Pass@1 (\%) \\
\midrule
Fixed-length (size=4)  & None & 8 ($V\!=\!256$)    & 1.4 & 8.5 \\
Fixed-length (size=8)  & None & 8 ($V\!=\!256$)    & 2.0 & 3.7 \\
Fixed-length (size=16) & None & 8 ($V\!=\!256$)    & 2.9 & 6.1 \\
Lines                  & None & 8 ($V\!=\!256$)    & 5.1 & 9.8 \\
Lines                  & 16   & 8 ($V\!=\!256$)    & 2.3 & 11.6 \\
Entropy (98th percentile) & 16   & 8 ($V\!=\!256$)    & 1.8 & 13.4 \\
Entropy (98th percentile) & 16   & 16 ($V\!=\!65536$) & 3.5 & 18.9 \\
Entropy (95th percentile) & 32   & 16 ($V\!=\!65536$) & 4.1 & 12.2 \\
\midrule
Entropy (95th percentile) & 16   & 16 ($V\!=\!65536$) & 3.1 & 18.3 \\
\bottomrule
\end{tabular}
\end{table}

\section{Additional Architectural, Training, and Evaluation Details}
\label{app:additional-details}

\subsection{Model Architectures}
\label{app:additional-details:arch}

\begin{figure}[tb]
\centering
\begin{subfigure}[b]{0.49\columnwidth} 
  \centering
  {\includegraphics[width=\textwidth]{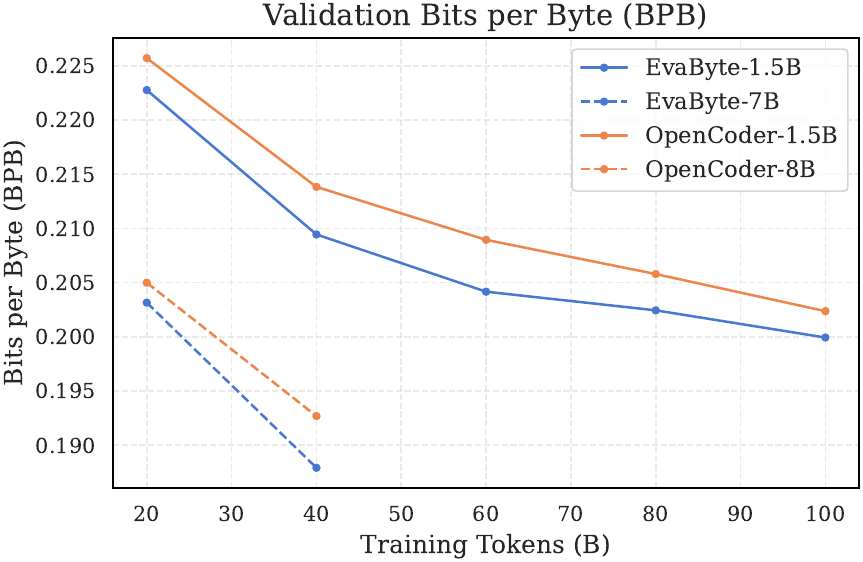}}
\end{subfigure}\hfill
\begin{subfigure}[b]{0.49\columnwidth} 
  \centering
  {\includegraphics[width=\textwidth]{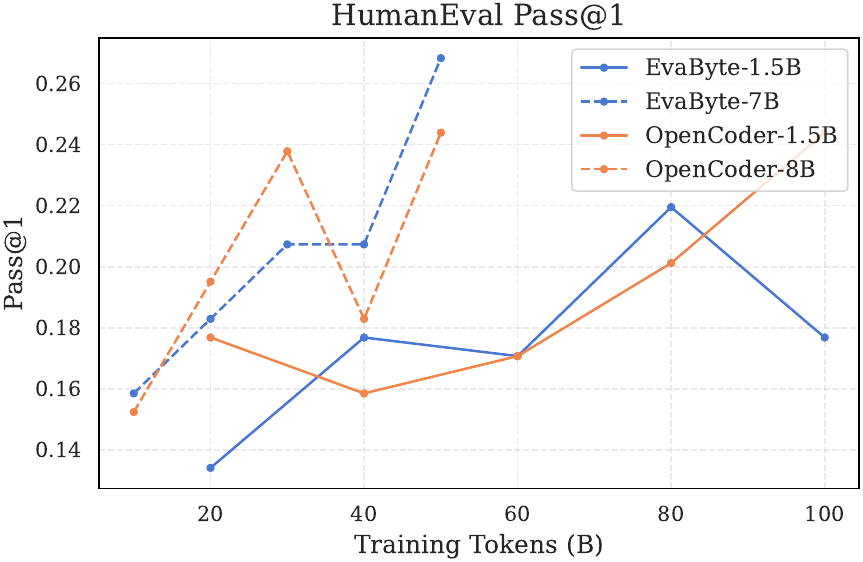}}
\end{subfigure}
\caption{Ablation on model architecture on validation BPB (left) and HumanEval pass@1 (right).}
\label{app:fig:ablation:model_arch}
\end{figure}

\begin{table}[ht]
\centering
\caption{Architectural hyperparameters. Vocabulary sizes for EvaByte are omitted here due to varying input representations in our study.}
\label{tb:model_arch_hp}
\begin{tabular}{l cc ccccc}
\toprule
\multirow{2}{*}{\textbf{Hyperparameter}} & \multicolumn{2}{c}{\textbf{OpenCoder}} & \multicolumn{5}{c}{\textbf{EvaByte}} \\
\cmidrule(lr){2-3} \cmidrule(lr){4-8}
& 1.5B & 8B & 0.5B & 1.5B & 4B & 7B & 14B \\
\midrule
Layers                  & 24     & 32      & 28     & 20      & 32      & 32     & 40     \\
Model Dimension         & 2240   & 4096    & 1024   & 2048    & 3072    & 4096   & 5120   \\
FFN Dimension           & 6144   & 14336   & 4096   & 8192    & 9216    & 12288  & 16384  \\
Attention Heads         & 14     & 32      & 8      & 16      & 24      & 32     & 40     \\
Key / Value Heads       & 14     & 8       & 8      & 16      & 24      & 32     & 40     \\
Vocab Size              & 96640  & 96640   & -      & -       & -       & -      & -      \\
RoPE $\theta$           & 10000  & 500000  & 100000 & 100000  & 100000  & 100000 & 100000 \\
Context Window Size     & 4096   & 4096    & 16384  & 16384   & 16384   & 16384  & 16384  \\
\bottomrule
\end{tabular}
\end{table}

\cref{app:fig:ablation:model_arch} compares EvaByte \citep{zheng2025evabyte} against OpenCoder \citep{huang2024opencoder}, a Llama-based transformer architecture \citep{dubey2024llama3}. We reproduce OpenCoder 1.5B and 8B under our codebase; our reproductions in a longer training run match or exceed the released intermediate checkpoints on HumanEval and MBPP benchmarks. EvaByte consistently achieves lower BPB while matching OpenCoder's downstream performance at comparable model sizes, where both models operate on tokens. Given its favorable efficiency–performance trade-off \citep{zheng2023eva,zheng2025evabyte}, we adopt EvaByte as the default architecture throughout this study. Following EvaByte \citep{zheng2025evabyte}, we employ multi-symbol (over either tokens or bytes) prediction \citep{stern2018blockwise,gloeckle2024multitoken,cai2024medusa,zheng2025evabyte,grivas2025mtpc} as the training objective above 4B parameters, and standard next-symbol prediction otherwise (enabling multi-symbol prediction for smaller models leads to slightly degraded performance, consistent with \citet{gloeckle2024multitoken}). Across different models, for byte-level predictions, we use 8 prediction heads; for token-level predictions from (proxy) compressors, we use either standard next-token prediction or 2-head multi-token prediction (we conducted ablation studies and found varying the number of token prediction heads does not improve downstream performance, including for our proxy models, which ultimately operate on raw bytes at inference). We choose the number of token- or byte-prediction heads to roughly match prediction granularity across representations, given that both tokenizer and neural proxy compressors achieve compression rates above $3\times$, although we found it does not lead to performance improvements over other configurations.

\subsection{Training Configuration}
\label{app:additional-details:training}
\begin{table}[ht]
\centering
\caption{Learning rates for different input representations and model sizes. All values are peak learning rates of the training schedule.}
\label{app:tb:lr_hp}
\begin{tabular}{l ccccc}
\toprule
Variant & 0.5B & 1.5B & 4B & 7B & 14B \\
\midrule
Tokenizer         & $1.2\text{e}\!-\!3$ & $9\text{e}\!-\!4$ & $7\text{e}\!-\!4$ & $5\text{e}\!-\!4$ & $5\text{e}\!-\!4$ \\
Byte-level        & $1.2\text{e}\!-\!3$ & $9\text{e}\!-\!4$ & $7\text{e}\!-\!4$ & $5\text{e}\!-\!4$ & $3\text{e}\!-\!4$ \\
Proxy (Neural)    & $1.2\text{e}\!-\!3$ & $9\text{e}\!-\!4$ & $3\text{e}\!-\!4$ & $5\text{e}\!-\!4$ & $3\text{e}\!-\!4$ \\
Proxy (Tokenizer) & $1.2\text{e}\!-\!3$ & $9\text{e}\!-\!4$ & $7\text{e}\!-\!4$ & $5\text{e}\!-\!4$ & $3\text{e}\!-\!4$ \\
\bottomrule
\end{tabular}
\end{table}

Model parameters are initialized from a truncated normal distribution with standard deviation 0.02, except for embedding parameters (with standard deviation 1.0). Gradient norms are clipped to 1.0. We use AdamW \citep{kingma2014adam,adamw} with weight decay 0.1, $\beta_1 = 0.9$, $\beta_2 = 0.95$, and $\epsilon = 1\text{e}\!-\!15$, following prior practices on stable large-scale training \citep{molybog2023adaminstability,wortsman2023smallforlargeinstabilities,olmo20242olmo2,zheng2025evabyte}. \cref{app:tb:lr_hp} reports peak learning rates for different input representations and model sizes. For each configuration, we perform a moderate learning rate sweep while keeping the batch size fixed, starting from common values and adjusting until observing either training instability or validation loss degradation.

For models trained on the Python subset, we use a fixed effective batch size of 2M sequence \emph{symbols}, independent of whether the inputs are represented as tokens, bytes, or other compressed formats. As a result, models with different input representations consume different amounts of raw data per batch. These models are trained for 50000 steps with a cosine learning-rate schedule, where the learning rate is linearly warmed up for the first 500 steps and then decayed down to 10\% of its peak value. For models trained on the full GitHub corpus (\cref{tab:downstream-transfer-longer-training}), we increase the batch size to 4M sequence symbols for 80000 training steps and use a constant learning rate schedule with 2000-step linear warm-up. All other settings remain unchanged.

\paragraph{Packing and Document Boundaries.}
Stateful proxy compressors such as gzip and neural compression operate on full raw documents \emph{before} packing; raw or compressed documents are then wrapped with format sentinels and packed into fixed-length contexts following standard practice. Thus, compressor state never crosses document boundaries. For neural compression, equal-information windows~\citep{lester2024training} additionally reset the compressor state at fixed bit budgets within a document (\cref{method:neural-proxies}), limiting the range over which compression state is carried. If long documents are truncated during context packing, truncation only removes already-compressed suffixes. Document-boundary attention masking within packed contexts (\cref{app:additional-exp-results:attnmask-vs-transfer}) further prevents cross-document attention.

\subsection{Evaluation Protocols}
\label{app:additional-details:evaluation}
We primarily focus on downstream code generation tasks for evaluation and report pass@$k$ rates following \citet{chen2021humaneval}, which estimate the probability of a model generating a correct solution within $k$ attempts. We evaluate on well-established code generation benchmarks, including HumanEval \citep{chen2021humaneval}, MBPP \citep{austin2021mbpp}, and their EvalPlus variants \citep{liu2023evalplus}. Pass@1 is calculated via greedy decoding, and pass@10 draws 20 samples at temperature 0.2 using nucleus sampling \citep{Holtzman2020nucleus} with $\text{top-}p = 0.95$. To accelerate decoding, we cap the maximum number of generated tokens to 512 for tokenizer-based baselines and 2048 for other input formats. We also measure \emph{Bits-Per-Byte} (BPB) for representation-agnostic comparison of modeling quality~\citep{rae2020compressive,xue2022byt5,yu2023megabyte,huang2024compression,xuyang2025compression}, computed on 40K held-out samples (${\sim}150$M bytes) from SwallowCode~\citep{fujii2025swallowcode}.

\paragraph{Note on BPB Evaluation.}
We focus on downstream task performance rather than validation BPB in this work. Validation BPB is known to be biased when comparing across different data representations~\citep{vieira2024language,hwang2025hnet}, and this bias is amplified in our setting where training involves tokens from different compressors, raw bytes, and their mixtures. Empirically, we observe that models trained with multi-symbol prediction objectives (whether over tokens or bytes) often achieve better downstream code generation but worse BPB, likely because their training objective diverges from next-symbol likelihood. Similarly, proxy-trained models exhibit worse BPB when evaluated on either compressed or raw representations alone, despite superior downstream performance. This is expected as at training time, the model must not only predict next symbols within each representation, but also align across representations to achieve transfer, which goes beyond standard next-symbol prediction. We therefore rely on downstream benchmarks as our primary metric, which we find yields reliable signals across models and scales.

\section{Additional Experimental Results and Analyses}
\label{app:additional-exp-results}

\subsection{Full Results of Downstream Transfer}
\label{app:additional-exp-results:downstream-transfer}
We list pass@1 rates on HumanEval and MBPP, as well as their EvalPlus variants for completeness, as shown in \cref{app:tab:downstream-transfer,app:tab:downstream-transfer-longer-training}. We also report HumanEval-Plus pass@10 rates in \cref{app:tab:downstream-transfer-pass10} and observe a consistent trend with pass@1 rates.

\begin{table*}[t]
    \centering
    \caption{Downstream pass@10 on HumanEval-Plus across model sizes and input representations under the same training setup as \cref{tab:downstream-transfer}.}
    \label{app:tab:downstream-transfer-pass10}
    \begin{tabular}{l c c c c c}
        \toprule
        \multirow{2}{*}{Model} & \multicolumn{5}{c}{Model Size} \\
        & 0.5B & 1.5B & 4B & 7B & 14B \\
        \midrule
        Tokenizer-based    & \textbf{24.3} & 28.3 & 38.4 & 43.4 & 44.2 \\
        Byte-level         & 22.9 & 27.5 & 32.0 & 32.7 & 35.3 \\
        Proxy (Neural)     & 21.6 & 26.5 & 33.6 & 42.1 & 44.5 \\
        Proxy (Tokenizer)  & 20.2 & \textbf{28.7} & \textbf{42.0} & \textbf{44.3} & \textbf{48.1} \\
        \bottomrule
    \end{tabular}
\end{table*}

\begin{table*}[t]
    \centering
    \caption{Downstream pass@1 performance on HumanEval(-Plus) and MBPP(-Plus) across different model sizes and input representations.}    
    \label{app:tab:downstream-transfer}
    \resizebox{0.8\textwidth}{!}{
    \begin{tabular}{l l c c c c c c}
        \toprule
        \multirow{2}{*}{Task} & \multirow{2}{*}{Model} & \multirow{2}{*}{Compression Rate} & \multicolumn{5}{c}{Model Size} \\
        & & & 0.5B & 1.5B & 4B & 7B & 14B \\
        \midrule
        \multirow{4}{*}{HumanEval}
            & Tokenizer-based & 3.7 & \textbf{21.3} & 21.3 & \textbf{33.5} & \textbf{32.9} & 34.1 \\
            & Byte-level & 1.0 & 17.7 & 21.3 & 26.8 & 26.2 & 27.4 \\
            & Proxy (Neural) & 2.6 & 15.2 & 21.3 & 26.2 & 30.5 & 32.9 \\
            & Proxy (Tokenizer) & 2.9 & 13.4 & \textbf{23.8} & 29.3 & \textbf{32.9} & \textbf{34.8} \\
        \midrule
        \multirow{4}{*}{HumanEval-Plus}
            & Tokenizer-based & 3.7 & \textbf{17.7} & 18.3 & \textbf{28.0} & \textbf{28.7} & 29.3 \\
            & Byte-level & 1.0 & 15.9 & 18.3 & 22.0 & 23.8 & 24.4 \\
            & Proxy (Neural) & 2.6 & 13.4 & 18.3 & 22.6 & 26.8 & 29.9 \\
            & Proxy (Tokenizer) & 2.9 & 12.2 & \textbf{20.7} & 24.4 & 26.2 & \textbf{30.5} \\
        \midrule
        \multirow{4}{*}{MBPP}
            & Tokenizer-based & 3.7 & \textbf{37.8} & \textbf{47.9} & \textbf{57.1} & 54.8 & \textbf{59.8} \\
            & Byte-level & 1.0 & 31.7 & 42.1 & 51.1 & 50.0 & 52.6 \\
            & Proxy (Neural) & 2.6 & 27.5 & 37.3 & 50.8 & 50.0 & 58.7 \\
            & Proxy (Tokenizer) & 2.9 & 30.7 & 46.0 & 54.2 & \textbf{56.3} & 58.2 \\
        \midrule
        \multirow{4}{*}{MBPP-Plus}
            & Tokenizer-based & 3.7 & \textbf{29.4} & \textbf{41.0} & \textbf{46.3} & 45.2 & 48.1 \\
            & Byte-level & 1.0 & 25.9 & 33.6 & 41.8 & 41.3 & 42.1 \\
            & Proxy (Neural) & 2.6 & 22.0 & 29.6 & 41.8 & 41.8 & 49.2 \\
            & Proxy (Tokenizer) & 2.9 & 25.4 & 38.4 & 44.4 & \textbf{45.5} & \textbf{49.5} \\
        \bottomrule
    \end{tabular}}
\end{table*}

\begin{table*}[t]
    \centering
    \caption{Downstream pass@1 performance on HumanEval(-Plus) and MBPP(-Plus) after training 320B sequence symbols on the full RefineCode GitHub corpus.}
    \label{app:tab:downstream-transfer-longer-training}
    \resizebox{0.9\textwidth}{!}{
    \begin{tabular}{c l c c c c c}
        \toprule
        \# Parameters & Model & Compression Rate & HumanEval & HumanEval-Plus & MBPP & MBPP-Plus \\
        \midrule
        \multirow{4}{*}{1.5B}
            & Tokenizer-based                        & 3.7 & \textbf{19.5} & \textbf{17.1} & \textbf{33.9} & \textbf{28.0} \\
            & Byte-level                             & 1.0 & 11.6 & 9.1  & 31.0 & 23.3 \\
            & Proxy (Neural)                         & 2.6 & 15.2 & 14.0 & 29.4 & 24.1 \\
            & Proxy (Tokenizer)                      & 2.9 & 14.6 & 12.8 & 32.8 & 25.1 \\
        \midrule
        \multirow{4}{*}{7B}
            & Tokenizer-based                        & 3.7 & \textbf{25.6} & 21.3 & 42.6 & \textbf{36.0} \\
            & Byte-level                             & 1.0 & 18.3 & 14.6 & 41.5 & 32.5 \\
            & Proxy (Neural)                         & 2.6 & 25.0 & 21.3 & \textbf{45.2} & \textbf{36.0} \\
            & Proxy (Tokenizer)                      & 2.9 & 23.2 & \textbf{22.0} & 43.4 & 34.9 \\
        \bottomrule
    \end{tabular}}
\end{table*}

\subsection{Additional Analyses of In-context Transfer}
\label{app:additional-exp-results:in-context-transfer}
Following \cref{experiments:in-context-transfer}, we provide additional experimental details and analyses for in-context transfer probing under controlled supervision. We consider the following mixed-representation prompt for evaluation,
\begin{align*}
    \big[ \boctok \circ \text{p}_{\text{comp}} \circ \text{s}_{\text{comp}} \circ \eoctok \circ \bortok \circ \text{p}_{\text{raw}} \big].
\end{align*}
We evaluate both tokenizer-based and neural compressors at 1.5B and 7B models, tracking oracle-translation pass@1 at intermediate training steps (10k, 20k, 30k, 40k, and 50k). As described in \cref{method:overview}, we optionally transform training inputs into explicit translation pairs by including both compressed and raw views of the same data in the same context. To study how in-context transfer strength depends on this pairing schedule, we consider three variants that share the same overall ratio $r=0.9$ of compressed samples and differ only in whether those samples are paired: (1) \emph{No pairs}, where training examples are independently sampled into either compressed or raw form without any paired data; (2) \emph{Warmup-only}, where the first 10k steps use translation pairs and later steps switch to independent sampling; and (3) \emph{Always-on}, where translation pairs are present throughout training. We identify two key findings from the extended in-context transfer experiments.

\paragraph{Structural differences in proxy compression drive transfer stability.}
As shown in \cref{tab:in-context-transfer}, under \emph{Warmup-only}, the tokenizer proxy drops after warmup but partially recovers (1.5B: $90.9\%{\to}31.1\%{\to}45.7\%$), while neural proxy decays more severely (1.5B: $90.9\%{\to}14.6\%{\to}38.4\%$), reflecting the structural properties of the two proxy compression types (\cref{fig:proxy-compression-comp}). Tokenizer-based compression uses a global static vocabulary where token IDs map consistently to byte patterns across the corpus, providing a stable anchor even without explicit supervision. Neural compression, in addition to its structured fuzziness (\cref{experiments:proxy-compression-comp}), produces context-dependent symbols that lack a fixed global mapping, where the same pattern may yield different compressed sequences depending on local context. This potentially explains why, under \textsc{Always-on}, both proxy compressors reach very high translation pass@1 (the model can always rely on the explicit translation pairing); whereas under \textsc{Warmup-only}, once explicit pairs disappear, neural proxies decay faster than tokenizer-based counterparts in oracle-translation pass@1.

\begin{figure}[t]
    \centering
    \definecolor{col1}{rgb}{0.12, 0.47, 0.71} 
    \definecolor{col2}{rgb}{1.00, 0.50, 0.05}
    \definecolor{col3}{rgb}{0.84, 0.15, 0.16}

    \begin{tikzpicture}[font=\small]
        \begin{axis}[
            width=\linewidth, height=5.4cm,
            xlabel={Training steps},
            ylabel={Oracle translation pass@1},
            xmin=5, xmax=55, ymin=0, ymax=1.05,
            xtick={10,20,30,40,50},
            xticklabels={10k, 20k, 30k, 40k, 50k},
            yticklabel={
              \pgfmathprintnumber[
                fixed,
                precision=2,
                use period,
              ]{\tick}
            },
            ytick={0,0.2,0.4,0.6,0.8,1.0},
            grid=both,
            minor grid style={dotted, gray!30},
            major grid style={gray!40},
            legend style={
                at={(0.03,0.05)}, anchor=south west,
                legend columns=1,
                draw=none, fill=none,
                font=\footnotesize,
                row sep=1pt
            },
            every axis plot post/.append style={line width=1.8pt},
            mark options={mark size=2.4pt, fill=white}
            ]
            \addplot+[mark=*, mark repeat=5, col1, solid]
            coordinates {(10,0.957)(20,0.939)(30,0.957)(40,0.963)(50,0.963)};
            \addlegendentry{1.5B \textsc{Always-On}}
            \addplot+[mark=square*, mark repeat=5, col2, solid]
            coordinates {(10,0.963)(20,0.957)(30,0.933)(40,0.957)(50,0.970)};
            \addlegendentry{7B \textsc{Always-On}}
            \addplot+[mark=triangle*, mark repeat=5, col1!70, dashed, dash pattern=on 5pt off 3pt]
            coordinates {(10,0.933)(20,0.396)(30,0.159)(40,0.226)(50,0.390)};
            \addlegendentry{1.5B \textsc{Warmup-Only}}
            \addplot+[mark=diamond*, mark repeat=5, col3, dashed, dash pattern=on 5pt off 3pt]
            coordinates {(10,0.957)(20,0.768)(30,0.280)(40,0.274)(50,0.152)};
            \addlegendentry{7B \textsc{Warmup-Only}}
        \end{axis}
    \end{tikzpicture}
    \caption{Oracle translation pass@1 on neural proxy compression. Solid lines represent continual translation pairs (\textsc{Always-on}), while dashed lines indicate abrupt removal after 10k steps (\textsc{Warmup-only}).}
    \label{app:fig:in-context-transfer-curves}
\end{figure}
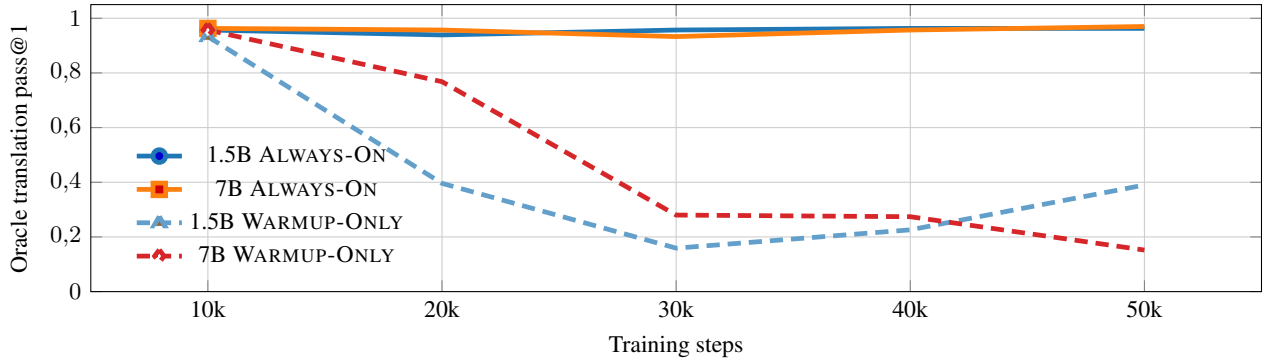

\paragraph{Larger models do not substitute for explicit pairing.}
\cref{app:fig:in-context-transfer-curves} visualizes the training dynamics for neural compression at 1.5B and 7B models. Under \emph{Warmup-only}, both 1.5B and 7B models exhibit rapid decay in oracle-translation accuracy once translation pairs are removed, despite 7B retaining slightly higher accuracy at intermediate steps (e.g., $73.8\%$ vs.\ $39.0\%$ at 20k). This suggests that in-context translation is fundamentally dependent on explicit pairing supervision: larger models do not internalize the superficial mapping more robustly, they simply start from a marginally better initial alignment. Once pairing is removed, the language modeling objective no longer constrains surface-level translation, and both scales converge toward similar degraded performance. In contrast, under \emph{Always-on}, both 1.5B and 7B maintain near-perfect translation ($>$94\%) throughout training, confirming that a small fraction of explicit pairs is both necessary and sufficient for reliable in-context transfer.

\subsection{Additional Results of Neural Proxy Compression}
\label{app:additional-exp-results:neural-proxy-analysis}

We provide additional analysis of the collision behavior induced by the neural compressor, complementing the main results in \cref{experiments:proxy-compression-comp}.

\begin{figure}[t]
\centering
\includegraphics[width=0.75\textwidth]{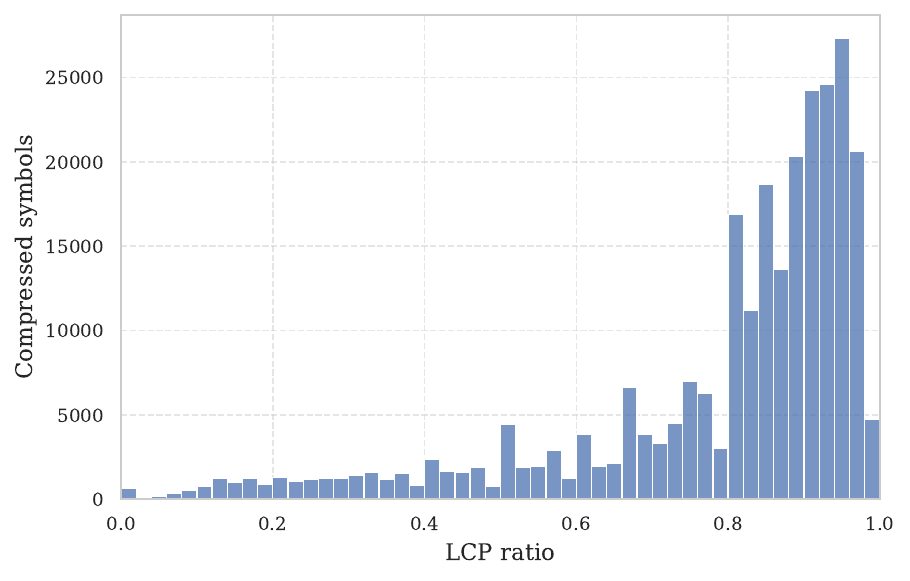}
\caption{Distribution of the LCP ratio across all compressed symbols that exhibit collisions. The distribution is heavily skewed towards high similarity (LCP ratio $> 0.8$), showing that collisions typically differ only in short suffixes.}
\label{app:fig:lcp-ratio}
\end{figure}

\paragraph{LCP Ratio Distribution.}
To quantify similarity among colliding chunks, we compute the \emph{longest common prefix (LCP) ratio}: the length of the shared prefix divided by average chunk length. \cref{app:fig:lcp-ratio} shows that over 90\% of collisions have LCP ratios above 0.8, meaning colliding chunks are nearly identical except for short suffixes. This confirms that neural compression induces structured rather than arbitrary ambiguity.

\paragraph{Representative Collision Cases.}
\cref{app:tab:collision-cases} presents statistics for four representative collision clusters, and \cref{app:lst:collision-examples} shows qualitative examples. Even in the largest cluster (Case~1, 61 variants), nearly all variants differ only by trailing whitespace and indentation. Cases~5--7 illustrate common patterns: URL suffixes, \texttt{if \_\_name\_\_} boilerplate with varying completions, and function calls with formatting differences. These examples demonstrate that neural compression merges semantically equivalent content while abstracting away superficial formatting noise.

\begin{table}[t]
\centering
\caption{Statistics of four representative collision clusters induced by the neural compressor.}
\label{app:tab:collision-cases}
\small
\begin{tabular}{lcccc}
\toprule
Case & \# Variants & LCP Ratio & Length (mean) & Std \\
\midrule
1. Maximum scale     & 61 & 0.49 & 59.00 & 17.60 \\
2. Minimum scale     &  2 & 0.56 & 64.00 & 28.00 \\
3. Highest LCP        &  2 & 0.87 & 7.50 &  0.50 \\
4. Lowest LCP         &  4 & 0.00 & 10.00 &  8.15 \\
\bottomrule
\end{tabular}
\end{table}

\begin{lstlisting}[
  caption={Collision examples from the neural compressor.},
  label={app:lst:collision-examples},
  basicstyle=\ttfamily\footnotesize,
  keepspaces=true,
  columns=fixed,
  basewidth=0.54em
]
Case 1: Maximum scale (61 variants, 4 shown)
[,\n                                             ]
[,\n                               ]
[,\n                                       ]
[,\n                                                                ]

Case 2: Minimum scale (2 variants)
[acc,\n                                                                           ]
[acc,\n                               ]

Case 3: Highest LCP ratio (2 variants)
[True\n\n ]
[True\n\n  ]

Case 4: Lowest LCP ratio (4 variants)
[ }]
[ }\n                ]
[!]
[ }\n            ]

Case 5: URL suffixes (4 variants)
[ = 1)\n\n# + colab={]
[ = 1)\n\n# + colab={"base_uri": "https://localhost:8080/]
[ = 1)\n\n# + colab={"base_uri": "https://localhost:8080/", "height": ]
[ = 1)\n\n# + colab]

Case 6: Main function boilerplate (11 variants)
[\nif __name__ == '_"]
[\nif __n]
[\nif __name__ == '__main]
[\nif __name__ == '__main__']
[\nif __]
[\nif _]
[\nif __name__ ]
[\nif __name__ =]
[\nif __name__ == '__]
[\nif __na]
[\nif __name__]

Case 7: Function call formatting (4 variants)
[process()\n        ]
[process(\n        ]
[process()\n    ]
[process(\n    ]
\end{lstlisting}

\subsection{Additional Results of Robustness Evaluation on ReCode}
\label{app:additional-exp-results:robustness}
\paragraph{Evaluation Metrics.}
We evaluate robustness on the HumanEval split of the ReCode benchmark \citep{wang2023recode}, which applies naturally occurring, semantics-preserving perturbations to coding problems. Following ReCode \citep{wang2023recode}, we consider four perturbation families: function name rewrites (\emph{Function}), formatting changes (\emph{Format}), syntactic rewrites (\emph{Syntax}), and docstring paraphrases (\emph{Docstrings}). We report the \emph{nominal} pass rate where no perturbations are applied. We measure robustness based on three metrics:
\emph{Robust Pass} $\mathrm{RP}_{s}@k$, measuring worst-case pass@$k$ under $s$ variants of the same problem. Using the standard pass@$k$ estimator with $n$ samples, robust pass rates are defined as $\mathrm{RP}_{s}@k=\mathbb{E}_{x}\!\big[1-\tbinom{n-r_c^{s}(x)}{k}\!/\!\tbinom{n}{k}\big]$, where $r_c^{s}(x)$ counts generations that pass \emph{all} $s$ variants \citep{wang2023recode}. Intuitively, RP measures the worst-case pass rates where a problem is solved only when the solution remains correct no matter how the prompt is perturbed (higher is better).
\emph{Robust Drop} $\mathrm{RD}_{s}@k=(\mathrm{pass}@k-\mathrm{RP}_{s}@k)/\mathrm{pass}@k$ measures the relative degradation from nominal to worst-case performance (lower is better; negative values indicate gains under perturbations).
\emph{Robust Relative (flip rate)} $\mathrm{RR}_{s}@k$ captures stability: the proportion of samples whose correctness flips between the nominal and perturbed inputs (lower is better).

For each original problem in HumanEval \citep{chen2021humaneval}, ReCode provides $s=5$ randomly perturbed variants for each perturbation type. We run greedy decoding with $n=1$ sample and $k=1$. We thus refer to the resulting quantities simply as RP, RD, and RR.

\paragraph{Models.}
We compare our proxy-trained models against the tokenizer-based and byte-level baselines at 7B parameters, using the same checkpoints as in \cref{tab:downstream-transfer}. To evaluate robustness more extensively, we additionally evaluate tokenizer-proxy-trained models by running inference on tokens, denoted by \textbf{Proxy (Tokenizer-tokens)}.

\paragraph{Per-Family Analysis.}

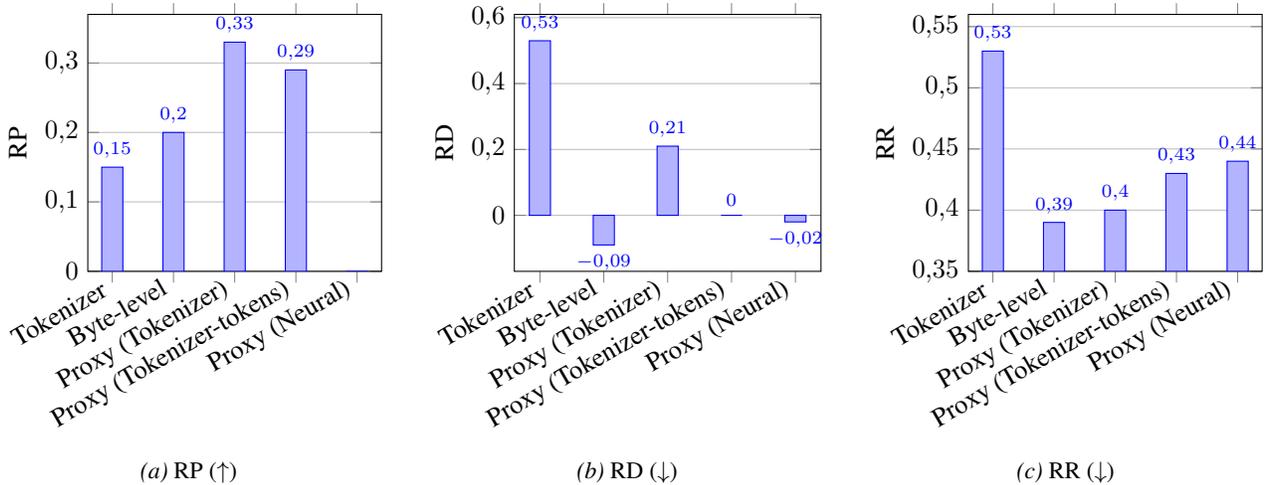
\begin{figure}[t]
\centering
\begin{subfigure}{.32\textwidth}
\centering
\begin{tikzpicture}
\begin{axis}[
    ybar, ymin=0, ymax=0.37,
    width=\linewidth, height=5.0cm,
    ylabel={RP}, ymajorgrids=true,
    symbolic x coords={tok,byte,praw,pspm,pneu},
    xticklabels={Tokenizer,Byte-level,Proxy (Tokenizer),Proxy (Tokenizer-tokens),Proxy (Neural)},
    xtick=data, x tick label style={rotate=30,anchor=east},
    yticklabel={
      \pgfmathprintnumber[
        fixed,
        precision=2,
        use period,
      ]{\tick}
    },
    bar width=8pt, nodes near coords, nodes near coords style={font=\scriptsize,/pgf/number format/use period,/pgf/number format/fixed,/pgf/number format/precision=2}
]
\addplot coordinates {(tok,0.15) (byte,0.29) (praw,0.33) (pspm,0.31) (pneu,0.29)};
\end{axis}
\end{tikzpicture}
\caption{RP ($\uparrow$)}
\end{subfigure}\hfill
\begin{subfigure}{.33\textwidth}
\centering
\begin{tikzpicture}
\begin{axis}[
    ybar, ymin=-0.17, ymax=0.61,
    width=\linewidth, height=5.0cm,
    ylabel={RD}, ymajorgrids=true,
    symbolic x coords={tok,byte,praw,pspm,pneu},
    xticklabels={Tokenizer,Byte-level,Proxy (Tokenizer),Proxy (Tokenizer-tokens),Proxy (Neural)},
    xtick=data, x tick label style={rotate=30,anchor=east},
    yticklabel={
      \pgfmathprintnumber[
        fixed,
        precision=2,
        use period,
      ]{\tick}
    },
    bar width=8pt, nodes near coords, nodes near coords style={font=\scriptsize,/pgf/number format/use period,/pgf/number format/fixed,/pgf/number format/precision=2}
]
\addplot coordinates {(tok,0.53) (byte,-0.09) (praw,0.21) (pspm,0.00) (pneu,-0.02)};
\end{axis}
\end{tikzpicture}
\caption{RD ($\downarrow$)}
\end{subfigure}\hfill
\begin{subfigure}{.32\textwidth}
\centering
\begin{tikzpicture}
\begin{axis}[
    ybar, ymin=0.35, ymax=0.56,
    width=\linewidth, height=5.0cm,
    ylabel={RR}, ymajorgrids=true,
    symbolic x coords={tok,byte,praw,pspm,pneu},
    xticklabels={Tokenizer,Byte-level,Proxy (Tokenizer),Proxy (Tokenizer-tokens),Proxy (Neural)},
    xtick=data, x tick label style={rotate=30,anchor=east},
    yticklabel={
      \pgfmathprintnumber[
        fixed,
        precision=2,
        use period,
      ]{\tick}
    },
    bar width=8pt, nodes near coords, nodes near coords style={font=\scriptsize,/pgf/number format/use period,/pgf/number format/fixed,/pgf/number format/precision=2}
]
\addplot coordinates {(tok,0.53) (byte,0.39) (praw,0.40) (pspm,0.43) (pneu,0.44)};
\end{axis}
\end{tikzpicture}
\caption{RR ($\downarrow$)}
\end{subfigure}
\caption{Format perturbations.}
\label{fig:format-bars}
\end{figure}

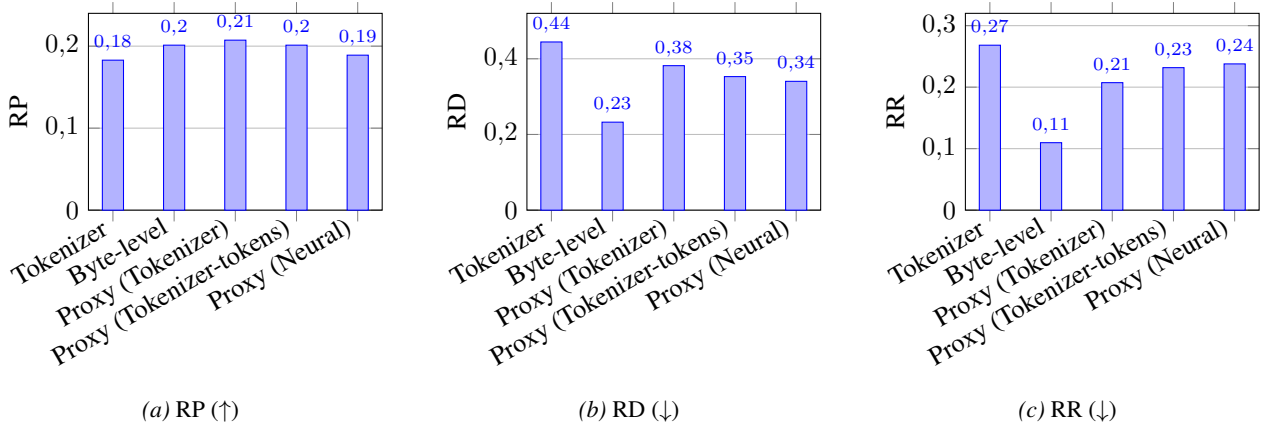
\begin{figure}[h]
\centering
\begin{subfigure}{.32\textwidth}
\centering
\begin{tikzpicture}
\begin{axis}[ybar,ymin=0,ymax=0.24,width=\linewidth,height=4.2cm,ylabel={RP},ymajorgrids=true,
symbolic x coords={tok,byte,praw,pspm,pneu},
xticklabels={Tokenizer,Byte-level,Proxy (Tokenizer),Proxy (Tokenizer-tokens),Proxy (Neural)},
xtick=data,x tick label style={rotate=30,anchor=east},bar width=8pt,
yticklabel={
  \pgfmathprintnumber[
    fixed,
    precision=2,
    use period,
  ]{\tick}
},
nodes near coords, nodes near coords style={font=\scriptsize,/pgf/number format/use period,/pgf/number format/fixed,/pgf/number format/precision=2}]
\addplot coordinates {(tok,0.1829) (byte,0.2012) (praw,0.2073) (pspm,0.2012) (pneu,0.1890)};
\end{axis}
\end{tikzpicture}
\caption{RP ($\uparrow$)}
\end{subfigure}\hfill
\begin{subfigure}{.32\textwidth}
\centering
\begin{tikzpicture}
\begin{axis}[ybar,ymin=0,ymax=0.52,width=\linewidth,height=4.2cm,ylabel={RD},ymajorgrids=true,
 symbolic x coords={tok,byte,praw,pspm,pneu},
 xticklabels={Tokenizer,Byte-level,Proxy (Tokenizer),Proxy (Tokenizer-tokens),Proxy (Neural)},
xtick=data,x tick label style={rotate=30,anchor=east},bar width=8pt,
yticklabel={
  \pgfmathprintnumber[
    fixed,
    precision=2,
    use period,
  ]{\tick}
},
nodes near coords, nodes near coords style={font=\scriptsize,/pgf/number format/use period,/pgf/number format/fixed,/pgf/number format/precision=2}]
\addplot coordinates {(tok,0.4444) (byte,0.2325) (praw,0.3818) (pspm,0.3529) (pneu,0.3404)};
\end{axis}
\end{tikzpicture}
\caption{RD ($\downarrow$)}
\end{subfigure}\hfill
\begin{subfigure}{.32\textwidth}
\centering
\begin{tikzpicture}
\begin{axis}[ybar,ymin=0.0,ymax=0.32,width=\linewidth,height=4.2cm,ylabel={RR},ymajorgrids=true,
symbolic x coords={tok,byte,praw,pspm,pneu},
xticklabels={Tokenizer,Byte-level,Proxy (Tokenizer),Proxy (Tokenizer-tokens),Proxy (Neural)},
xtick=data,x tick label style={rotate=30,anchor=east},bar width=8pt,
yticklabel={
  \pgfmathprintnumber[
    fixed,
    precision=2,
    use period,
  ]{\tick}
},
nodes near coords, nodes near coords style={font=\scriptsize,/pgf/number format/use period,/pgf/number format/precision=2}]
\addplot coordinates {(tok,0.2682) (byte,0.1097) (praw,0.2073) (pspm,0.2317) (pneu,0.2378)};
\end{axis}
\end{tikzpicture}
\caption{RR ($\downarrow$)}
\end{subfigure}
\caption{Function perturbations.}
\label{fig:func-bars}
\end{figure}

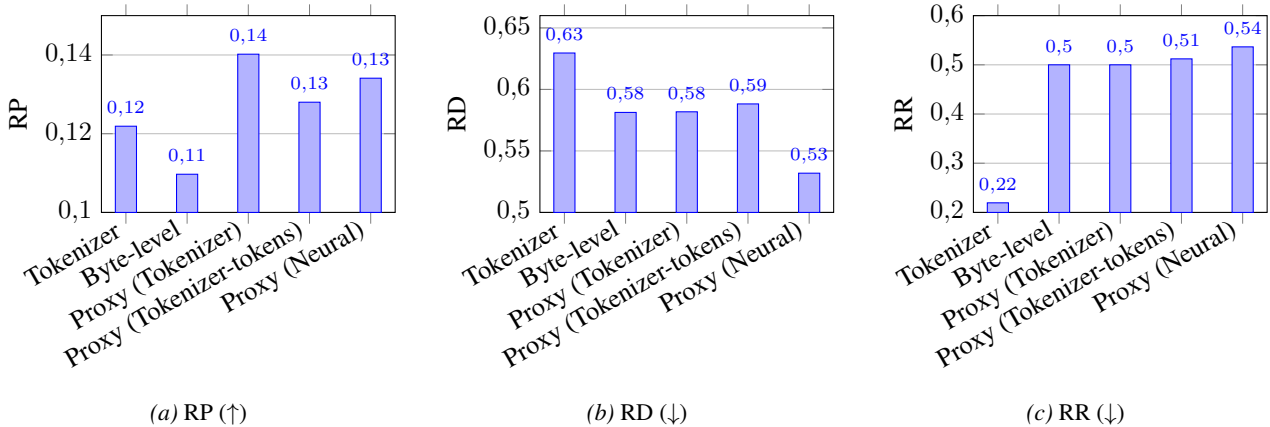
\begin{figure}[h]
\centering
\begin{subfigure}{.32\textwidth}
\centering
\begin{tikzpicture}
\begin{axis}[ybar,ymin=0.10,ymax=0.15,width=\linewidth,height=4.2cm,ylabel={RP},ymajorgrids=true,
symbolic x coords={tok,byte,praw,pspm,pneu},
xticklabels={Tokenizer,Byte-level,Proxy (Tokenizer),Proxy (Tokenizer-tokens),Proxy (Neural)},
xtick=data,x tick label style={rotate=30,anchor=east},bar width=8pt,
yticklabel={
  \pgfmathprintnumber[
    fixed,
    precision=2,
    use period,
  ]{\tick}
},
nodes near coords, nodes near coords style={font=\scriptsize,/pgf/number format/use period,/pgf/number format/precision=2}]
\addplot coordinates {(tok,0.1219) (byte,0.1097) (praw,0.1402) (pspm,0.1280) (pneu,0.1341)};
\end{axis}
\end{tikzpicture}
\caption{RP ($\uparrow$)}
\end{subfigure}\hfill
\begin{subfigure}{.32\textwidth}
\centering
\begin{tikzpicture}
\begin{axis}[ybar,ymin=0.50,ymax=0.66,width=\linewidth,height=4.2cm,ylabel={RD},ymajorgrids=true,
symbolic x coords={tok,byte,praw,pspm,pneu},
xticklabels={Tokenizer,Byte-level,Proxy (Tokenizer),Proxy (Tokenizer-tokens),Proxy (Neural)},
xtick=data,x tick label style={rotate=30,anchor=east},bar width=8pt,
yticklabel={
  \pgfmathprintnumber[
    fixed,
    precision=2,
    use period,
  ]{\tick}
},
nodes near coords, nodes near coords style={font=\scriptsize,/pgf/number format/use period,/pgf/number format/precision=2}]
\addplot coordinates {(tok,0.6296) (byte,0.5813) (praw,0.5818) (pspm,0.5882) (pneu,0.5319)};
\end{axis}
\end{tikzpicture}
\caption{RD ($\downarrow$)}
\end{subfigure}\hfill
\begin{subfigure}{.32\textwidth}
\centering
\begin{tikzpicture}
\begin{axis}[ybar,ymin=0.20,ymax=0.60,width=\linewidth,height=4.2cm,ylabel={RR},ymajorgrids=true,
symbolic x coords={tok,byte,praw,pspm,pneu},
xticklabels={Tokenizer,Byte-level,Proxy (Tokenizer),Proxy (Tokenizer-tokens),Proxy (Neural)},
xtick=data,x tick label style={rotate=30,anchor=east},bar width=8pt,
yticklabel={
  \pgfmathprintnumber[
    fixed,
    precision=2,
    use period,
  ]{\tick}
},
nodes near coords, nodes near coords style={font=\scriptsize,/pgf/number format/use period,/pgf/number format/precision=2}]
\addplot coordinates {(tok,0.2195) (byte,0.5000) (praw,0.5000) (pspm,0.5121) (pneu,0.5365)};
\end{axis}
\end{tikzpicture}
\caption{RR ($\downarrow$)}
\end{subfigure}
\caption{Syntax perturbations.}
\label{fig:syntax-bars}
\end{figure}

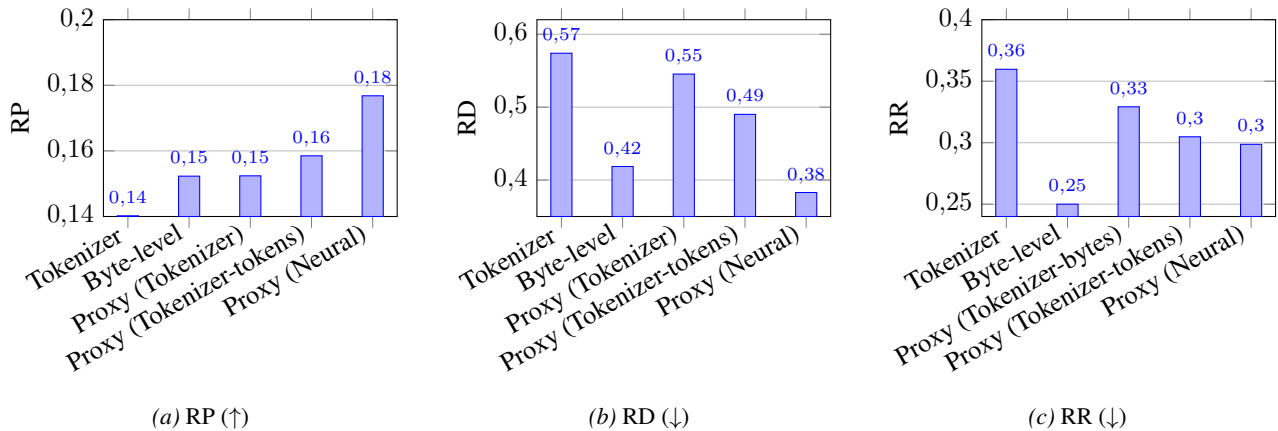
\begin{figure}[h]
\centering
\begin{subfigure}{.32\textwidth}
\centering
\begin{tikzpicture}
\begin{axis}[ybar,ymin=0.14,ymax=0.20,width=\linewidth,height=4.2cm,ylabel={RP},ymajorgrids=true,
symbolic x coords={tok,byte,praw,pspm,pneu},
xticklabels={Tokenizer,Byte-level,Proxy (Tokenizer),Proxy (Tokenizer-tokens),Proxy (Neural)},
xtick=data,x tick label style={rotate=30,anchor=east},bar width=8pt,
yticklabel={
  \pgfmathprintnumber[
    fixed,
    precision=2,
    use period,
  ]{\tick}
},
nodes near coords, nodes near coords style={font=\scriptsize,/pgf/number format/use period,/pgf/number format/precision=2}]
\addplot coordinates {(tok,0.1402) (byte,0.1523) (praw,0.1524) (pspm,0.1585) (pneu,0.1768)};
\end{axis}
\end{tikzpicture}
\caption{RP ($\uparrow$)}
\end{subfigure}\hfill
\begin{subfigure}{.32\textwidth}
\centering
\begin{tikzpicture}
\begin{axis}[ybar,ymin=0.35,ymax=0.62,width=\linewidth,height=4.2cm,ylabel={RD},ymajorgrids=true,
symbolic x coords={tok,byte,praw,pspm,pneu},
xticklabels={Tokenizer,Byte-level,Proxy (Tokenizer),Proxy (Tokenizer-tokens),Proxy (Neural)},
xtick=data,x tick label style={rotate=30,anchor=east},bar width=8pt,
yticklabel={
  \pgfmathprintnumber[
    fixed,
    precision=2,
    use period,
  ]{\tick}
},
nodes near coords, nodes near coords style={font=\scriptsize,/pgf/number format/use period,/pgf/number format/precision=2}]
\addplot coordinates {(tok,0.5740) (byte,0.4186) (praw,0.5454) (pspm,0.4901) (pneu,0.3829)};
\end{axis}
\end{tikzpicture}
\caption{RD ($\downarrow$)}
\end{subfigure}\hfill
\begin{subfigure}{.32\textwidth}
\centering
\begin{tikzpicture}
\begin{axis}[ybar,ymin=0.24,ymax=0.40,width=\linewidth,height=4.2cm,ylabel={RR},ymajorgrids=true,
symbolic x coords={tok,byte,praw,pspm,pneu},
xticklabels={Tokenizer,Byte-level,Proxy (Tokenizer),Proxy (Tokenizer-tokens),Proxy (Neural)},
xtick=data,x tick label style={rotate=30,anchor=east},bar width=8pt,
yticklabel={
  \pgfmathprintnumber[
    fixed,
    precision=2,
    use period,
  ]{\tick}
},
nodes near coords, nodes near coords style={font=\scriptsize,/pgf/number format/use period,/pgf/number format/precision=2}]
\addplot coordinates {(tok,0.3597) (byte,0.2500) (praw,0.3292) (pspm,0.3048) (pneu,0.2987)};
\end{axis}
\end{tikzpicture}
\caption{RR ($\downarrow$)}
\end{subfigure}
\caption{Docstring perturbations.}
\label{fig:doc-bars}
\end{figure}

We visualize RP, RD, and RR for each perturbation family. \emph{Format perturbations} (\cref{fig:format-bars}) exhibit the largest gap between representations. The tokenizer baseline suffers severe degradation, while byte-level and proxy models remain stable or even improve. This indicates that partial exposure to raw bytes during training suffices to improve model robustness against surface-level formatting noise. We also visualize results for \emph{Function perturbations} (\cref{fig:func-bars}), \emph{Syntax perturbations} (\cref{fig:syntax-bars}), and \emph{Docstring perturbations} (\cref{fig:doc-bars}).

\subsection{Additional Analyses on Data Representations}
\label{app:additional-exp-results:data-representation}
While tokenization remains the dominant approach for transforming raw inputs into discrete units for language models, we systematically investigate a spectrum of input representations ranging from BPE tokens down to individual bits. Specifically, we consider BPE tokens, double-bytes (16-bit), bytes (8-bit), half-bytes (4-bit), double-bits (2-bit), and bits (1-bit), training models with matched parameter counts and training FLOPs across all representations.

\begin{figure}[tb]
\centering
\begin{subfigure}[b]{0.48\columnwidth} 
  \centering
  {\includegraphics[width=\textwidth]{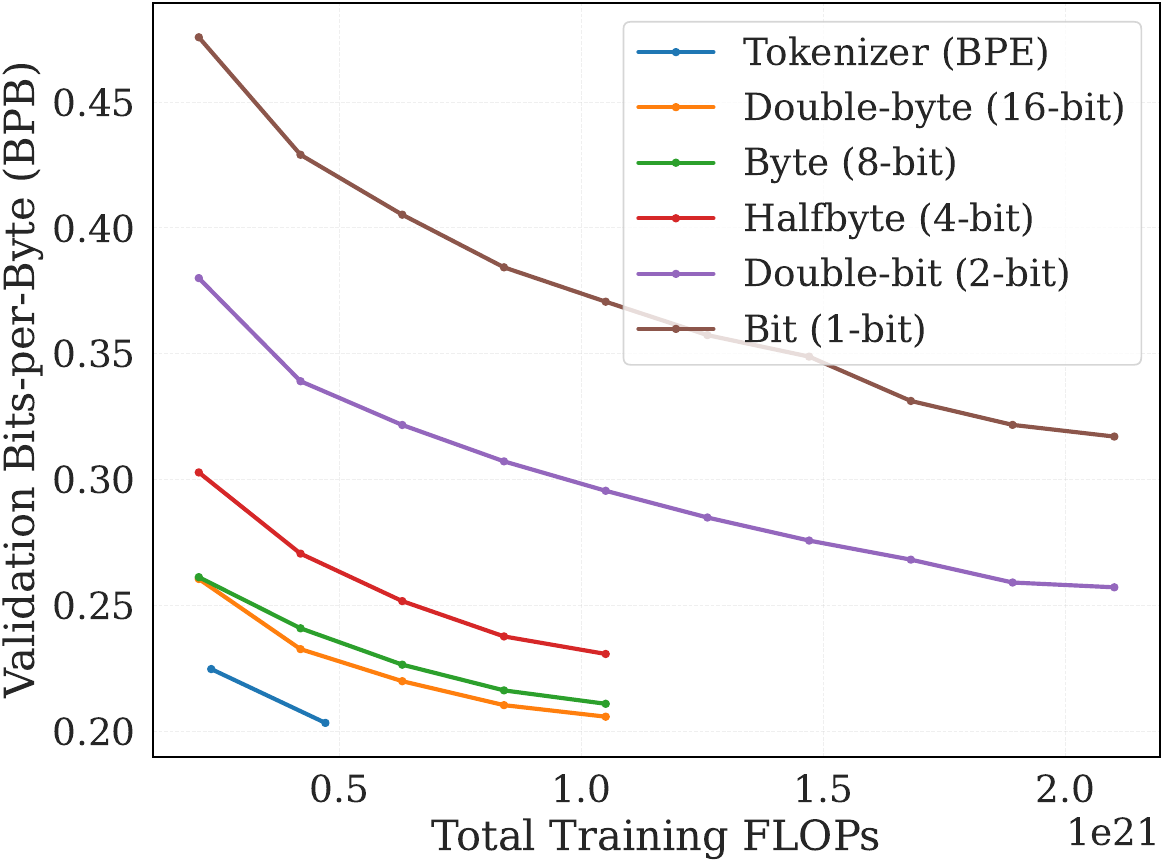}}
\end{subfigure}\hfill
\begin{subfigure}[b]{0.48\columnwidth}
  \centering
  {\includegraphics[width=\textwidth]{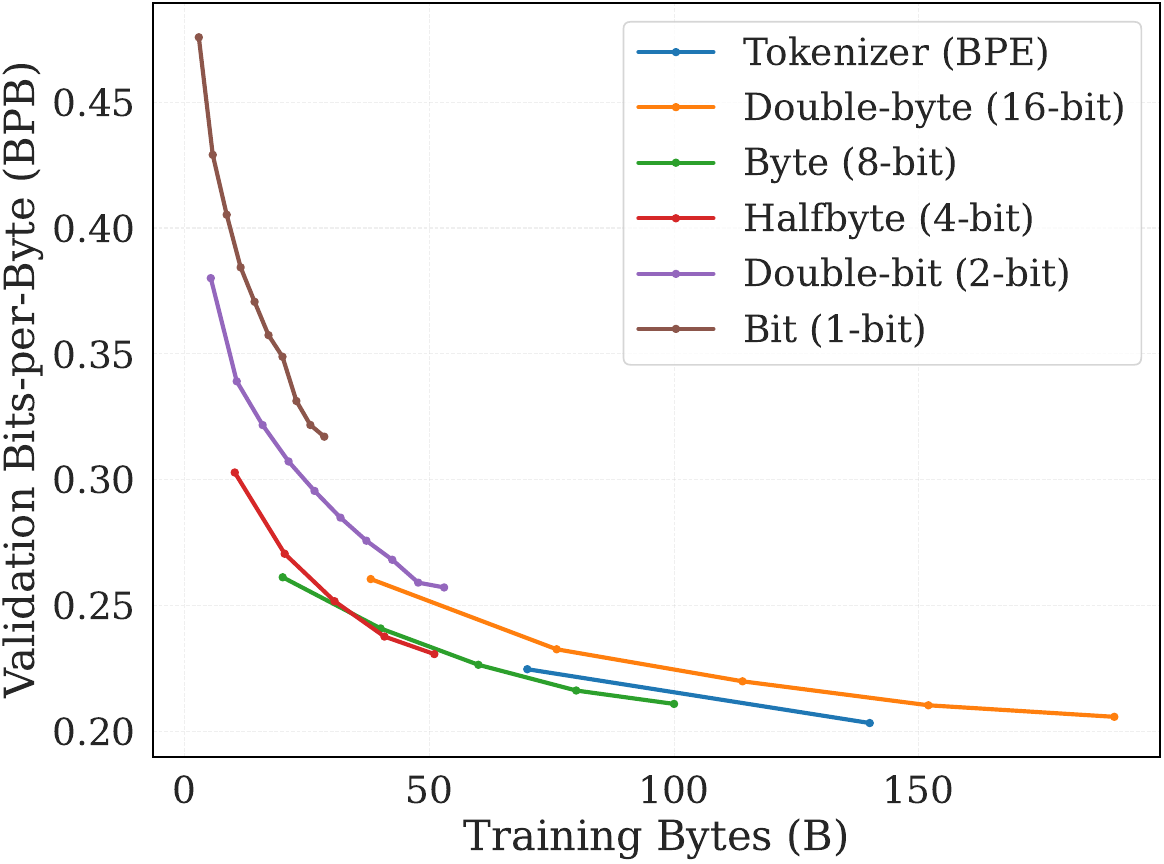}}
\end{subfigure}
\caption{Validation BPB performance of different data representations spanning tokens to bits, under FLOPs-matched (left) and data-matched (right) comparison.}
\label{app:fig:data_representations}
\end{figure}

\cref{app:fig:data_representations} reveals two complementary trends. Under a fixed compute budget, representations that process more data per FLOP achieve lower validation BPB, indicating superior \emph{compute efficiency}. Conversely, under a fixed data budget (right), lower-level representations generally outperform higher-level ones, reflecting better \emph{data efficiency}, consistent with prior findings on byte-level models \citep{xue2022byt5,zheng2025evabyte}. However, this trend does not extrapolate to the finest granularities: 2-bit and 1-bit models consistently underperform across both regimes. This suggests that excessively fine-grained representations waste compute greatly and lack sufficient abstraction for effective learning, even when granted additional training steps or data. We observe consistent trends in longer training runs (\cref{app:additional-exp-results:data-compute-efficiency}). These findings motivate representations that balance abstraction with granularity, rather than pursuing either extreme.

\subsection{On Document-boundary Attention Masking}
\label{app:additional-exp-results:attnmask-vs-transfer}
In proxy compression training, multiple samples potentially under different representations are concatenated and packed into fixed-length contexts, following standard language model training practice. This raises a natural question: does cross-representation transfer arise from in-context attention between samples of different representations, or from shared model parameters?

\begin{figure}[tb]
\centering
{\includegraphics[width=0.5\columnwidth]{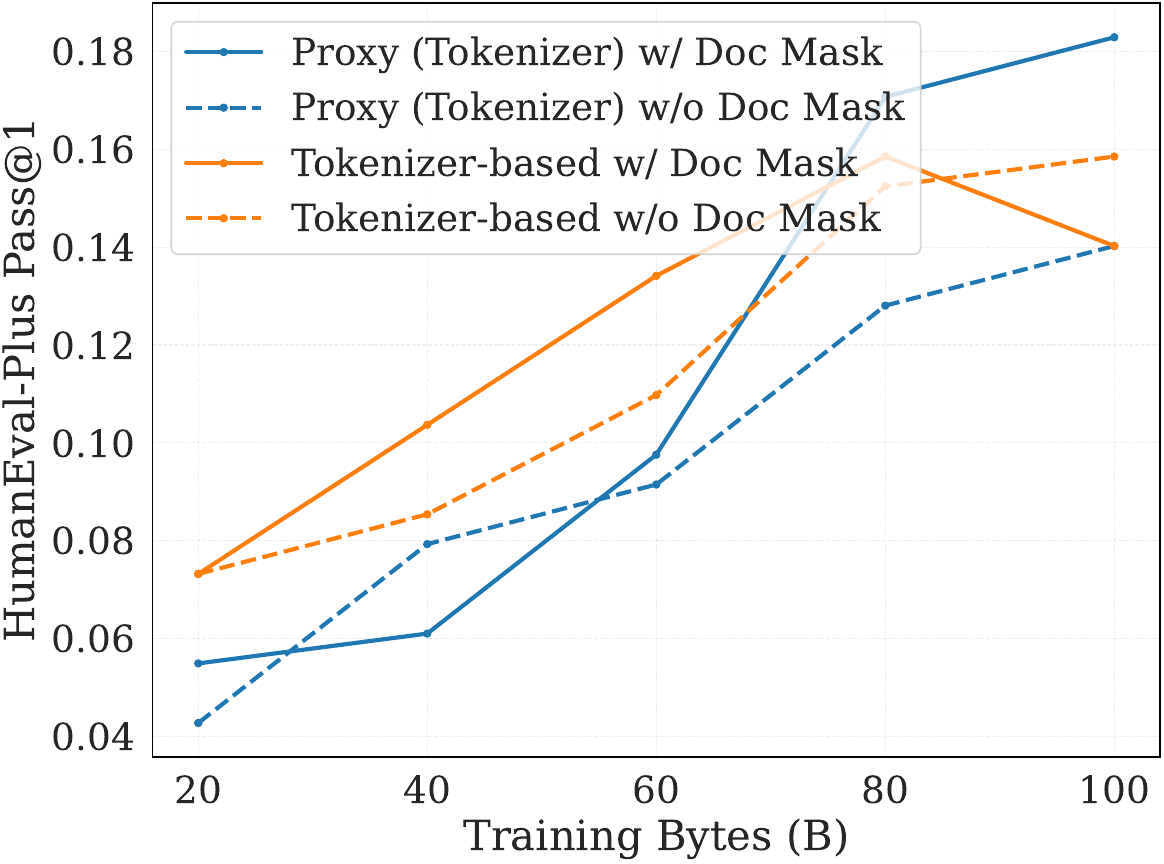}}
\caption{Ablation on document-boundary attention masking.}
\label{app:fig:ablation:docmask}
\end{figure}

To isolate this, we compare two settings: (1) standard packing, where samples within a context can attend to each other, and (2) document-boundary attention masking, which restricts attention to within-document tokens only. As shown in \cref{app:fig:ablation:docmask}, we observe performance improvements when preventing cross-document attention. This suggests that cross-representation transfer stems primarily from shared parameters rather than in-context interactions between representations. Following recent work demonstrating benefits of document-boundary masking for training \citep{zhu2025skyladder}, we enable it by default in all other experiments unless otherwise specified.

\subsection{Additional Results on Data Efficiency versus Compute Efficiency}
\label{app:additional-exp-results:data-compute-efficiency}

In the main text (\cref{experiments:main-transfer}), we showed that proxy compression captures the best of both data efficiency and compute efficiency for 14B models (\cref{fig:14B-humaneval_pass1-vs-flops-and-data}). Here we provide extended analysis across model scales and training horizons.

We note that intermediate checkpoint comparisons can be partially confounded by learning-rate annealing; models compared at the same amount of consumed data can correspond to different training steps and thus different points on the learning-rate schedule. Final-checkpoint pass@1 (\cref{tab:downstream-transfer}) remains the primary signal, with the FLOPs/data plots showing the trajectory during training.

\begin{figure}[p]
\centering

\begin{subfigure}[b]{0.78\textwidth}
  \centering
  \includegraphics[width=0.45\linewidth]{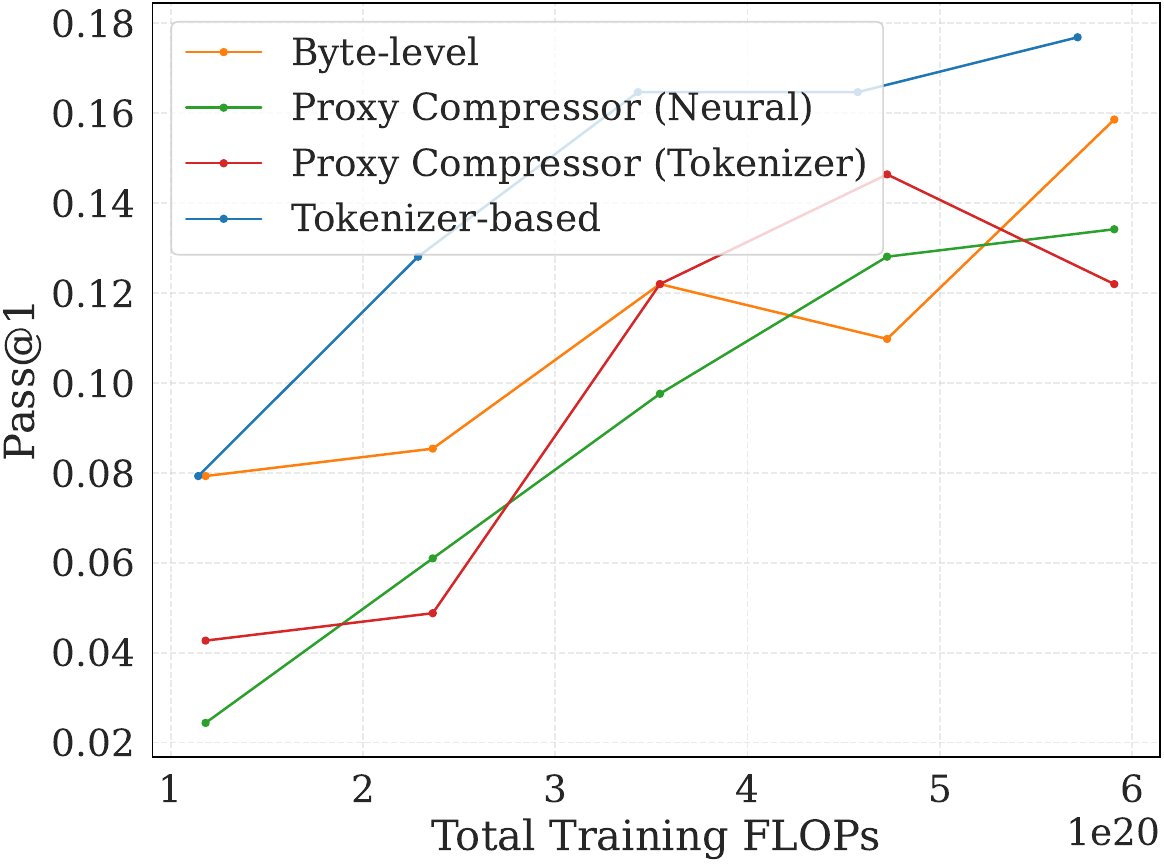}
  \hspace{0.04\linewidth}
  \includegraphics[width=0.45\linewidth]{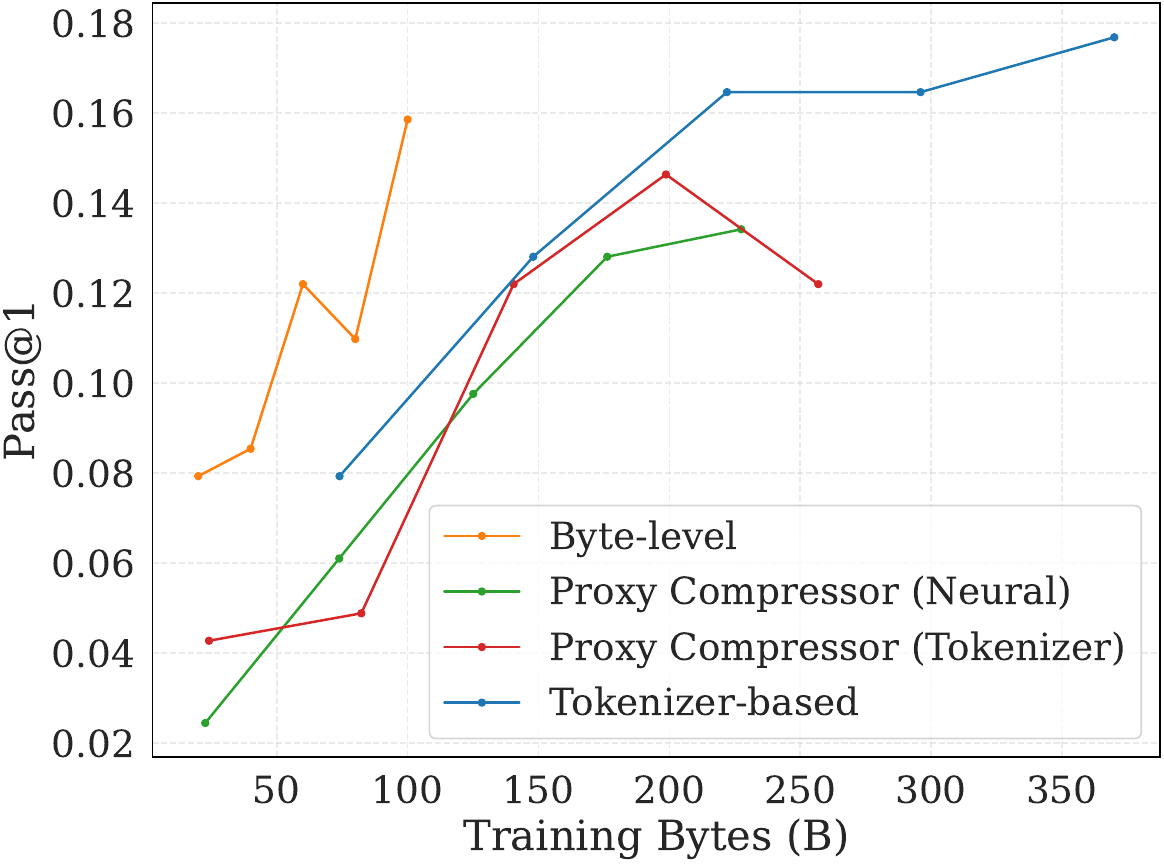}
  \caption{Pass@1 performance on HumanEval-Plus for 0.5B models.}
  \label{fig:0B5-humaneval_pass1-vs-flops-and-data}
\end{subfigure}

\vspace{0.7em}

\begin{subfigure}[b]{0.78\textwidth}
  \centering
  \includegraphics[width=0.45\linewidth]{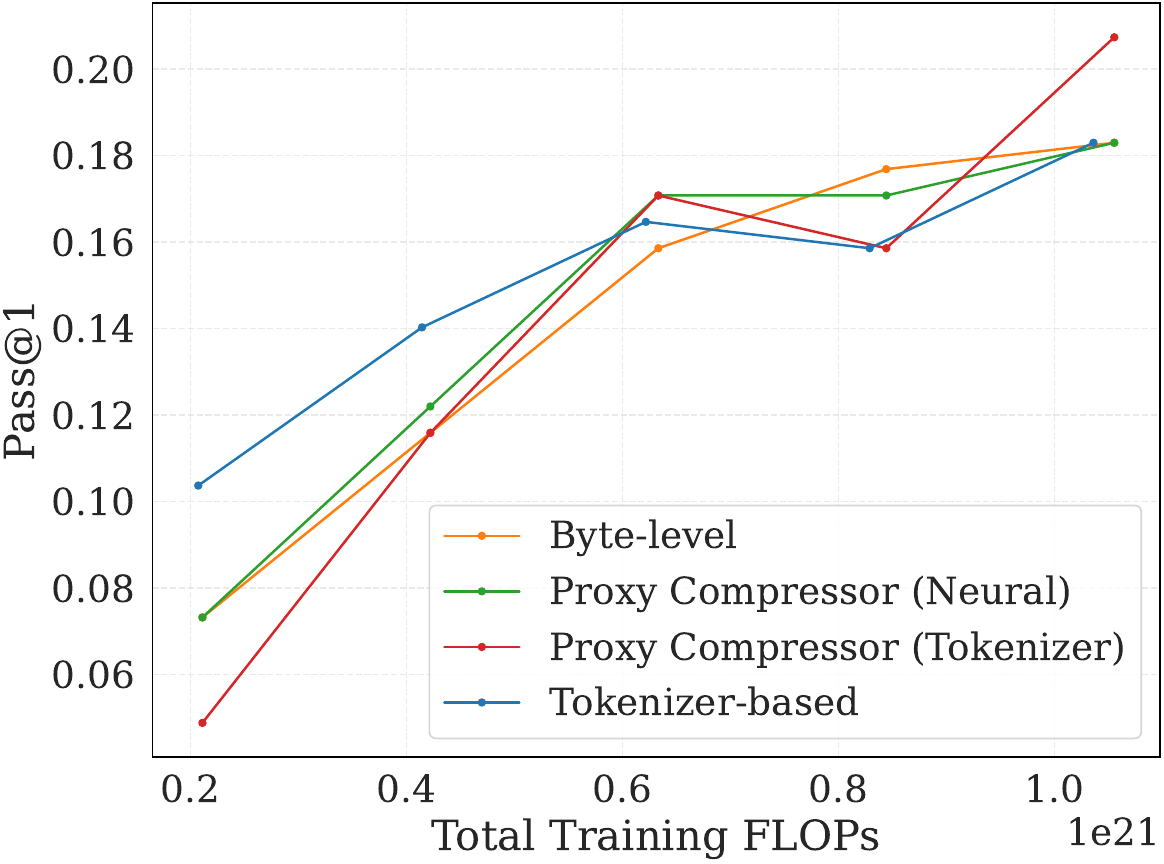}
  \hspace{0.04\linewidth}
  \includegraphics[width=0.45\linewidth]{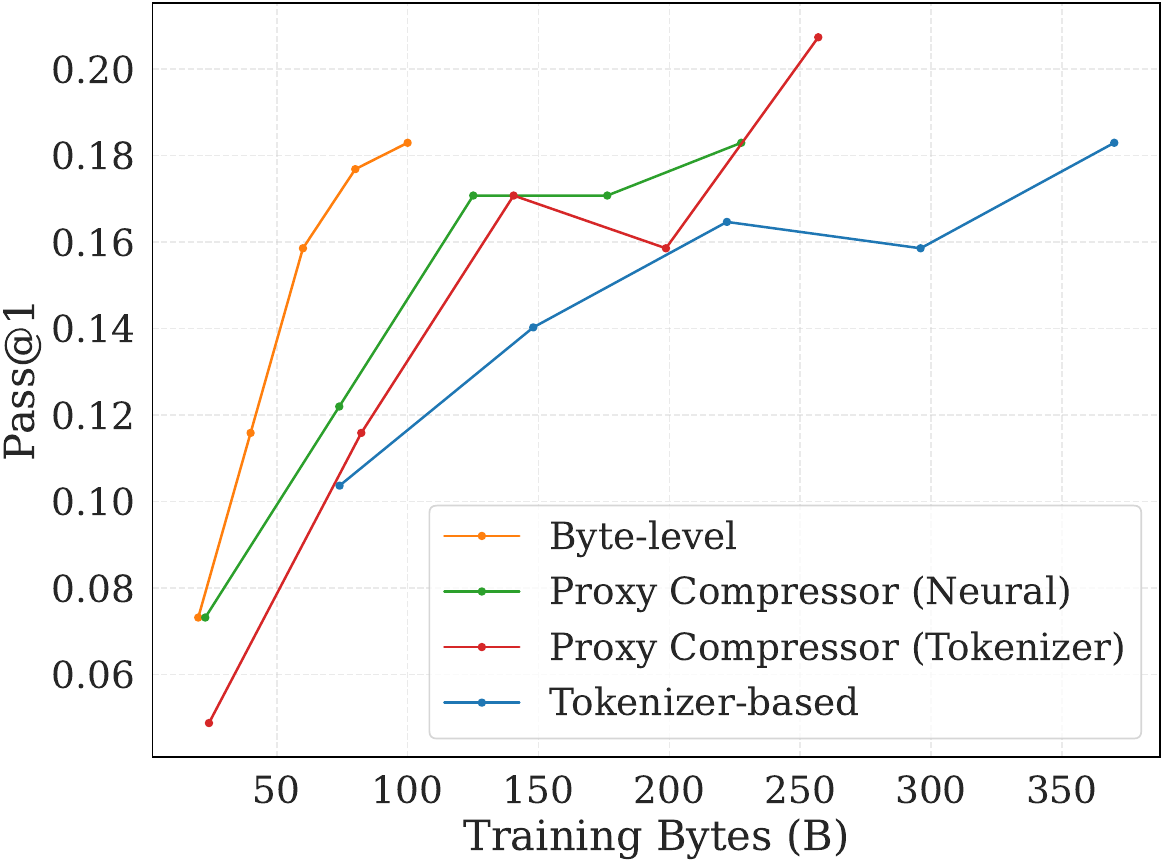}
  \caption{Pass@1 performance on HumanEval-Plus for 1.5B models.}
  \label{fig:1B5-humaneval_pass1-vs-flops-and-data}
\end{subfigure}

\vspace{0.7em}

\begin{subfigure}[b]{0.78\textwidth}
  \centering
  \includegraphics[width=0.45\linewidth]{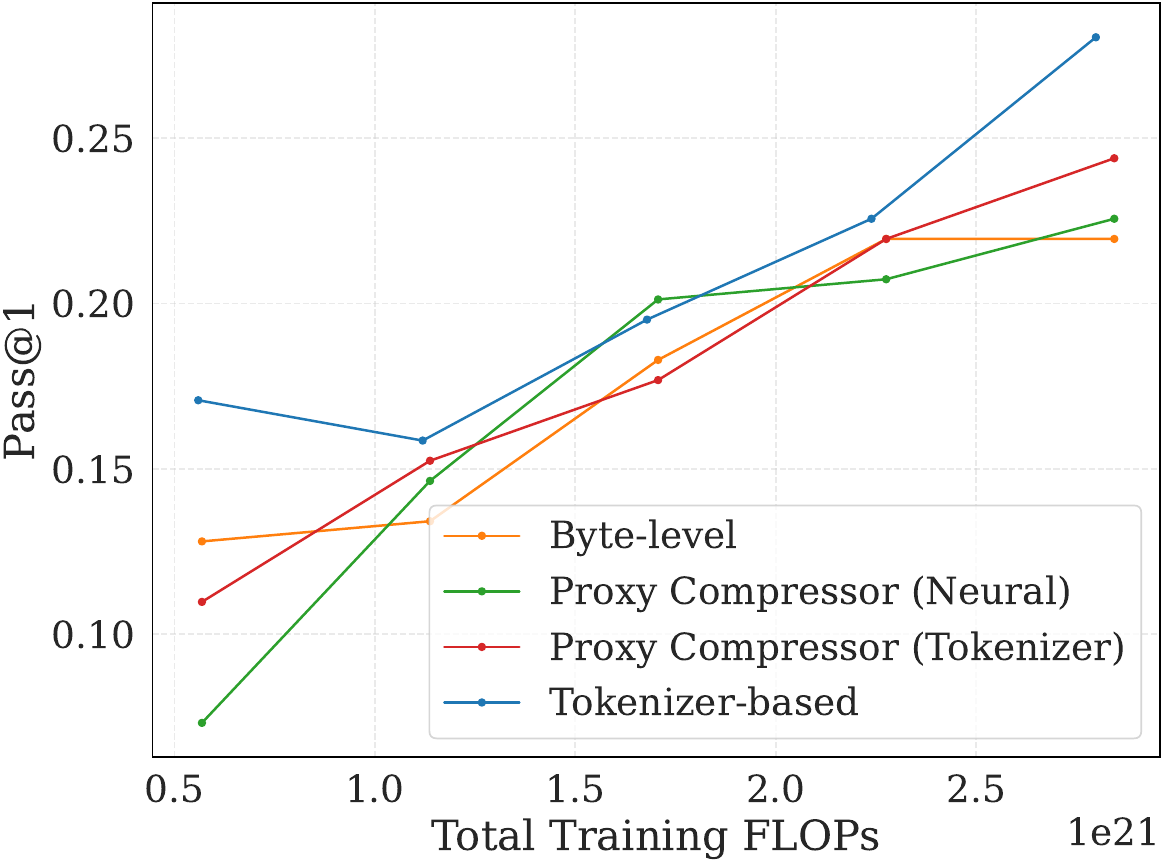}
  \hspace{0.04\linewidth}
  \includegraphics[width=0.45\linewidth]{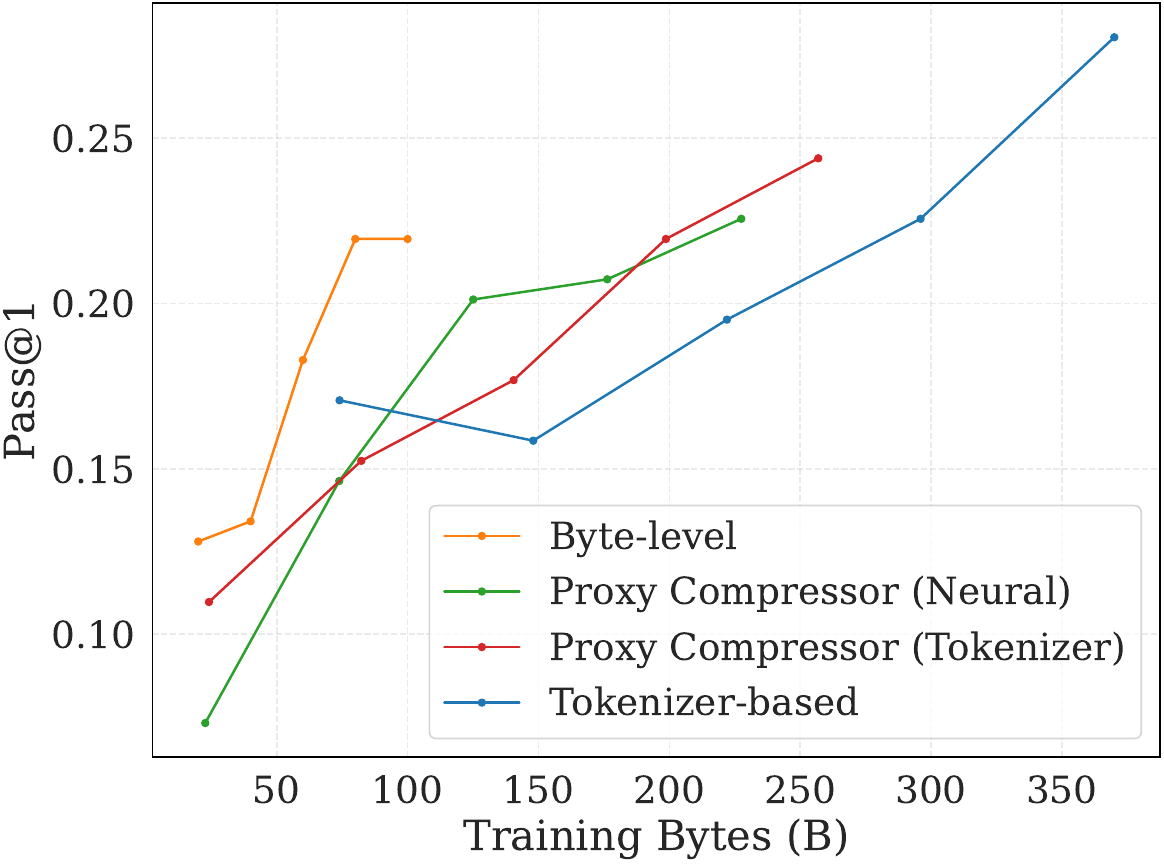}
  \caption{Pass@1 performance on HumanEval-Plus for 4B models.}
  \label{fig:4B-humaneval_pass1-vs-flops-and-data}
\end{subfigure}

\vspace{0.7em}

\begin{subfigure}[b]{0.78\textwidth}
  \centering
  \includegraphics[width=0.45\linewidth]{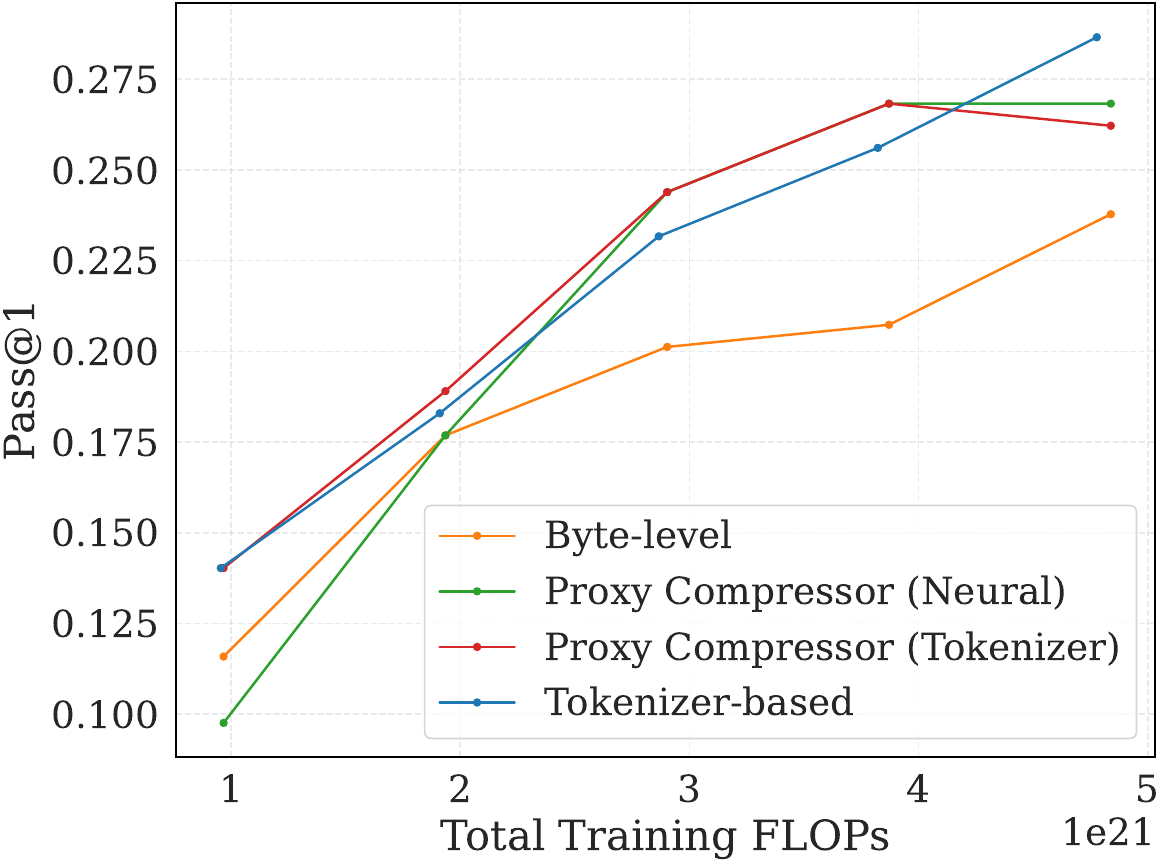}
  \hspace{0.04\linewidth}
  \includegraphics[width=0.45\linewidth]{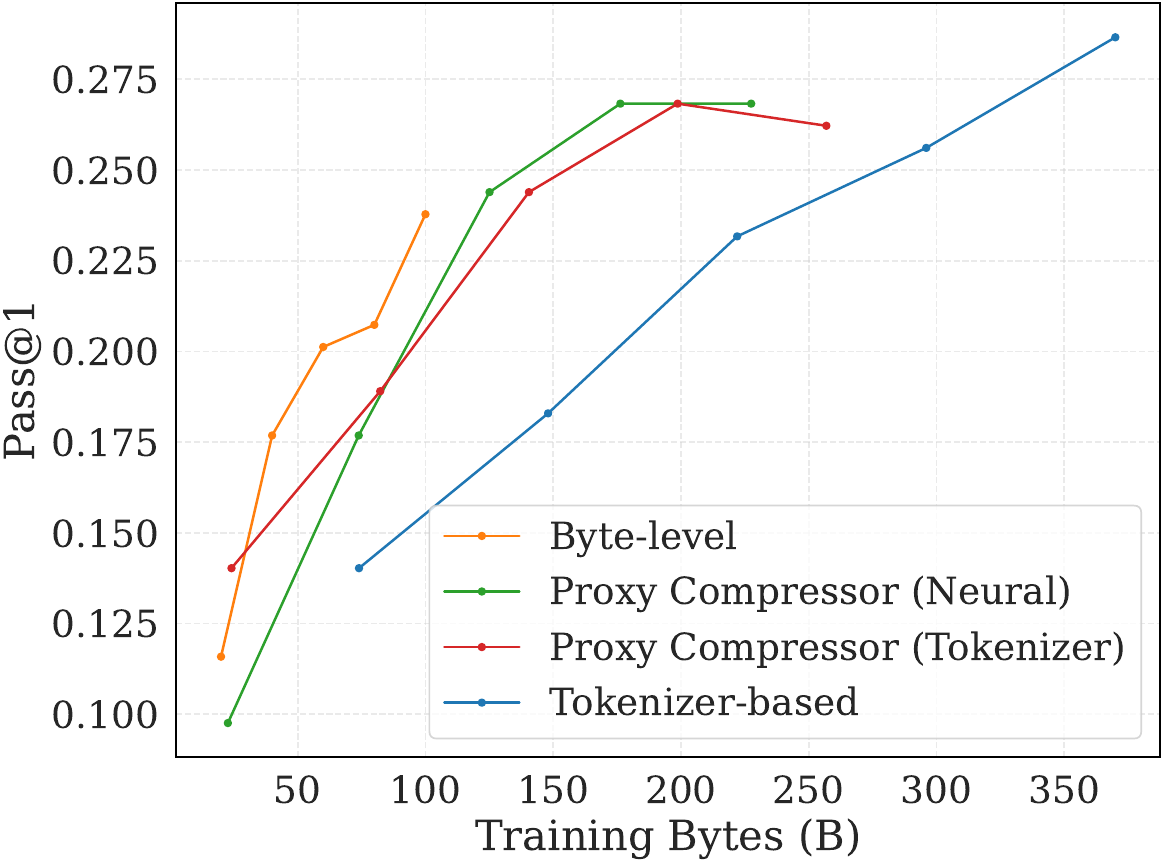}
  \caption{Pass@1 performance on HumanEval-Plus for 7B models.}
  \label{fig:7B-humaneval_pass1-vs-flops-and-data}
\end{subfigure}

\caption{Pass@1 performance on HumanEval-Plus under different input representations and model sizes, compared as a function of training FLOPs (left in each row) and amount of training data (right in each row).}
\label{fig:humaneval_pass1-vs-flops-and-data-all-scales}
\end{figure}

\paragraph{Scaling Behavior.}
\cref{fig:0B5-humaneval_pass1-vs-flops-and-data,fig:1B5-humaneval_pass1-vs-flops-and-data,fig:4B-humaneval_pass1-vs-flops-and-data,fig:7B-humaneval_pass1-vs-flops-and-data} visualize performance under matched FLOPs (left) and matched data (right) for models at 0.5B, 1.5B, 4B, and 7B parameters. At smaller scales (0.5B), proxy-trained models underperform baselines in both regimes; however, as model size increases, proxy compression becomes progressively more competitive, eventually achieving the best of both regimes at 14B scale (\cref{fig:14B-humaneval_pass1-vs-flops-and-data}). This supports our hypothesis that larger models have greater capacity to learn cross-representation alignment for effective transfer.

\begin{figure}[!t]
\centering
\begin{subfigure}[b]{0.45\columnwidth} 
  \centering
  {\includegraphics[width=\textwidth]{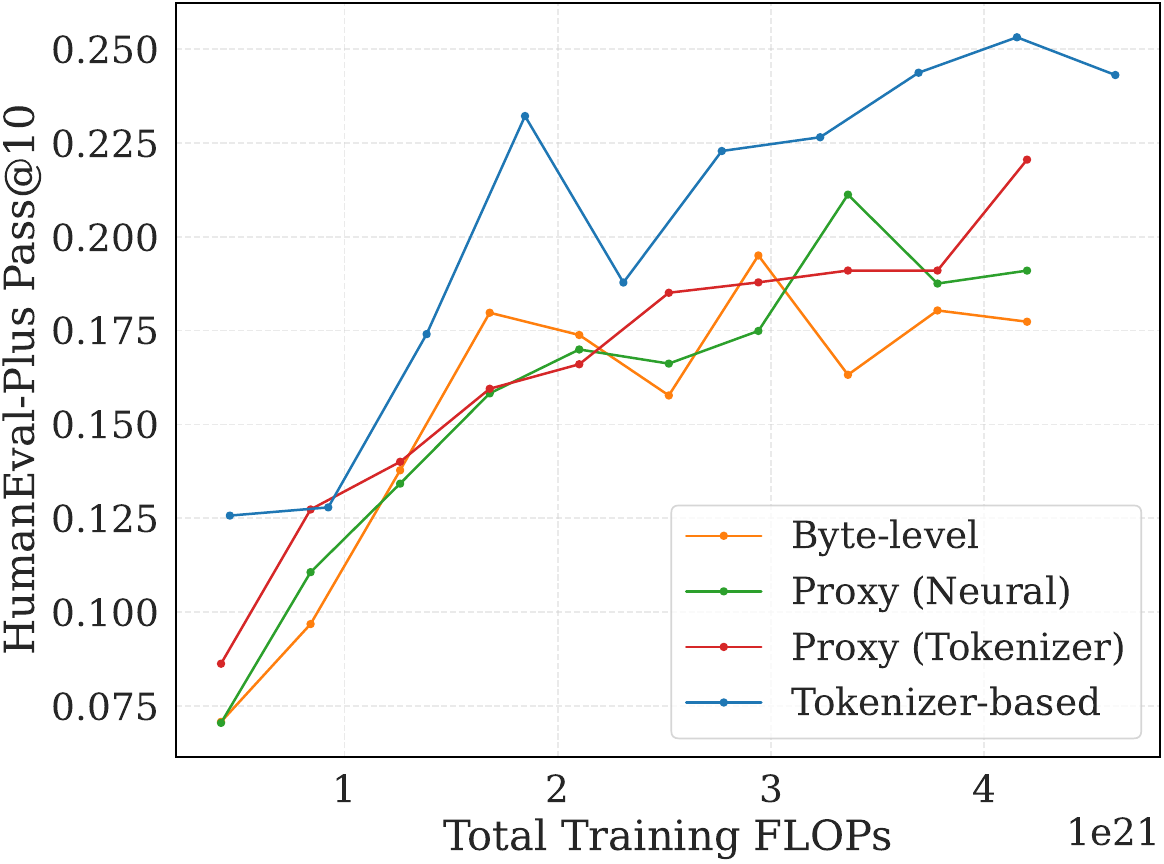}}
\end{subfigure}\hspace{0.06\linewidth}
\begin{subfigure}[b]{0.45\columnwidth}
  \centering
  {\includegraphics[width=\textwidth]{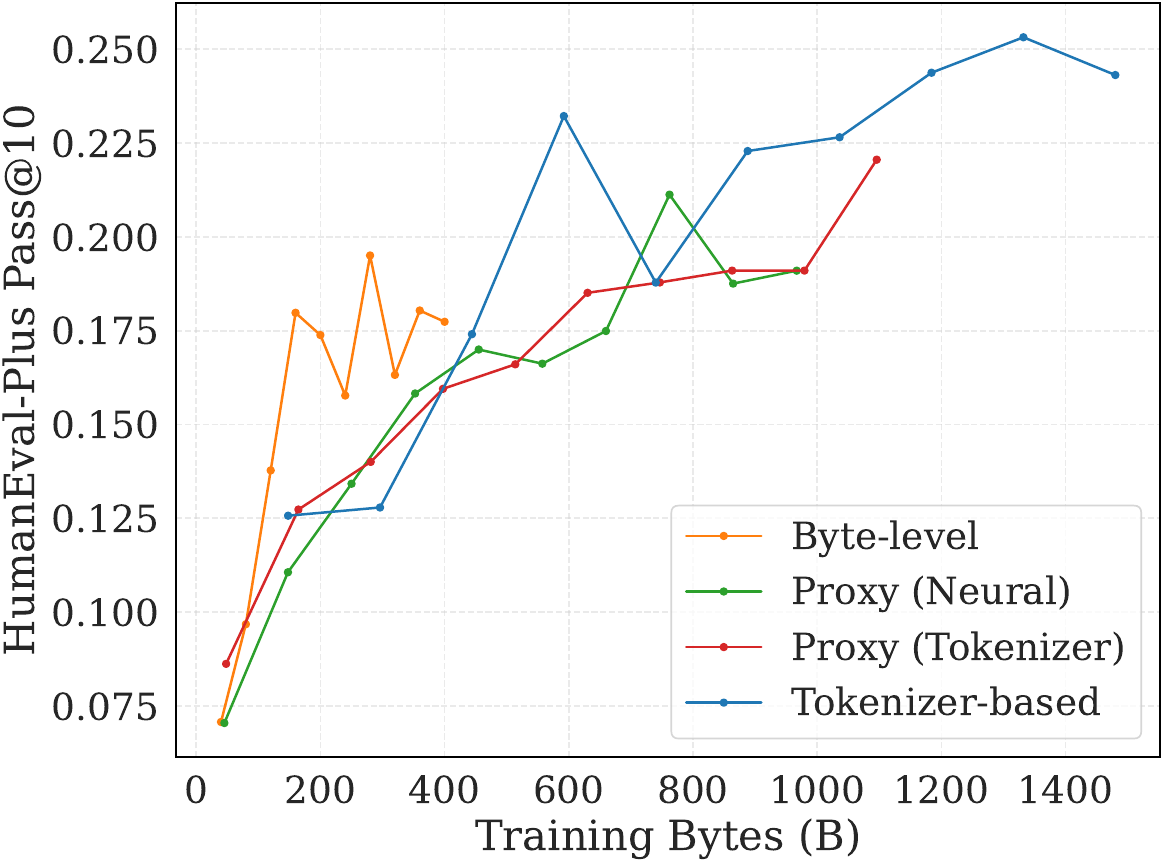}}
\end{subfigure}
\caption{HumanEval-Plus pass@10 for 1.5B models under extended training versus training FLOPs (left) and data consumed (right). We report pass@10 for stability across checkpoints during pretraining.}
\label{app:fig:longer_schedule_efficiency_1b5}
\end{figure}

\begin{figure}[!t]
\centering
\begin{subfigure}[b]{0.45\columnwidth} 
  \centering
  {\includegraphics[width=\textwidth]{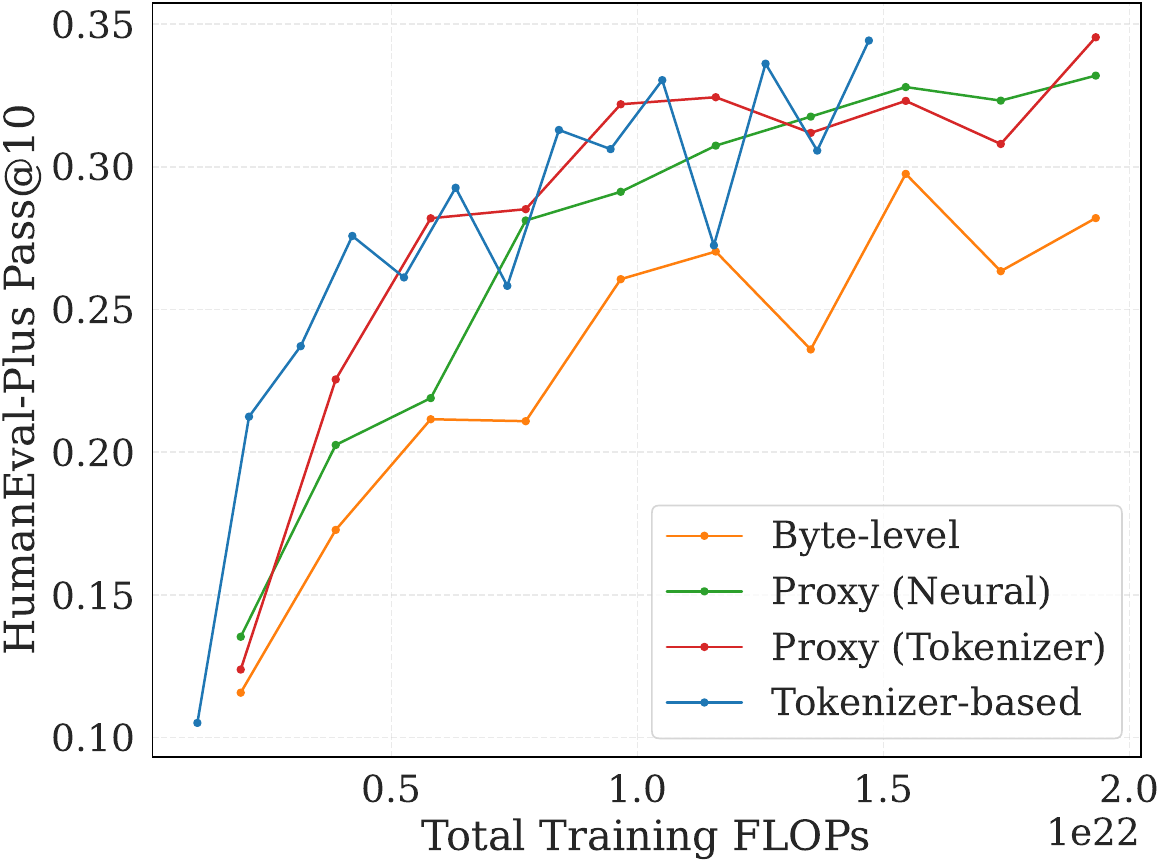}}
\end{subfigure}\hspace{0.06\linewidth}
\begin{subfigure}[b]{0.45\columnwidth}
  \centering
  {\includegraphics[width=\textwidth]{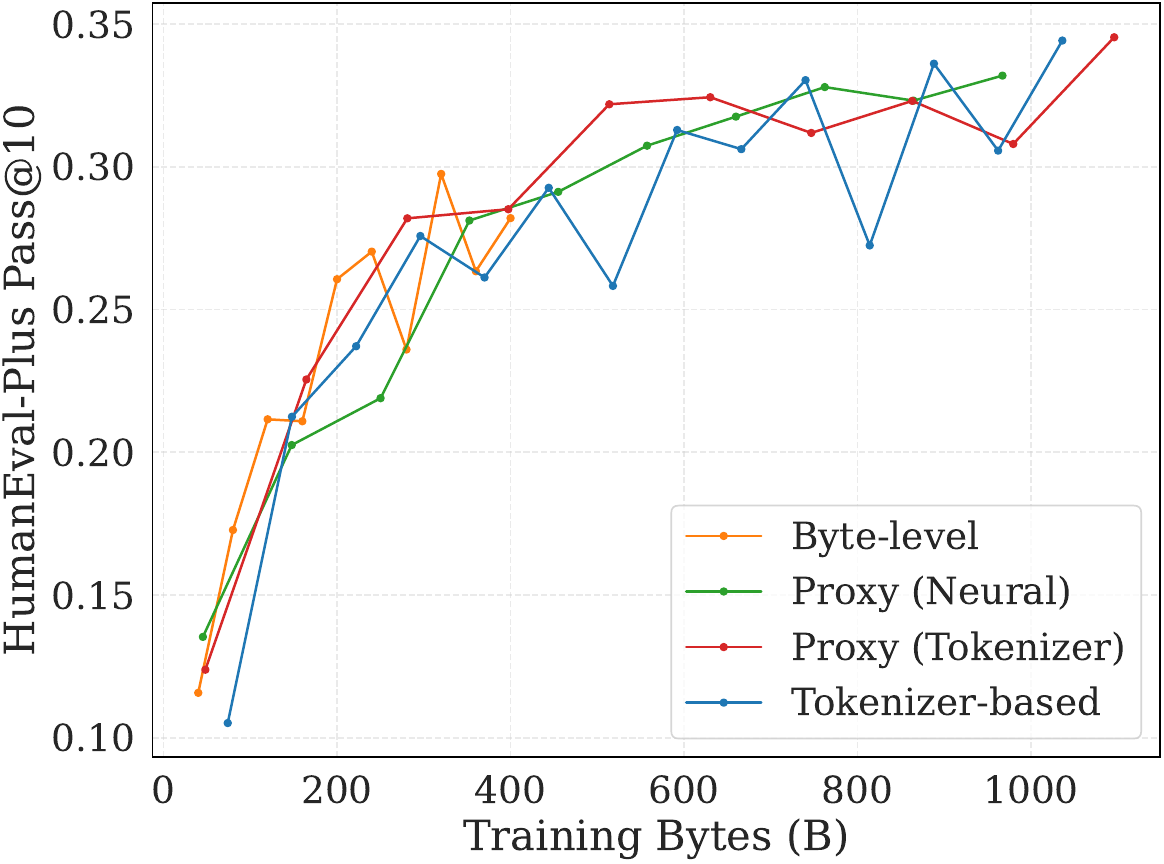}}
\end{subfigure}
\caption{HumanEval-Plus pass@10 for 7B models under extended training versus training FLOPs (left) and data consumed (right). We report pass@10 for stability across checkpoints during pretraining.}
\label{app:fig:longer_schedule_efficiency_7b}
\end{figure}

\paragraph{Longer Training Horizons.}
We also examine whether the conventional wisdom, where byte-level models are more data-efficient, tokenizer-based models are more compute-efficient, holds under different training schedules. \cref{app:fig:longer_schedule_efficiency_1b5,app:fig:longer_schedule_efficiency_7b} plot performance over longer training runs for 1.5B and 7B models, respectively. Notably, byte-level models do not consistently outperform tokenizer-based models under matched data in this regime. This suggests that the data-efficiency advantage of byte-level training might diminish given sufficient training data: the additional compute spent per byte yields diminishing returns, and compression remains beneficial for long-horizon training schedules. However, our proxy compression training still preserves the trend as observed in \cref{experiments:main-transfer}.

\paragraph{Related Observations.}
Our findings align with prior work on the nuanced relationship between compression rate and performance. SuperBPE \citep{liu2025superbpe} found that higher compression rates do not necessarily improve performance despite consuming more data per step, and discusses the associated trade-off between effective context size and training steps.\footnote{See \url{https://superbpe.github.io/faq.html\#context_adjustment} for more details.} Possible explanations: (i) higher compression produces harder tokens for the model to learn and predict; or (ii) lower compression yields more optimization steps and thus more compute per byte, allowing better fitting of the data distribution.

\subsection{Pairing Strategy Ablation}
\label{app:additional-exp-results:pairing-strategy}
As described in \cref{method:overview}, we optionally pair compressed and raw views of the same document during training to encourage explicit cross-representation alignment. \cref{tab:ablation_format} ablates this design choice along three axes: (i) \emph{pairing strategy}: whether pairs are used throughout training (Always-on), only during warmup (Warmup-only), or not at all (None); (ii) \emph{pairing order}: whether raw precedes compressed (R$\rightarrow$C), compressed precedes raw (C$\rightarrow$R), or order is randomized (R$\leftrightarrow$C); and (iii) \emph{rate warmup}: whether the mixing rate $r$ is gradually increased from $0.4$ to $0.9$ over the first 10K steps.

We observe that \emph{Warmup-only} pairing with randomized order achieves the best downstream performance (20.1\% pass@1). This aligns with findings in \cref{experiments:in-context-transfer}: while Always-on pairing yields near-perfect in-context translation, it reduces the effective compression rate by duplicating data as pairs. Warmup-only pairing provides sufficient signal for the model to learn cross-representation alignment early in training, then reverts to independent sampling to maximize data efficiency. Pairing order has a modest effect, with C$\rightarrow$R slightly under-performing other orderings. Rate warmup provides marginal gains on the performance.

\begin{table}[t]
\centering
\caption{Ablation of rate warmup and pairing design for proxy compressor training. We report HumanEval-Plus pass@1 and pass@10 on 1.5B models with tokenizer-based proxy compressors. R$\leftrightarrow$C indicates random ordering with equal probability of R$\rightarrow$C and C$\rightarrow$R, while R$\rightarrow$C and C$\rightarrow$R enforce a fixed order, where R$\rightarrow$C means that the raw sequence precedes the compressed sequence in the input.}
\label{tab:ablation_format}
\begin{tabular}{lcccc}
\toprule
Pairing Strategy & Pairing Order & Rate Warmup & Pass@1 & Pass@10 \\
\midrule
None        & ---              & \xmark & 16.7 & 25.7 \\
None        & ---              & \cmark & 16.0 & 27.6 \\
\midrule
Always-on   & R$\leftrightarrow$C & \xmark & 16.6 & 27.2 \\
Always-on   & R$\leftrightarrow$C & \cmark & 16.1 & 27.5 \\
Always-on   & R$\rightarrow$C & \cmark & 16.5 & 27.0 \\
Always-on   & C$\rightarrow$R & \cmark & 16.1 & 25.6 \\
\midrule
Warmup-only & R$\leftrightarrow$C & \cmark & 20.1 & 28.7 \\
\bottomrule
\end{tabular}
\end{table}

\subsection{Transfer Strength under Controlled Raw-Byte Exposure}
\label{app:additional-exp-results:transfer-vs-absolute-rawbytes}
To isolate the effect of transfer from the effect of increased data volume, we compare two training configurations with matched raw-byte exposure but different total training bytes. Specifically, we train 1.5B models on: (i) 95\% tokens + 5\% raw bytes for 48K steps, and (ii) 90\% tokens + 10\% raw bytes for 25K steps. Both configurations expose the model to approximately 9.1B raw bytes, but the first sees more total data due to additional compressed bytes.

\cref{tab:absolute_bytes_comp} shows that under matched raw-byte exposure, increasing total training bytes via compressed data yields higher downstream performance on byte-level inference. This confirms \emph{positive transfer} from compressed to raw representations: the model benefits from more compressed training data even when evaluated purely on raw bytes. The same pattern holds for token-level inference (15.2\% vs.\ 10.4\%), though the absolute performance is lower than byte-level interface.

\begin{table}[t]
\centering
\caption{Effect of mixing rate under fixed raw-byte exposure for 1.5B models. Performance is measured with pass@1 rates on HumanEval-Plus.}
\label{tab:absolute_bytes_comp}
\begin{tabular}{l c c c c}
\toprule
\multirow{2}{*}{Mixing Scheme} & \multirow{2}{*}{Training Steps} & \multirow{2}{*}{Raw bytes Seen} & \multicolumn{2}{c}{Pass@1 (Inference Mode)} \\ \cmidrule(lr){4-5}
& & & \multicolumn{1}{c}{Bytes} & \multicolumn{1}{c}{Tokens} \\
\midrule
95\% Tokens + 5\% Bytes  & 48000 & 9.1B & 15.9 & 15.2 \\
90\% Tokens + 10\% Bytes & 25000 & 9.1B & 12.2 & 10.4 \\
\bottomrule
\end{tabular}
\end{table}

\subsection{On the Effect of Format Sentinels}
\label{app:additional-exp-results:sentinel-ablation}
\begin{figure}[tb]
\centering
\begin{subfigure}[b]{0.45\columnwidth} 
  \centering
  {\includegraphics[width=\textwidth]{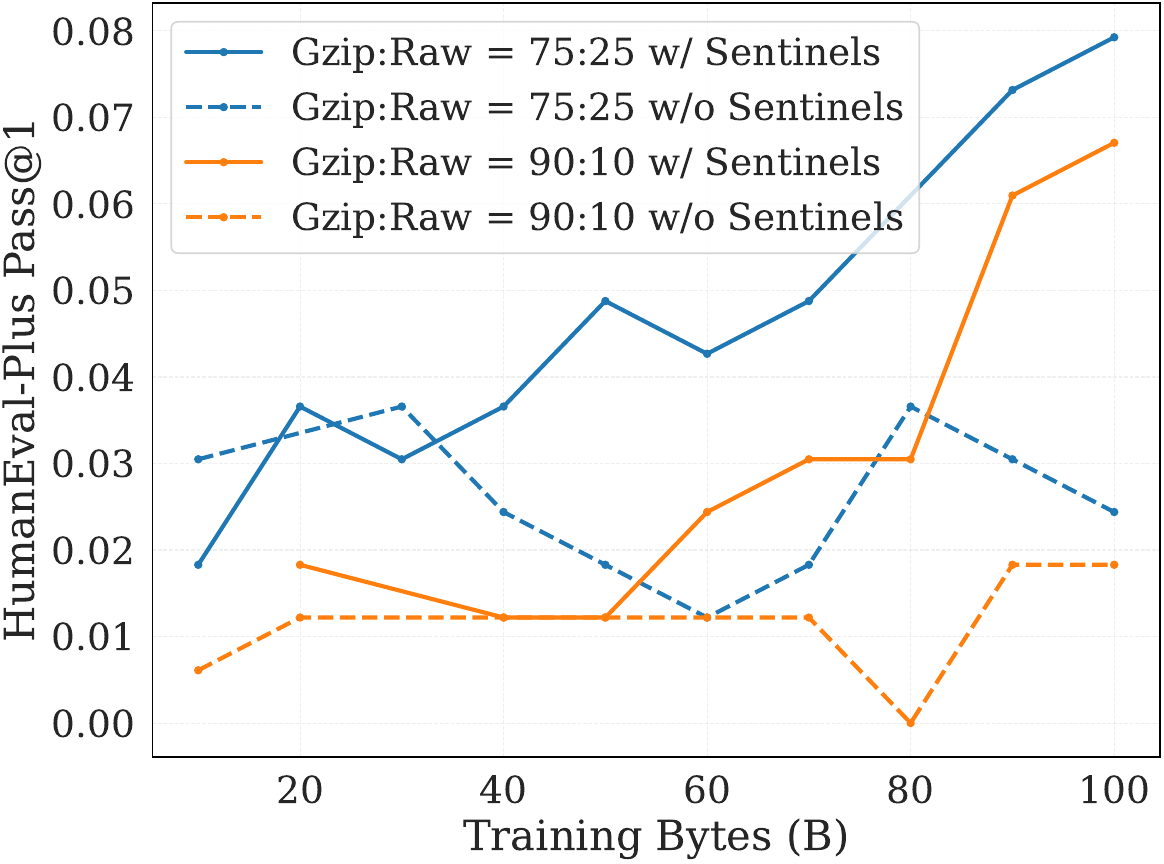}}
\end{subfigure}\hfill
\begin{subfigure}[b]{0.45\columnwidth}
  \centering
  {\includegraphics[width=\textwidth]{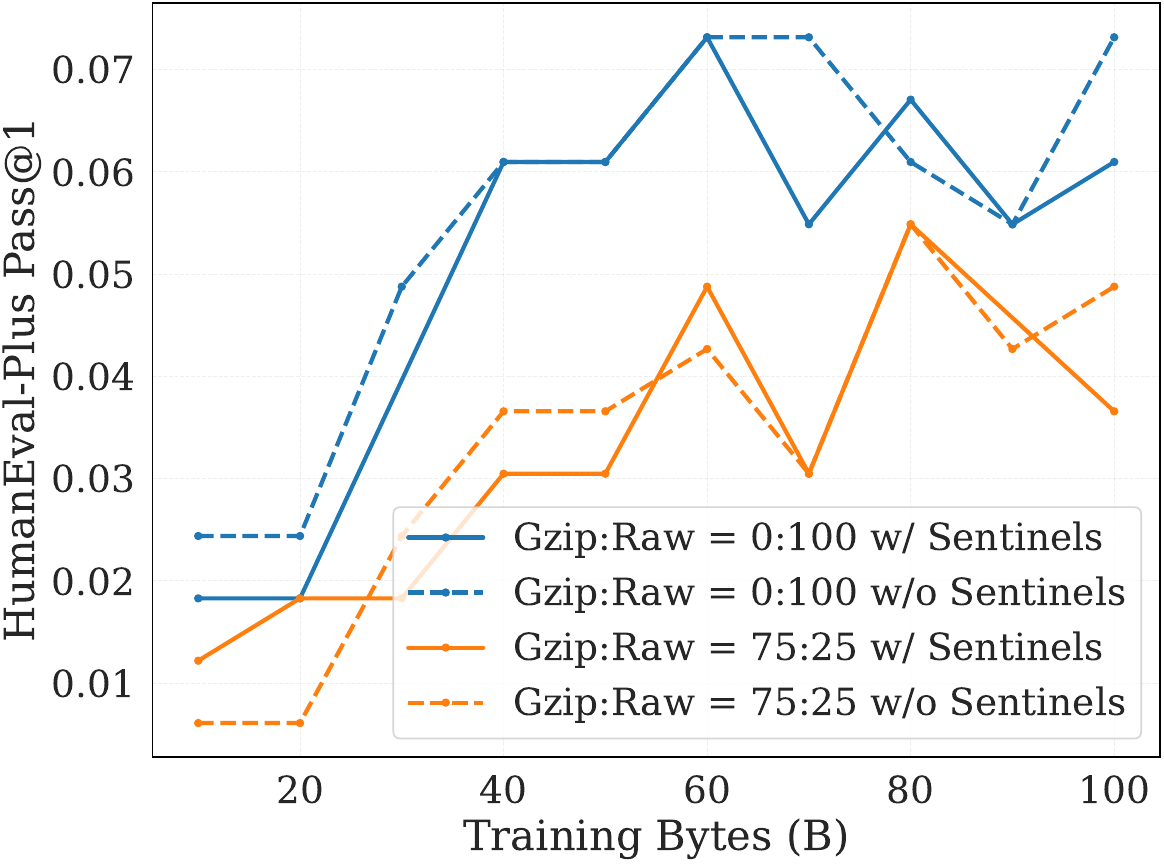}}
\end{subfigure}
\caption{Ablation on format sentinels with gzip proxy compression at 1.5B (\textbf{left}; Python corpus) and 7B (\textbf{right}; full corpus) scales.}
\label{app:fig:ablation:sentinel}
\end{figure}

Format sentinels (\cref{method:overview}) explicitly demarcate representation boundaries within packed sequences. We ablate their effect across model scales using preliminary gzip-based proxy compression. As shown in \cref{app:fig:ablation:sentinel}, sentinels significantly improve performance at the 1.5B scale, but their benefit becomes marginal at 7B. We hypothesize that larger models can more readily distinguish between representations without explicit markers. Additionally, when training pure byte models (100\% raw-format data), we observe no clear performance differences from sentinels, which is expected since no representation boundaries exist. Based on these findings, we enable format sentinels by default, as they provide benefits at smaller scales with negligible overhead at larger scales.

\subsection{Prompt Boundary Robustness}
\label{app:additional-exp-results:prompt-boundary}
A central motivation for byte-level modeling is to avoid tokenizer artifacts such as prompt boundaries, where minor whitespace or formatting changes at the end of the prompt can shift the tokenization decisions and drastically alter model behavior \citep{microsoft2023guidance,lundberg2023tokenhealing,athiwaratkun2024token,hayase2025sampling}. \cref{experiments:robustness} evaluates broad robustness via ReCode; here we complement that study with a targeted probe of prompt-boundary sensitivity. We construct 8 perturbations each appending a semantically neutral suffix (whitespace, newlines, or tabs) to a HumanEval prompt (164 problems) after stripping any existing trailing whitespace, plus an unperturbed baseline (P1 in \cref{app:tab:prompt_boundary_results}). None of these suffixes alters the semantics of the prompt, but they can lead to different tokenization boundaries, potentially producing entirely different leading token sequences at the start of generation. We evaluate the same 7B model checkpoints as in \cref{tab:downstream-transfer}.

\begin{table}[htbp]
\centering
\caption{Prompt-boundary robustness results on HumanEval (HE) and HumanEval-Plus (HE+) with 7B models. We report pass@1 (\%). Each row appends a semantically neutral suffix to the prompt after stripping trailing whitespace; P1 is the unperturbed baseline.}
\label{app:tab:prompt_boundary_results}
\begin{tabular}{l l p{3.2cm} cc cc cc}
\toprule
\textbf{ID} & \textbf{Description} & \textbf{Suffix} 
& \multicolumn{2}{c}{\textbf{Proxy (Tokenizer)}} 
& \multicolumn{2}{c}{\textbf{Byte-level}} 
& \multicolumn{2}{c}{\textbf{Tokenizer-based}} \\
\cmidrule(lr){4-5} \cmidrule(lr){6-7} \cmidrule(l){8-9}
& & & HE & HE+ & HE & HE+ & HE & HE+ \\
\midrule
P1 & Baseline & (none)
& 31.71 & 26.83 & 27.44 & 24.39 & 34.15 & 29.88 \\
P2 & Space + newline & \texttt{" \textbackslash n"} 
& 32.32 & 26.22 & 26.83 & 23.78 & \textit{0.61} & \textit{0.61} \\
P3 & Newline & \texttt{"\textbackslash n"} 
& 34.15 & 27.44 & 26.83 & 23.78 & \textit{3.05} & \textit{3.05} \\
P4 & Newline + 1 space & \texttt{"\textbackslash n~"} 
& 31.71 & 25.61 & 27.44 & 23.78 & 27.44 & 23.78 \\
P5 & Newline + 2 spaces & \texttt{"\textbackslash n~~"} 
& 32.32 & 27.44 & 28.05 & 24.39 & 26.83 & 23.11 \\
P6 & Newline + 3 spaces & \texttt{"\textbackslash n~~~"} 
& 32.32 & 25.00 & 27.44 & 23.78 & 33.54 & 29.88 \\
P7 & Newline + 4 spaces & \texttt{"\textbackslash n~~~~"} 
& 32.32 & 26.22 & 27.44 & 24.39 & 25.61 & 20.12 \\
P8 & Double newline & \texttt{"\textbackslash n\textbackslash n"} 
& 31.10 & 23.78 & 28.05 & 24.39 & \textit{0.00} & \textit{0.00} \\
P9 & Newline + tab & \texttt{"\textbackslash n\textbackslash t"} 
& 33.54 & 25.61 & 25.61 & 21.34 & 34.15 & 29.27 \\
\midrule
\multicolumn{3}{l}{Mean} 
& \textbf{32.39} & \textbf{26.02} & 27.24 & 23.78 & 20.60 & 17.74 \\
\multicolumn{3}{l}{Std. dev.} 
& 0.88 & 1.11 & \textbf{0.70} & \textbf{0.91} & 14.05 & 12.12 \\
\bottomrule
\end{tabular}
\end{table}

The tokenizer-based model exhibits severe sensitivity to prompt boundary perturbations: on HumanEval, its pass@1 rate ranges from $0.00\%$ to $34.15\%$, with two suffixes, P2 (\texttt{" \textbackslash n"}) and P8 (\texttt{"\textbackslash n\textbackslash n"}), collapsing generation to near-zero performance. Its standard deviation across the nine prompt variants is $14.05\%$ on HumanEval and $12.12\%$ on HumanEval-Plus. Operating on raw bytes at inference, the proxy-trained model treats appended characters simply as additional byte input and avoids the issue of tokenization boundaries by construction. It maintains a $32.39\%$ mean pass@1 on HumanEval with a standard deviation of only $0.88\%$, and its average performance even slightly exceeds the unperturbed baseline. The pure byte-level model is similarly robust and stable, as expected, but trails the proxy model in average performance, illustrating that proxy compression effectively combines compressed-training efficiency with byte-level inference robustness. These results directly support the prompt-boundary aspect of the motivation in \cref{introduction}.

\subsection{Natural Language Experiments}
\label{app:additional-exp-results:natural-language}

\paragraph{Setup.} To validate the effectiveness of proxy compression beyond code, we train 1.5B models on the Nemotron-CLIMB corpus~\citep{diao2026nemotronclimb} under the same architecture, optimizer, and compute-matched protocol as in \cref{experiments:setup}, swapping only the training corpus. Each run uses 50K training steps at a batch size of $0.5$M symbols, totaling $25$B symbols per run, where symbols denote tokens or bytes depending on the input representation. We compare a tokenizer-based baseline, a raw byte-level baseline, and proxy compression with a tokenizer compressor at $r=0.9$. We evaluate on $9$ standard natural language understanding benchmarks following the OLMES setup~\citep{gu2024olmes}.

\begin{table}[thb]
\centering
\caption{Downstream task accuracy on natural language understanding benchmarks for 1.5B models under compute-matched training. Proxy~(Tokenizer) is evaluated on raw bytes at inference.}
\label{tab:nl-generalization}
\resizebox{0.9\columnwidth}{!}{%
\begin{tabular}{lccccccccc|c}
\toprule
Model & ARC-c & ARC-e & BoolQ & CSQA & HellaSwag & OBQA & PIQA & SIQA & WinoGrande & Avg. \\
\midrule
Tokenizer-based   & 37.1 & 67.3 & 63.3 & 47.9 & 40.4 & 40.2 & 68.5 & 45.3 & 53.7 & 51.5 \\
Proxy~(Tokenizer) & 34.9 & 65.4 & 59.5 & 45.7 & 42.9 & 41.2 & 68.2 & 43.6 & 51.5 & 50.3 \\
Byte-level        & 35.8 & 61.4 & 56.6 & 38.9 & 44.0 & 34.0 & 67.1 & 43.1 & 53.0 & 48.2 \\
\bottomrule
\end{tabular}
}
\end{table}
\cref{tab:nl-generalization} reports task accuracy under different approaches. With raw-byte exposure on only $10\%$ of training samples, proxy compression improves over the byte-level baseline by $+2.1$ average points ($50.3$ vs.\ $48.2$), closing roughly $64\%$ of the gap between the byte-level and tokenizer-based methods. In addition, it slightly underperforms the tokenizer baseline ($51.5$), qualitatively consistent with the trend observed in our code experiments at 1.5B, where the gap gradually narrows at larger scales. We also tested whether byte-level capability can be unlocked by late raw-byte finetuning. Starting from a tokenizer-only checkpoint and finetuning for 5k steps on raw bytes improves byte inference from $36.8$ to $40.7$ average accuracy, but remains far below proxy training at $50.3$; after finetuning, token inference also drops to $37.2$, suggesting catastrophic forgetting. This supports mixed-representation training during pretraining rather than late conversion. We leave validating the scaling behavior for natural language tasks to future work.

\end{document}

%% file: math_commands.tex
\usepackage{amsmath,amsfonts,bm,mathtools}

\def\eqref#1{equation~\ref{#1}}

\def\1{\bm{1}}

\DeclareMathAlphabet{\mathsfit}{\encodingdefault}{\sfdefault}{m}{sl}
\SetMathAlphabet{\mathsfit}{bold}{\encodingdefault}{\sfdefault}{bx}{n}